\documentclass[times,twocolumn,final]{elsarticle}
\usepackage[ruled,vlined]{algorithm2e}
\usepackage{arxiv}
\usepackage{graphicx}%
\usepackage{multirow}%
\usepackage{amsmath,amssymb,amsfonts}%
\usepackage{amsthm}%
\usepackage{mathrsfs}%
\usepackage[title]{appendix}%
\usepackage{xcolor}%
\usepackage{textcomp}%
\usepackage{manyfoot}%
\usepackage{booktabs}%
\usepackage{algpseudocode}%
\usepackage{listings}%
\usepackage{multirow}
\usepackage{pifont}%
\usepackage{adjustbox}
\usepackage{graphicx}
\usepackage{hyperref}
\usepackage{float}
\usepackage{placeins}
\usepackage{fancyhdr}
\usepackage{dsfont}
\usepackage{threeparttable}
\usepackage{color,colortbl}
\definecolor{LightCyan}{rgb}{0.88,1,1}

\begin{document}

\fancypagestyle{firstpagestyle}{
    \fancyhf{} 
    \renewcommand{\headrulewidth}{0pt} 
    \fancyhead{} 
    \fancyfoot{} 
    \fancyhead[CO]{\em \fontsize{9pt}{8pt}\selectfont This article has been accepted to the Elsevier Medical Image Analysis Journal, 2026.}
}

\fancypagestyle{default}{
    \fancyhf{}
    \fancyhead[R]{\thepage} 
    \renewcommand{\headrulewidth}{0.4pt} 
    \fancyhead[LO]{\em \fontsize{9pt}{8pt}\selectfont}
}



\title{State-Change Learning for Prediction of Future Events in Endoscopic Videos}
\author[1,2]{Saurav \snm{Sharma} \fnref{corresp}}
\fntext[corresp]{Corresponding author: \texttt{ssharma@unistra.fr}}
\author[1,2]{Chinedu Innocent \snm{Nwoye}}
\author[2,3]{Didier \snm{Mutter}}
\author[1,2]{Nicolas \snm{Padoy}}

\address[1]{University of Strasbourg, CNRS, INSERM, ICube, UMR7357, France}
\address[2]{IHU Strasbourg, Strasbourg, France}
\address[3]{University Hospital of Strasbourg, France}

\received{XXX}
\finalform{XXX}
\accepted{XXX}
\availableonline{XXX}
\communicated{XXX}

\begin{abstract}
Surgical future prediction, driven by real-time AI analysis of surgical video, is critical for operating room safety and efficiency. Future prediction could provide actionable insights into upcoming events, their timing, and associated risks—enabling better resource allocation, timely instrument readiness, and early warnings for emergent complications (e.g., bleeding, bile duct injury).
Despite this need, current surgical AI research focuses on understanding what is happening rather than predicting future events. Existing methods target specific tasks such as phase or instrument anticipation in isolation, lacking unified approaches that span both short-term (action triplets, surgical events) and long-term horizons (remaining surgery duration, phase/step transitions).
These methods rely on coarse-grained supervision at the phase or instrument level, while fine-grained surgical action triplets and steps remain underexplored despite their potential to capture nuanced temporal dynamics. 
Furthermore, methods based only on future feature prediction (i.e., predicting the next-step latent visual features from current frames) struggle to generalize across different surgical contexts and procedures.
We address these limitations by reframing surgical future prediction as state-change learning. Rather than forecasting raw observations directly, our approach classifies state transitions between current and future timesteps, building transition-aware representations that improve generalization across tasks and procedures.
In this work, we introduce SurgFUTR, implementing this paradigm through a teacher-student architecture. Video clips are compressed into state representations via Sinkhorn-Knopp clustering; the teacher network learns from both current and future clips, while the student network predicts future states from current observations alone, guided by our Action Dynamics (ActDyn) module that models state transition patterns.
For comprehensive evaluation, we establish SFPBench, spanning five prediction tasks across different temporal horizons: short-term anticipation (cystic-structure triplets, surgical events) and long-term forecasting (remaining surgery duration, phase/step transitions).
Extensive experiments across four datasets spanning three laparoscopic procedures demonstrate consistent improvements over existing methods. Cross-procedure transfer from cholecystectomy to gastric bypass further validates our approach's generalizability. 
\\

\end{abstract}

\maketitle
\thispagestyle{firstpagestyle}

\begin{figure*}[!ht]
\centering
    \includegraphics[width=2.05\columnwidth]{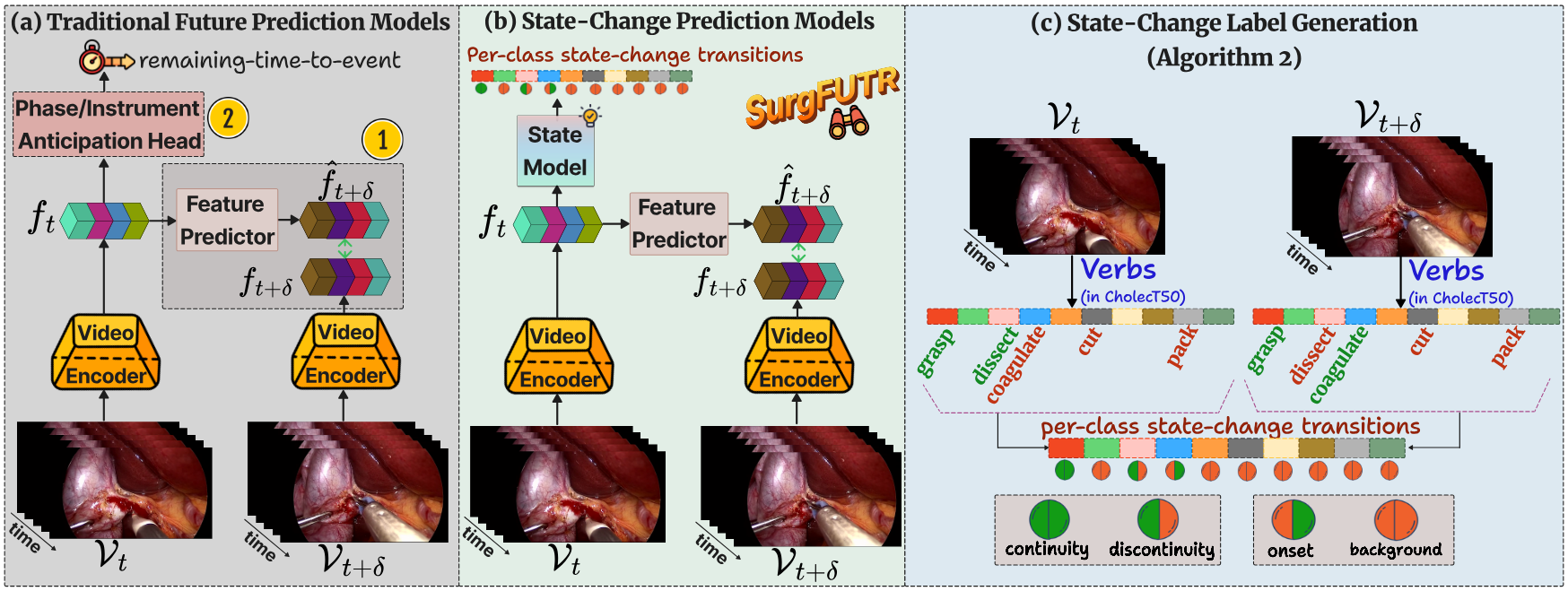} 
    \caption{Traditional anticipation models (a) predict raw future features (1) or use phase/instrument anticipation heads (2) without considering semantic state transitions. Our approach (b) reformulates anticipation as state-change classification, categorizing transitions between consecutive clips into four labels: continuity, discontinuity, onset, or background. (c) illustrates this for CholecT50 verbs: each half-circle can be green (class present) or red (class absent), with the left representing time $t$ and right representing $t+\delta$. The combination yields four state-change labels: (green, green) = continuity, (green, red) = discontinuity, (red, green) = onset, (red, red) = background. This state-transition learning builds future-aware representations that enhance downstream prediction performance.}
    \label{fig:task_intro}
\end{figure*}

\section{Introduction}
Over the past decade, surgical data science~\citep{vercauteren2019cai4cai} has increasingly centered on surgical workflow analysis, performing computational modeling of procedural context to enable context-aware systems in the operating room (OR). Computer-assisted intervention (CAI) approaches leverage rich signals from the OR, with endoscopic video emerging as a particularly informative source for understanding surgical scenes.
Building on these video signals, recent computer vision methods have advanced automatic recognition of surgical phases~\citep{twinanda2016endonet,czempiel2020tecno, jaspers2025scaling,guo2025surgical}, steps~\citep{ramesh2021multi,Lavanchy2024,ayobi2025pixel}, instrument-tissue interactions~\citep{nwoye2021rendezvous,sharma2023rendezvous,ssharma2023mcitig,li2024parameter}, surgical skill or experience assessment~\citep{aklilu2024artificial,jin2018tool,wagner2023comparative}, and assessment of critical view of safety~\citep{mascagni2021artificial,murali2023latent}. By understanding what is currently happening in surgical scenes, these recognition systems aim to enhance patient safety, mitigate intraoperative risks, and improve resource allocation, laying the foundation for robust, real-time decision support.

Despite these advances, surgical future prediction, which involves forecasting upcoming events during procedures, remains significantly underexplored despite its transformative potential for predictive decision support (e.g., resource allocation, instrument readiness, anesthesia dosing) and risk mitigation (e.g., early warnings for bleeding, vascular injury). Initial approaches targeted remaining surgery duration (RSD) estimation~\citep{twinanda2018rsdnet} and surgery type prediction~\citep{kannan2019future}, while recent efforts have expanded to anticipating phases and instruments~\citep{yuan2022anticipation,rivoir2020rethinking,boels2024supra,boels2025swag}. These anticipation capabilities are pivotal for enabling early intervention before critical events such as bile duct injury or bleeding, and for providing automatic alerts about safety-critical interactions with anatomical structures like the cystic duct and artery~\citep{mascagni2021artificial}. 
Recent datasets containing comprehensive phase and step annotations~\citep{Lavanchy2024,ayobi2025pixel} now enable the development of more granular transition prediction models.

However, existing approaches (Figure~\ref{fig:task_intro}(a)) to surgical future prediction face several key limitations that limit their clinical applicability and generalization capabilities.
\textbf{Coarse-Grained Supervision:} Current methods~\citep{yuan2022anticipation,rivoir2020rethinking,boels2024supra,boels2025swag} rely on coarse-grained supervision from phases or tool presence, overlooking fine-grained annotations that capture richer temporal dynamics. Surgical action triplets in CholecT50~\citep{nwoye2021rendezvous} detail specific instrument-tissue interactions, while procedural steps in GraSP~\citep{ayobi2025pixel} and MultiBypass140~\citep{Lavanchy2024} represent collections of triplets within surgical phases. 
These fine-grained annotations provide more detailed temporal information that could enhance surgical future prediction tasks across multiple anticipation horizons: short-term action anticipation and long-term workflow forecasting.
\textbf{Limited Spatiotemporal Modeling:} Most models rely on pre-extracted frame features aggregated by temporal models, underutilizing clip-level spatiotemporal dynamics that could provide more comprehensive temporal understanding.
\textbf{Task-Specific Approaches:} The absence of unified frameworks capable of handling both short-term anticipation and long-term forecasting limits holistic modeling capabilities in surgical domains.

In general computer vision, future prediction approaches include action anticipation~\citep{Damen2021PAMI}, which forecasts upcoming activities, and object state modeling~\citep{souvcek2022look}, which tracks object transitions between different states (e.g., initial, during action, final states) over time.
Inspired by these approaches and to address the mentioned limitations, we propose a unified deep learning framework that handles diverse surgical future prediction tasks across multiple temporal scales. We recast surgical future prediction as state-change prediction: learning compact clip-level state representations and training models to forecast how those states evolve over time (Figure~\ref{fig:task_intro}). 

Our central hypothesis is that learning to classify state changes between consecutive clips creates representations that generalize better across downstream future prediction tasks.

To implement this hypothesis, we introduce state-change supervision by labeling transitions between video clips using fine-grained annotations such as surgical action triplets and procedural steps. While related work has explored implicit state prediction for early surgery recognition~\citep{kannan2019future}, our formulation defines discrete state-change categories and uses them as explicit learning targets.

Our approach (Figure~\ref{fig:task_intro}(b)) operates as follows: given a video clip at time $t$, the model compresses spatio-temporal features into a compact state vector $\mathbf{S}_t$ via Sinkhorn-Knopp clustering, then projects it into per-class embeddings. The model learns to predict transitions from the current state at time $t$ to the future state at $t+\delta$, with each embedding aligned to discrete state-change categories (Figure~\ref{fig:task_intro}(c)). This objective compels the network to exploit spatio-temporal dynamics, yielding representations that capture temporal transition patterns and enhance performance across diverse future prediction tasks.

\textit{What do these state-change labels look like?} We leverage fine-grained annotations to define meaningful state-change categories (Figure~\ref{fig:task_intro}(c)): action labels (\textit{verb}) from surgical action triplets in CholecT50~\citep{nwoye2021rendezvous} and steps in GraSP~\citep{ayobi2025pixel}. For each class, we compare binary labels between current clip at time $t$ and future clip at $t+\delta$, assigning one of four state-change categories: continuity ($1\longmapsto1$) for persistent activities, discontinuity ($1\longmapsto0$) for concluding activities, onset ($0\longmapsto1$) for emerging activities, and background ($0\longmapsto0$) for continued absence. 
These four categories exhaustively cover all possible binary transitions, providing complete descriptions of temporal state evolution.

To learn these state-change transitions effectively, we develop SurgFUTR, a novel teacher-student framework that transforms state-change classification into a knowledge distillation problem. The teacher network processes both current and future clips to understand complete temporal transitions, while the student network learns to predict future states using only current observations, forcing it to develop predictive representations. 
To model temporal state evolution, we introduce ActDyn, a graph-based module that operates over state centroids and predicts future centroid configurations by propagating information across cluster nodes. ActDyn enables the student to learn from the teacher's future-state knowledge by minimizing divergence between predicted and target state distributions, effectively capturing temporal state dynamics without requiring future context at inference time.

\begin{figure*}[!ht]
\centering
    \includegraphics[width=2.0\columnwidth]{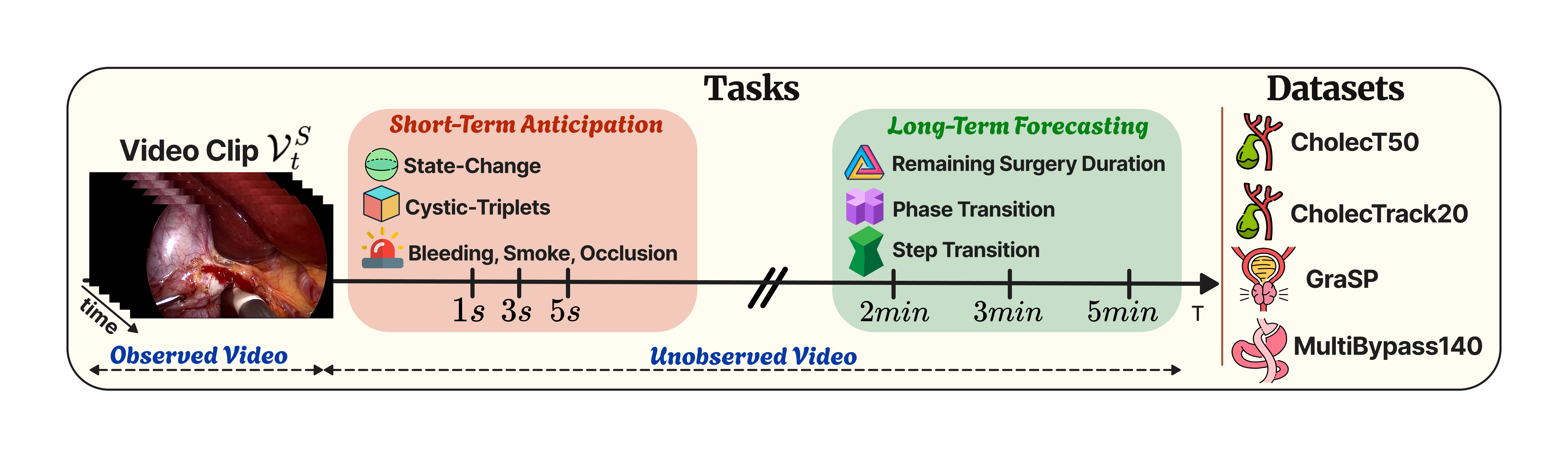} 
    \caption{SFPBench: We introduce a comprehensive benchmark for evaluating state-change pretraining across multiple surgical procedures and prediction tasks. The benchmark encompasses four datasets spanning three laparoscopic surgical procedures (cholecystectomy, robotic-assisted radical prostatectomy, gastric bypass) and includes both long-term forecasting tasks (remaining surgery duration, phase/step transitions) and short-term anticipation tasks (cystic-structure related action triplets, surgical events). Cross-procedure transfer evaluation from cholecystectomy to gastric bypass validates generalization capabilities. See Section~\ref{task_desc} for details.}
    \label{fig:task_overview}
\end{figure*}

We introduce SFPBench (\textbf{S}urgical \textbf{F}uture \textbf{P}rediction \textbf{Bench}mark) for comprehensive evaluation, bringing together datasets from CholecT50~\citep{nwoye2021rendezvous}, GraSP~\citep{ayobi2025pixel}, CholecTrack20~\citep{nwoye2023cholectrack20}, and MultiBypass140~\citep{Lavanchy2024}.
As illustrated in Figure~\ref{fig:task_overview}, SFPBench encompasses five prediction tasks across different temporal horizons: three long-term forecasting tasks (remaining surgery duration, phase transition, and step transition) and two short-term anticipation tasks (cystic-structure triplet anticipation and surgical event anticipation). We pretrain SurgFUTR variants using state-change prediction with fine-grained supervision (verb or step labels) and evaluate their transfer performance across all SFPBench tasks, showing generally favorable and often competitive performance relative to strong baselines, including recent surgical foundation models.

Our contributions are summarized as follows:

\begin{enumerate}
    \item \textbf{State-Change Learning Formulation:} We introduce a state-change learning objective that unifies future prediction across multiple anticipation horizons by learning anticipative features from temporal state transitions.
    \item \textbf{SurgFUTR Architecture:} We develop SurgFUTR, a teacher-student framework that models semantic transitions between current and future video clips, training the student to predict future states from present observations through our novel ActDyn module.
    \item \textbf{Surgical Future Prediction Benchmark (SFPBench):} We introduce a comprehensive benchmark spanning multiple surgical procedures and annotation granularities.
    \item \textbf{Comprehensive Evaluation:} We present extensive evaluation across SFPBench using complementary metrics and ablation studies to analyze the contribution of the proposed framework components.
    \item \textbf{Cross-Procedure Transfer Learning:} We demonstrate effective transfer of state-change pretrained models from cholecystectomy to gastric bypass procedures, validating generalization across different surgical contexts.
\end{enumerate}

\section{Related work}
\subsection{Surgical Workflow Analysis}
The automatic extraction of surgical workflows has been a longstanding area of interest for enhancing context-aware decision support systems~\citep{maier2017surgical}. Traditional methods~\citep{padoy2012statistical,blum2008workflow} primarily focused on instrument utilization as a proxy for identifying surgical phases, utilizing Hidden Markov Models (HMMs) to classify different phases with validation provided by surgeons. 
The emergence of deep learning introduced new possibilities through endoscopic video analysis. Modern approaches~\citep{twinanda2016endonet,dergachyova2016automatic,czempiel2020tecno,funke2018temporal,jin2017sv} moved beyond instrument tracking to leverage comprehensive visual information for automated phase recognition, demonstrating significant improvements in understanding surgical workflows and enabling more robust phase identification. Recent methods~\citep{ramesh2023dissecting,batic2024endovit,jaspers2025scaling} have focused on self-supervised learning and dataset scaling to demonstrate generalization capabilities, primarily for phase recognition tasks.
While phase recognition represents a crucial advancement, surgical procedures require more detailed analysis. Subsequent work~\citep{ramesh2021multi,Lavanchy2024,ayobi2025pixel,valderrama2022towards,huaulme2021micro} has introduced surgical steps as fundamental units within each phase, representing specific actions required to complete surgical phases and providing the granular understanding necessary for effective surgical assistance systems.
However, steps still provide only a broad view of instrument-tissue interactions. Surgical action triplets~\citep{nwoye2021rendezvous} provide the most detailed semantic representation, explicitly encoding which instrument performs what action on which anatomical structure. This hierarchical progression from phases to steps to triplets provides increasingly detailed semantic detail for surgical workflow analysis.

\subsection{Surgical Action Triplets}
Instrument-tissue interaction represents the fundamental level of activity in surgical procedures, capturing detailed nuances of surgical actions. Initial formulations of these interactions used ontological concepts~\citep{neumuth2009validation,katic2014knowledge}, defining surgical activities through instrument actions on specific anatomical structures. In laparoscopic cholecystectomy procedure, these interactions are formalized as \textlangle{}{\textit{instrument, verb, target}}\textrangle{} triplets. The CholecT40 dataset~\citep{nwoye2020recognition}, comprising 40 videos, introduced this representation along with Tripnet, a multi-task model incorporating weak instrument localization feature maps. This framework evolved with CholecT50~\citep{nwoye2021rendezvous}, expanding to 50 videos and introducing the attention-based Rendezvous~\citep{nwoye2021rendezvous} approach that leverages features from each triplet component for improved prediction through attention-based reasoning. Subsequent works enhanced this foundation: Rendezvous-in-Time~\citep{sharma2023rendezvous} exploited temporal information of triplets to improve interaction recognition performance, while MCIT-IG~\citep{ssharma2023mcitig} addressed the absence of instance-level information by incorporating an instrument detector and graph formulation to connect instrument instances with target class embeddings for triplet detection on the detection test set of MICCAI2022 CholecTriplet Challenge~\citep{nwoye2023cholectriplet2022}. 
Similar approaches have been adopted across surgical domains: SARAS-ESAD~\citep{bawa2021saras} uses \textlangle{}{\textit{verb, target}}\textrangle{} pairs with bounding boxes for prostatectomy, other prostatectomy datasets~\citep{ayobi2025pixel,valderrama2022towards} focus on \textlangle{}{\textit{action}}\textrangle{} recognition alone, and cataract surgery work~\citep{lin2022instrument} extends triplets with bounding box annotations.

\subsection{Surgical Future Prediction}
Anticipating future events in surgical workflow analysis holds immense value for improving patient safety through predictive decision support (e.g., resource allocation, instrument readiness) and risk mitigation (e.g., early warnings for bleeding, vascular injury, bile-duct injuries, visual challenges like smoke and occlusion)~\citep{wei2021intraoperative,bose2025feature,nwoye2023cholectrack20,mascagni2021artificial}. Predicting remaining surgery duration helps anesthesiologists optimize medication timing, while forecasting upcoming phase transitions allows nurses to prepare specific instrument sets and surgical teams to coordinate role changes. In laparoscopic cholecystectomy procedures, such anticipation capabilities enable early intervention and automatic alerts about safety-critical interactions with anatomical structures such as the cystic duct and cystic artery.

Anticipation tasks in surgery can be conceptually divided into two types: long-term forecasting (minutes) and short-term anticipation (seconds). \textbf{Long-term forecasting} (minutes) focuses on predictions over extended horizons to support preparation and resource management. Early research addressed remaining surgery duration (RSD) prediction, crucial for determining anesthesia requirements. RSDNet~\citep{twinanda2018rsdnet} introduced a self-supervised approach that simultaneously predicted surgical progress and remaining duration through multi-task learning across Cholecystectomy and Bypass procedures. A parallel development~\citep{kannan2019future} addressed early surgery type identification using a teacher-student framework where future frame features at time $t+\delta$ were distilled into a student model with access only to the current frame at time $t$. More recent work has focused on predicting time until specific events such as instrument usage~\citep{rivoir2020rethinking} or phase transitions~\citep{yuan2022anticipation}, evaluated at 2-5 minute intervals to support instrument readiness~\citep{maier2017surgical} and operating room coordination. 

However, existing methods face significant limitations. IIA-Net~\citep{yuan2022anticipation} requires instrument bounding boxes and segmentation masks in a two-stage setup for phase transition tasks, introducing additional complexity and dependency on accurate segmentation pipelines. Recent methods SuPRA~\citep{boels2024supra} and SWAG~\citep{boels2025swag} enhance phase recognition by incorporating future predictions as auxiliary tasks, but SWAG's reliance on ground truth transition probabilities from training data limits its generalizability to unseen surgical variations and procedures.
\textbf{Short-term anticipation} (seconds) has received limited attention in the surgical domain due to the inherent difficulty of modeling rapid, fine-grained surgical dynamics and the lack of densely annotated datasets at sub-minute temporal resolutions. A notable exception is HGT~\citep{yin2024hypergraph}, which employs a hypergraph structure derived from $\langle$\textit{instrument, verb, target}$\rangle$ components to anticipate various elements: all triplets~\citep{nwoye2020recognition}, triplets that occur in \textit{clipping and cutting} phases, and critical view of safety (CVS)~\citep{rios2023cholec80} for durations up to 10 seconds. These approaches generate predictions for specific events within limited time windows where preventive or preparatory actions can be taken effectively. To address these limitations, we propose a unified framework based on state-change pretraining that can perform effectively across multiple temporal horizons without requiring additional annotations or ground truth dependencies.

\subsection{State-Change Modeling}
State-based modeling approaches in computer vision typically focus on directly predicting future states or using implicit state representations. In surgical domains, early surgery identification~\citep{kannan2019future} employed LSTM internal cells as implicit states for future feature prediction. The natural image domain has explored explicit state-change classification through the ChangeIt dataset~\citep{souvcek2022look}, which manually annotates videos into four states (background, initial, action, end) to capture natural progression in untrimmed videos.

We take a fundamentally different approach by learning state-change prediction as a pretraining objective rather than directly predicting future states. Instead of forecasting what will happen, we train models to recognize how states transition between time steps $t$ and $t+\delta$. This state-change supervision infuses learned features with future predictive context, enabling better transfer to diverse downstream anticipation tasks. Our hypothesis is that understanding state transitions provides richer temporal representations than direct future prediction alone.
\section{Methodology}
\begin{figure*}[h]
\centering
    \includegraphics[width=2.0\columnwidth]{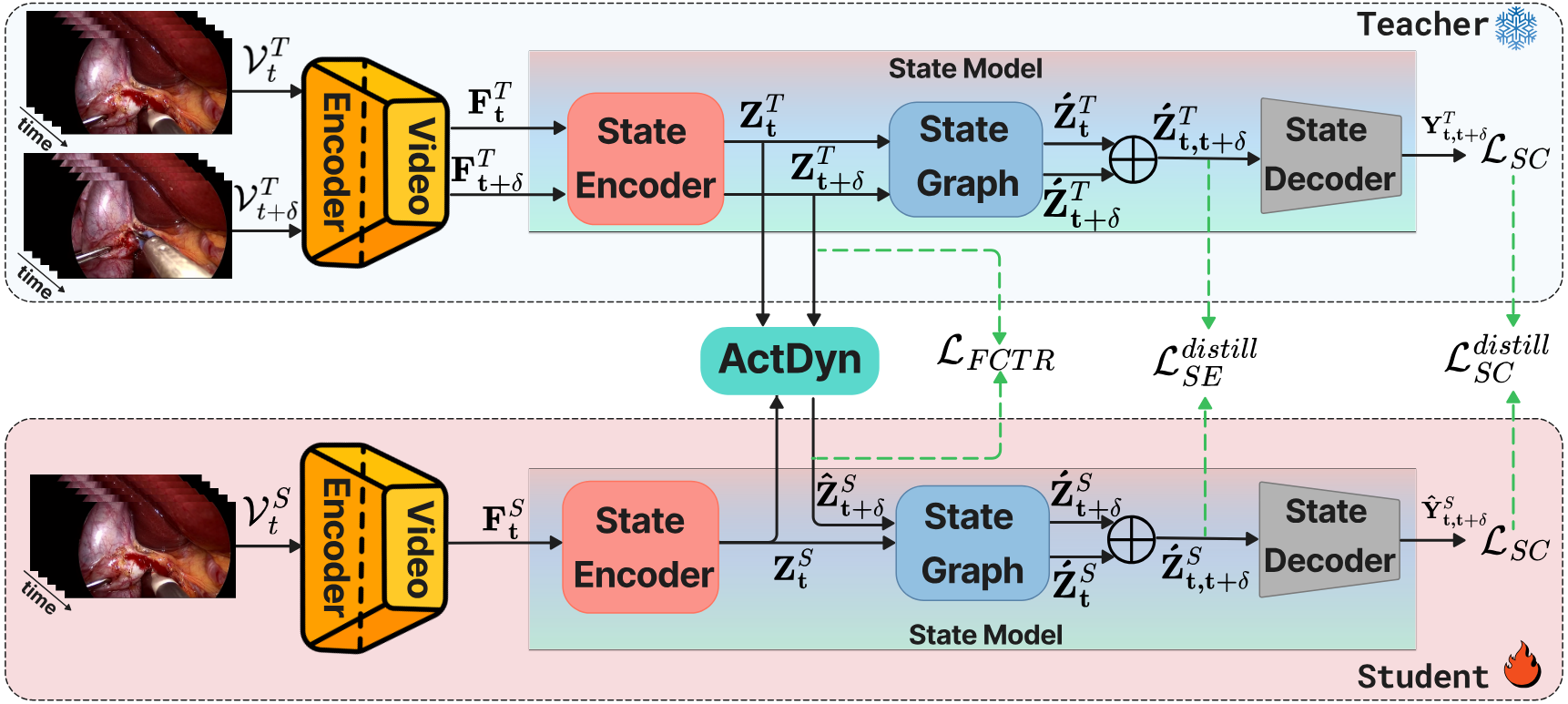}
    \caption{\textbf{SurgFUTR:} a teacher-student framework comprising: (a) a video encoder for spatio-temporal feature extraction; (b) a state encoder using Sinkhorn-Knopp clustering to generate centroid-based state representations; (c) a state graph enabling cross-cluster information exchange through message passing; (d) an Action Dynamics (ActDyn) module predicting transitions from current-clip to future-clip centroids; and (e) a state decoder mapping states to per-class embeddings for state-change classification.}
    \label{fig:scadyn}
\end{figure*}

In this section, we present \textbf{SurgFUTR}, our deep learning framework that enhances surgical future prediction through state-change modeling. While existing approaches rely on direct feature prediction or general video understanding, we hypothesize that explicitly modeling state transitions provides superior temporal representations for surgical anticipation tasks. We first formulate the state-change classification task that serves as the core pretraining objective, then present three architectural variants: SurgFUTR-Lite (future feature prediction), SurgFUTR-S (adds clustering-based state representations), and SurgFUTR-TS (full teacher-student framework). The complete architecture is illustrated in Figure~\ref{fig:scadyn}.

\subsection{State-Change Task Formulation}
At its core, \textbf{SurgFUTR} employs a state-change classification objective that unifies short and long-horizon forecasting under supervision from structured labels: verbs in action triplets~\citep{nwoye2021rendezvous} and steps~\citep{ayobi2025pixel}. Our state-change task design is based on a key hypothesis: learning semantic transitions between video clips at time steps $t$ and $t+\delta$ using state-change labels enables learning contextual cues beneficial for downstream future prediction tasks. 
Based on binary labels at time $t$ and $t+\delta$, we generate state-change categories that capture how surgical procedures evolve over time. For each class, the binary labels indicate presence (1) or absence (0) of that semantic content at each time step. Inspired by the ChangeIt~\citep{souvcek2022look} framework, we define four state change labels per class (illustrated in Figure~\ref{fig:task_intro}): Continuity, Discontinuity, Onset, and Background.

\paragraph{\textbf{Continuity ($1\longmapsto1$)}} A class is present at both time steps $t$ and $t+\delta$. For instance, the \textit{grasp} action persists throughout the \textit{calot triangle dissection} phase in Cholec80~\citep{twinanda2016endonet}. 

\paragraph{\textbf{Discontinuity ($1\longmapsto0$)}} A class is present at time step $t$ but absent at $t+\delta$. Consider the \textit{dissect} action during \textit{calot triangle dissection} that disappears when transitioning to the \textit{clipping and cutting} phase. This captures when an ongoing activity concludes as new actions or procedural steps begin.

\paragraph{\textbf{Onset ($0\longmapsto1$)}} A class that is absent at time step $t$ becomes present at $t+\delta$. This transition directly mirrors anticipation tasks by capturing the emergence of new semantic content from current video context. For instance, when \textit{dissect} actions conclude, \textit{clip} actions typically emerge during the transition, followed by \textit{cut} actions once clipping is complete. The onset label captures this predictive emergence of future activities.

\paragraph{\textbf{Background ($0\longmapsto0$)}} A class remains absent at both time steps $t$ and $t+\delta$. For example, during the \textit{preparation} phase in Cholec80~\citep{twinanda2016endonet}, actions like \textit{clip} or \textit{cut} are not observed at either time step.

\paragraph{\textbf{Summary}} By framing future prediction as state-change learning, we transform the anticipation problem from predicting specific future details to understanding how semantic transitions occur. This formulation enables the model to learn discriminative spatio-temporal cues and improve generalization across diverse future prediction tasks.

\subsection{Architecture Overview}
\textbf{SurgFUTR} comprises four key components: (1) a feature extraction backbone (Section~\ref{sec:feat_backbone}), (2) a state encoder with clustering (Section~\ref{sec:state_encoding}), (3) a state graph (Section~\ref{sec:state_graph}), and (4) a state decoder (Section~\ref{sec:state_decoding}). We employ a two-stage teacher-student distillation (Section~\ref{sec:ts_distillation}) pipeline for state-change classification, featuring a novel module that predicts future states from current video clip state vectors. 
As a robust baseline, we develop \textbf{SurgFUTR-Lite}, a direct future feature prediction model without explicit state modeling (Section~\ref{sec:surgfutr_lite}).

\subsection{Feature Extraction Backbone}
\label{sec:feat_backbone}
Although frame-based backbones can be applied to videos via per-frame processing, treating frames independently prevents them from modeling temporal dynamics. To extract spatio-temporal features from video clips, we employ a Vision Transformer (ViT)~\citep{dosovitskiy2020image} pretrained using the VideoMAEv2~\citep{wang2023videomaev2} \textit{mask-and-predict} pretraining strategy. The VideoMAEv2-based ViT models spatio-temporal structure by forming tubelets—patches extended across multiple frames—and processing their tokens through a stack of transformer layers. 
Given a clip $\mathbf{V}_t \in \mathbb{R}^{C \times T \times H \times W}$, the backbone outputs features $\mathbf{F}_t \in \mathbb{R}^{N \times d}$, where $N = (T/\tau) \times (H W / P^2)$ with spatial patch size $P$, temporal tubelet size/stride $\tau=2$, and embedding dimension $d$. This patch-level tokenization enables finer control over how visual entities and their parts contribute to the task.
\subsection{State Encoder}
\label{sec:state_encoding}
Recognizing state changes between video clips requires understanding essential scene dynamics and semantics. Raw high-dimensional features are computationally expensive and inefficient for downstream reasoning. To address this, we distill each clip into a compact state vector: an abstract, object-centric representation that reduces computational overhead while preserving semantic content for effective state-change analysis.

A straightforward yet powerful strategy is to cluster the spatio-temporal features $\mathbf{F}_t$ extracted from the video backbone. Our approach draws inspiration from CrOC~\citep{stegmuller2023croc}, which leverages the Sinkhorn-Knopp algorithm~\citep{cuturi2013sinkhorn} for object-centric clustering. In CrOC, two augmented views of the same image are concatenated, yielding features of shape $\mathbb{R}^{2N \times d}$, where $N$ is the number of tokens per view, $d$ is the feature dimension. These concatenated features are then clustered into $K$ centroids using the Sinkhorn-Knopp algorithm, producing a discrete and structured state representation.
By adopting this clustering-based abstraction, we transform the rich feature space into discrete states representing different scene configurations. This allows us to model temporal dynamics through state transitions rather than direct feature prediction.

In our implementation, we adapt the clustering approach in CrOC~\citep{stegmuller2023croc} to the video domain by clustering spatio-temporal features $\mathbf{F}_t \in \mathbb{R}^{N \times d}$, where $N$ is the number of spatio-temporal tokens. The attention map $A \in \mathbb{R}^{N \times N}$, derived from the model’s self-attention, is used to guide the clustering.
The attention matrix $A_{ij}$ from the last layer, averaged across heads, contains overall contribution from token $i$ to token $j$. Specifically, we aggregate $A$ to obtain a \textit{row marginal distribution} $m_r$ over tokens, reflecting their semantic importance in the sequence. 
As a result, tokens with greater saliency in the attention map are given more weight during clustering.

Unlike CrOC, which uses the [CLS] token’s attention as an external prior and iteratively merges centroids to adaptively determine the number of clusters, our method fixes the number of clusters $K$ and performs clustering in a single step. This design choice significantly reduces computational overhead and runtime, which is especially beneficial for long video sequences. The attention-guided marginals ensure that a fixed $K$ still yields semantically meaningful clusters, making adaptive cluster selection less critical.
Based on the row marginals $m_r$, we sample $K$ tokens to serve as centroids, resulting in $\mathbf{C}_t \in \mathbb{R}^{K \times d}$, and construct a binary assignment matrix $M_t \in \{0,1\}^{N \times K}$. We then compute the cost matrix $P_{\text{cost}} \in \mathbb{R}^{N \times K}$, which measures the (negative) similarity between each token and each centroid:

\begin{equation}
\mathbf{P}_{\text{cost}} = - \mathbf{F}_t (\mathbf{C}_t)^\top
\end{equation}

We then compute \textit{column marginals} $m_c$ to represent the distribution over centroids. The Sinkhorn-Knopp algorithm then solves for the optimal assignment matrix $\Pi_t^*$:

\begin{equation}
\Pi_t^* = \underset{\Pi_t \in \mathbf{U}(m_r, m_c)}{\arg\min} \langle \Pi_t, \mathbf{P}_{\text{cost}} \rangle - \frac{1}{\lambda} \mathrm{H}(\Pi_t),
\label{sinkhorn}
\end{equation}

where $\mathbf{U}(m_r, m_c)$ is the set of matrices with row and column sums $m_r$ and $m_c$, respectively, and $\mathrm{H}(\cdot)$ denotes the entropy regularization.
After $r$ Sinkhorn iterations, we obtain the final soft assignment matrix $\Pi_t^* \in \mathbb{R}^{N \times K}$, where $\Pi_t^*[i,k]$ denotes the soft assignment (e.g., probability) of token $i$ to cluster $k$ (rows sum to 1, and columns match the target cluster distribution).
To summarize the video’s spatio-temporal dynamics, we compute the state vector $\mathbf{Z}_t \in \mathbb{R}^{K \times d}$ as a weighted aggregation of the token features, with each centroid vector formed by pooling the features according to their assignment probabilities:
\begin{align}
\mathbf{Z}_t &= (\Pi_t^*)^\top \mathbf{F}_t
\end{align}
where the $k$-th row of $\mathbf{Z}_t$ is computed as $\mathbf{Z}_t[k, :] = \sum_{i=1}^N \Pi_t^*[i, k]\, \mathbf{F}_t[i, :]$. 
Our novel approach generates state vectors that capture both temporal dynamics and semantic structure. 
The resulting clusters are object-centric, highlighting surgical instruments and their interaction zones with anatomical structures, yielding interpretable representations. Moreover, the clustering process operates without trainable parameters, relying solely on the pre-computed attention matrix from the ViT backbone.


\subsection{State Graph}
\label{sec:state_graph}
As shown in Figure~\ref{fig:scadyn}, we perform message passing over the $K$ centroid features to model their interactions and reduce dimensionality. We first apply a linear projection layer $\phi_1$ to the centroid feature $\mathbf{Z}_t \in \mathbb{R}^{K \times d}$, obtaining $\mathbf{\tilde{Z}}_t = \phi_1(\mathbf{Z}_t) \in \mathbb{R}^{K \times d_1}$. We then $\ell_2$-normalize the rows of $\mathbf{\tilde{Z}}_t$ and construct an adjacency matrix from their pairwise similarities as:
\begin{align}
\mathbf{D}_{ij} &= \left\| \mathbf{\tilde{Z}}_i - \mathbf{\tilde{Z}}_j \right\|_2, \\
\mathbf{Adj}_{ij} &= \frac{\exp\left(-\frac{\mathbf{D}_{ij}}{\tau_{1}}\right)}{\sum_k \exp\left(-\frac{\mathbf{D}_{ik}}{\tau_{1}}\right)}, \\
\mathbf{Adj}_{ij} &= \mathbb{I}\left[\mathbf{Adj}_{ij} > \theta\right],
\end{align}
where $\mathbf{Adj}$ is the adjacency matrix, $\mathbf{D}$ is the Euclidean distance matrix, $\tau_1$ is a temperature parameter, $\theta$ is a similarity threshold, and $\mathbb{I}$ denotes the indicator function. Finally, we apply a single layer GATv2 graph attention network~\citep{brody2021attentive} ($\mathbf{G}_t$) to the projected centroid representations $\mathbf{\tilde{Z}}_t$. This computes attention-weighted aggregations over centroids, enabling adaptive information integration based on learned coefficients. The output is refined centroid features $\mathbf{\acute{Z}}_t = \mathbf{G}_t(\mathbf{\tilde{Z}}_t) \in \mathbb{R}^{K \times d_1}$.

\subsection{State Decoder}
\label{sec:state_decoding}

To decode the state representation into intermediate features suitable for model training and downstream future prediction tasks, we transform the processed state representation through a series of projections. Given the processed state representation $\mathbf{\acute{Z}}_t$ from the state graph, we first apply a linear projection layer $\phi_2$ to transform it into per-class embeddings $\mathbf{\breve{{Z}}}_t \in \mathbb{R}^{N_c \times d_2}$, where $N_c$ is the number of class labels and $d_2$ is the projection dimension. These embeddings represent verbs or procedural steps. We then apply a final linear layer $\phi_3$ to produce logits $\mathbf{\hat{Y}}_t \in \mathbb{R}^{N_c \times N_s}$, where $N_s$ is the number of state-change categories per class. The model predicts which state-change category applies to each class, ensuring that state-change learning is grounded in the video-derived state features. By pretraining on this objective, the model learns rich future context beneficial for downstream future prediction tasks.

\subsection{Naive Future Prediction: \textbf{SurgFUTR-Lite}}
\label{sec:surgfutr_lite}
\begin{figure}[!ht]
\centering
    \includegraphics[width=0.95\columnwidth]{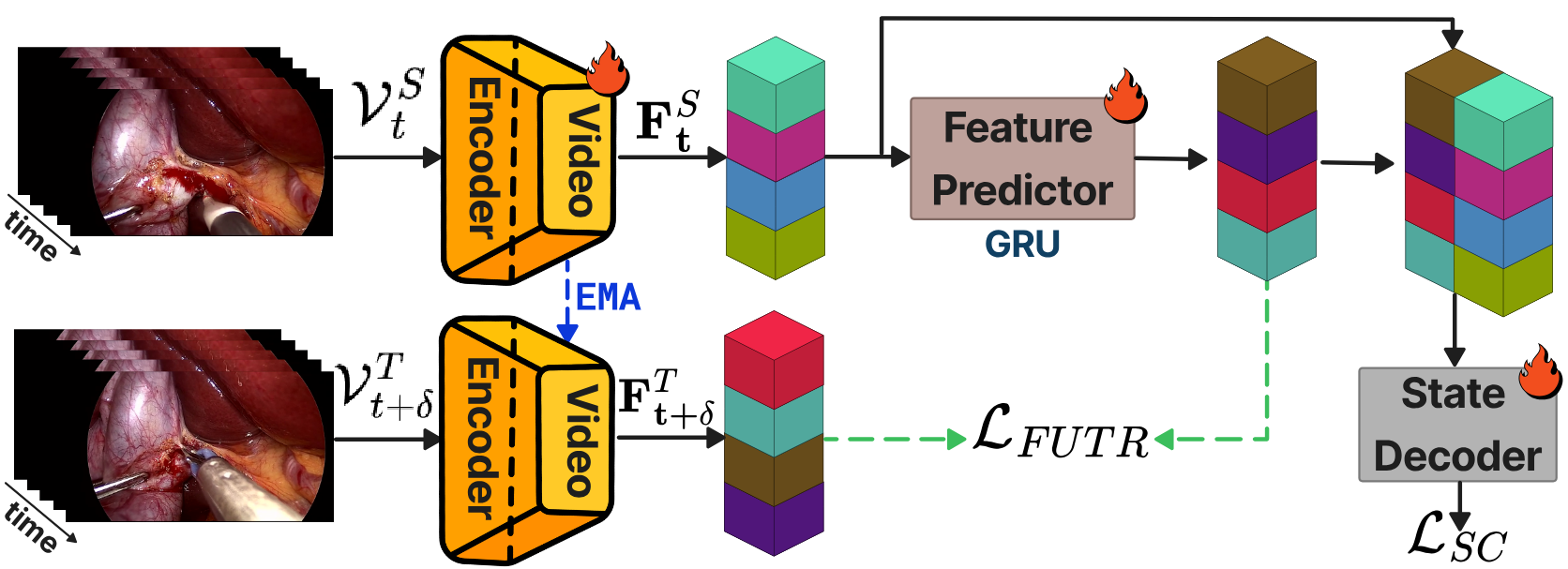}
    \caption{\textbf{SurgFUTR-Lite:} A single-stage future feature prediction model. The main encoder processes a clip at time $t$; the target encoder processes a clip at $t+\delta$, with the target backbone as an exponential moving average of the main encoder's. Spatially pooled features $\mathbf{F}_{t}$ are used for future feature prediction to output $\mathbf{\hat{F}}_{t+\delta}$, and are trained to match $\mathbf{F}_{t+\delta}$ via $\mathcal{L}_{FUTR}$. The concatenated features $[\mathbf{F}_{t},\mathbf{\hat{F}}_{t+\delta}]$ also drive a state-change head optimized with $\mathcal{L}_{SC}$.}
    \label{fig:surgfutr_lite}
\end{figure}

As illustrated in Figure~\ref{fig:task_intro}(a), we first establish a straightforward yet competitive baseline for future prediction that does \textit{not} rely on any explicit state representation. We refer to this model as \textbf{SurgFUTR-Lite}, depicted in Figure~\ref{fig:surgfutr_lite}.
In this approach, the current video clip $\mathbf{V}_t$ is passed through a video encoder to extract spatio-temporal features $\mathbf{F}_t \in \mathbb{R}^{THW \times d}$, where $T$, $H$, and $W$ represent temporal, height, and width dimensions, and $d$ is the feature dimension. These features are then average-pooled across the spatial dimension to yield a compact representation $\mathbf{F}_t \in \mathbb{R}^{T \times d}$.
To obtain the prediction target, the future video clip $\mathbf{V}_{t+\delta}$ is processed by an exponentially moving averaged (EMA) version of the video encoder, producing future features $\mathbf{F}_{t+\delta} \in \mathbb{R}^{T \times d}$. The model's objective is to predict these future features given only the current features $\mathbf{F}_t$.
We employ a temporal model (GRU) that autoregressively predicts future features from $t$ to $t+\delta$, trained with smooth L1 loss between predicted and actual features.
Finally, we concatenate the current features $\mathbf{F}_t$ and predicted future features $\mathbf{\hat{F}}_{t+\delta}$, and input them to a state decoder (see Section~\ref{sec:state_decoding}). The decoder outputs logits for each class label, supervised using cross-entropy loss to recognize state changes.
In summary, \textbf{SurgFUTR-Lite} provides a lightweight future prediction framework that bypasses explicit state representations and state transition graphs.

\subsection{Teacher-Student Distillation}
\label{sec:ts_distillation}
While SurgFUTR-Lite provides a strong baseline for state-change learning, it lacks explicit state representations capturing spatio-temporal dynamics. We address this through a teacher-student distillation strategy. The teacher model is trained with video clips from both time-steps $t$ and $t+\delta$ ($\mathbf{V}_t$ and $\mathbf{V}_{t+\delta}$), while the student model uses only single time-step inputs. Superscripts T and S distinguish teacher and student model components respectively.
The state vectors $\mathbf{Z}^T_t$ and $\mathbf{Z}^T_{t+\delta}$ are processed through the state graph (Section~\ref{sec:state_graph}) to obtain $\mathbf{\acute{Z}}^T_t$ and $\mathbf{\acute{Z}}^T_{t+\delta}$. These features are averaged to produce $\mathbf{\acute{Z}}_{t,t+\delta}$, which contains semantic information from both clips indicative of state changes. The averaged features $\mathbf{\acute{Z}}_{t,t+\delta}$ are input to the state decoder to output per-class embeddings $\mathbf{\breve{Z}}_{t,t+\delta}$, which are transformed to logits $\mathbf{\hat{Y}}_{t,t+\delta}$ and trained using cross-entropy loss:
\begin{equation}
\mathcal{L}_{SC} = -\frac{1}{N_c} \sum_{c=1}^{N_c} \sum_{s=1}^{N_s}
    y_{c, s} \log \left(
        \frac{\exp(\hat{y}_{c, s})}{\sum_{s'=1}^{N_s} \exp(\hat{y}_{c, s'})}
    \right),
\label{sc_loss}
\end{equation}
where $N_c$ is the number of classes, $N_s$ is the number of state-change categories, $y_{c, s}$ is the ground-truth indicator for class $c$ and state $s$, and $\hat{y}_{c, s}$ denotes the predicted logit for class $c$ and state $s$. This formulation encourages the teacher model to learn discriminative state representations sensitive to changes between current and future clips.

Once the teacher (Figure~\ref{fig:scadyn}) is trained, we distill its knowledge into a student model that only has access to the current video clip $\mathbf{V}_t$. The student learns to predict the future-aware state features generated by the teacher, anticipating state changes using only present information. We input a video clip $\mathbf{V}^{S}_t$ to the video encoder to obtain $\mathbf{F}^S_t$ spatio-temporal features, which are transformed to a state vector $\mathbf{Z}^S_t$. 
However, the student model cannot access future clips $\mathbf{V}_{t+\delta}$, preventing direct generation of future-aware state vectors.

Our solution relies on two key insights from teacher training.
First, each video clip produces a patch-to-centroid assignment matrix $\textbf{M}_t$ and $\textbf{M}_{t+\delta}$ for time steps $t$ and $t+\delta$. 
Second, these assignment matrices can enable modeling centroid evolution from $t$ to $t+\delta$ via transition matrices.
We derive the transition matrix through two steps:

(1) We compute a patch affinity matrix to capture correspondences between current and future clips. Using $\ell_2$-normalized features $\mathbf{F}^T_t, \mathbf{F}^T_{t+\delta} \in \mathbb{R}^{N \times d}$ (rows are patch embeddings), we form $\mathbf{A}^T_{patch} \in \mathbb{R}^{N \times N}$ as the matrix of pairwise dot products between rows of $\mathbf{F}^T_t$ and $\mathbf{F}^T_{t+\delta}$, yielding a soft correspondence across time.

(2) We construct the transition matrix by identifying top-$k$ most similar patches between current and future clips using the affinity matrix. For each current patch, we record transitions between centroid assignments of matched patches, weighted by normalized affinity scores. This produces a soft transition matrix $\mathcal{Q}^T_{t,t+\delta}$, capturing centroid evolution from $t$ to $t+\delta$. The complete algorithm is detailed in Algorithm~\ref{alg:transition_matrix}.

\begin{algorithm}[!htbp]
\caption{Create Centroid Transition Matrix}
\label{alg:transition_matrix}
\KwData{
    Current patch-to-cluster assignment $M^{T}_t \in \mathbb{R}^{N \times K}$; \\
    Future patch-to-cluster assignment $M^{T}_{t+\delta} \in \mathbb{R}^{N \times K}$; \\
    Patch affinity matrix $A^T_{patch} \in \mathbb{R}^{N \times N}$; \\
    Top-$k$ parameter $k$
}
\KwResult{Transition matrix $\mathcal{Q}^T_{t,t+\delta} \in \mathbb{R}^{K \times K}$}
Initialize $\mathcal{Q}^T_{t,t+\delta}$ as a zero matrix of shape $K \times K$\; 
Compute $c_t[i] \gets \arg\max_{k} M^{T}_t[i, k]$ for $i = 1, \ldots, N$\;
Compute $c_{t+\delta}[j] \gets \arg\max_{k} M^{T}_{t+\delta}[j, k]$ for $j = 1, \ldots, N$\;
\For{each patch $i = 1$ to $N$}{
    $topk\_vals \gets$ top-$k$ values from $A^T_{patch}[i, :]$\;
    $topk\_idx \gets$ indices of top-$k$ values from $A^T_{patch}[i, :]$\;
    $topk\_vals \gets$ softmax$(topk\_vals)$\;
    \For{each $j = 1$ to $k$}{
        $src \gets c_t[i]$\;
        $tgt \gets c_{t+\delta}[topk\_idx[j]]$\;
        $\mathcal{Q}^T_{t,t+\delta}[src, tgt] \mathrel{+}= topk\_vals[j]$\;
    }
}
Normalize rows: $\mathcal{Q}^T_{t,t+\delta}[p, :] \mathrel{/}= \sum_q \mathcal{Q}^T_{t,t+\delta}[p, q]$
\end{algorithm}

\begin{figure}[!ht]
\centering
    \includegraphics[width=0.60\columnwidth]{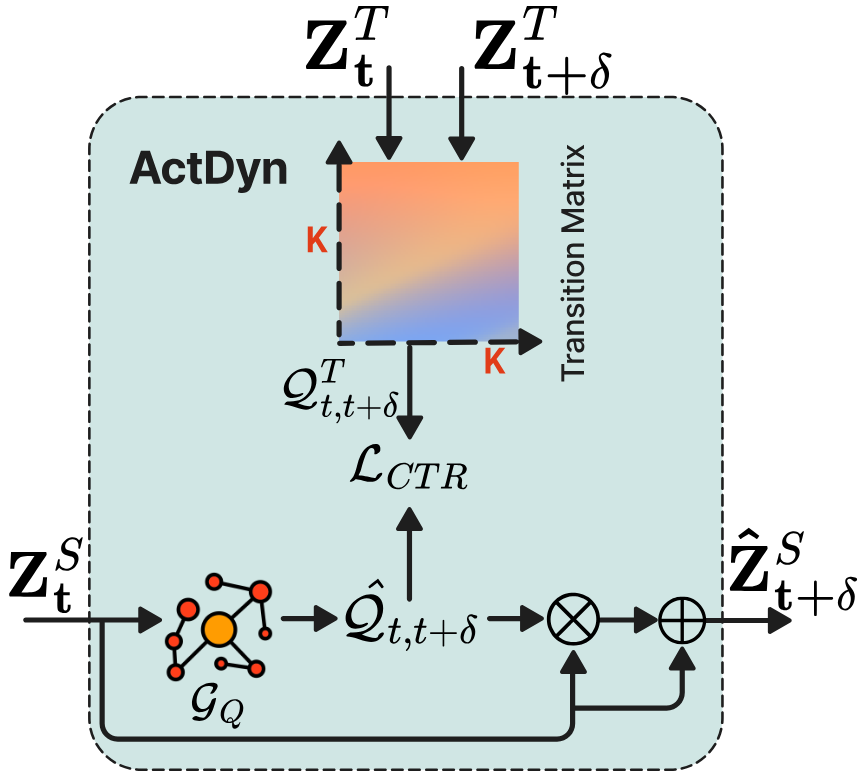} 
    \caption{Action dynamics modeling with centroid transition prediction}
    \label{fig:actdyn}
\end{figure}

After obtaining the centroid transition matrix $\mathcal{Q}^T_{t,t+\delta}$, the student model's state vector $\mathbf{Z}^S_t$ should learn to predict transitions from state at time step $t$ to future state at time step ${t+\delta}$. We propose an action dynamics module (\textbf{ActDyn}), shown in green in Figure~\ref{fig:scadyn}, which uses a 1-layer GATv2~\citep{brody2021attentive} network to generate the predicted transition matrix $\mathcal{\hat{Q}}_{t,t+\delta} \in \mathbb{R}^{K \times K}$ from current state features $\mathbf{Z}^S_t$. We normalize $\mathcal{\hat{Q}}_{t,t+\delta}$ using:
\begin{equation}
\mathcal{\hat{Q}}_{t,t+\delta} = \frac{ \mathcal{\hat{Q}}_{t, t+\delta}}{\sum_{k'=1}^{K}  \mathcal{\hat{Q}}_{t, t+\delta}^{(:,k')}}
\end{equation}
To ensure proper learning, we use Wasserstein (Earth Mover's Distance, EMD) loss~\citep{feydy2019interpolating}. 
Since transition matrices represent mass transport between centroids over time, Wasserstein distance naturally measures the minimal transformation cost between distributions while considering centroid geometry. This enables learning of transition matrices that accurately capture centroid dynamics.
\begin{equation}
\mathcal{L}_{CTR} = \sum_{i=1}^{K} \mathrm{EMD}\left( \mathcal{\hat{Q}}_{t, t+\delta}[i, :],\ \mathcal{Q}^T_{t, t+\delta}[i, :] \right),
\end{equation}
where, $i$ indexes over the number of centroids $K$. 
This forces the student to learn patterns in current spatio-temporal features that predict future centroid evolution, effectively learning to anticipate state changes from present observations alone.
Finally, we obtain the predicted future centroids for time $t+\delta$ using:
\begin{equation}
\mathbf{\hat{Z}}^S_{t+\delta} = \left( \mathcal{\hat{Q}}_{t, t+\delta} + \alpha \mathbf{I} \right) \mathbf{Z}^S_t,
\label{ctr_update}
\end{equation}
where $\mathbf{I}$ is the identity matrix. This update rule computes each future centroid as a combination of propagated current centroids (via the learned transition matrix) and a direct contribution from its previous state. The parameter $\alpha$ controls the balance between following predicted transitions and retaining original centroid positions, providing flexibility and stability in temporal evolution modeling.
Figure~\ref{fig:actdyn} deconstructs the steps to obtain predicted centroids at time step $t+\delta$.
We apply smooth-L1 loss between the teacher's learned state $\mathbf{Z}^T_{t+\delta}$ and the ActDyn module's predicted state $\mathbf{\hat{Z}}^S_{t+\delta}$:
\begin{equation}
    \mathcal{L}_{FCTR} = \text{smooth-L1}(\mathbf{Z}^T_{t+\delta}, \mathbf{\hat{Z}}^S_{t+\delta}, \delta)
\end{equation}
where $\delta$ is a threshold hyperparameter. Following SurgFUTR-Lite, we also add an auxiliary feature predictor that predicts spatially pooled features $\mathbf{\hat{F}}^S_{t+\delta}$ from $\mathbf{F}^S_{t}$ and apply smooth-L1 loss:
\begin{equation}
    \mathcal{L}_{FUTR} = \text{smooth-L1}(\mathbf{F}^T_{t+\delta}, \mathbf{\hat{F}}^S_{t+\delta}, \delta)
\end{equation}
where $\delta$ is a threshold hyperparameter. 
The student model averages its current state vector $\mathbf{Z}^S_t$ and predicted future state vector $\mathbf{\hat{Z}}^S_{t+\delta}$ to form $\mathbf{\hat{Z}}^S_{t,t+\delta}$. This representation flows through the state graph and through state decoder to produce per-class embeddings $\mathbf{\breve{Z}}^S_{t,t+\delta}$ and final state-change logits $\mathbf{\hat{Y}}^S_{t,t+\delta}$. Training occurs with the teacher frozen.

\subsection{Distillation Losses}
In our teacher-student framework, we employ a dual-level distillation strategy to transfer knowledge from the teacher to the student model. 
The intuition is to leverage the teacher's spatio-temporal knowledge from current and future clips to refine the student's features.
At the first level, we apply a distillation loss to align class embeddings $\mathbf{\breve{Z}}_{t,t+\delta}$ by matching the student's state representation to the teacher's corresponding embedding:
\begin{equation}
\mathcal{L}_{SE}^{distill} = {||\mathbf{\breve{Z}}^S_{t,t+\delta} - \mathbf{\breve{Z}}^T_{t,t+\delta}||}_1,
\end{equation}
where SE denotes distilled state embedding.
In the second level, we distill the state-change context logits from the teacher to the student using:
\begin{equation}
\mathcal{L}_{SC}^{distill} = \left|\left|\text{softmax}\left(\frac{\mathbf{\hat{Y}}^S_{t,t+\delta}}{\tau_2}\right) - \text{softmax}\left(\frac{\mathbf{Y}^T_{t,t+\delta}}{\tau_2}\right)\right|\right|_1
\label{loss_sc_distill}
\end{equation}
where $\tau_2$ represents temperature. Our comprehensive training objective for \textbf{SurgFUTR-TS} combines centroid-based learning with state-change aware distillation:
\begin{align}
\mathcal{L}_{centroid} &= \lambda_1 \mathcal{L}_{SC} +
\lambda_2 \mathcal{L}_{CTR} +
\lambda_3 \mathcal{L}_{FCTR}, \\
\mathcal{L}_{distill} &= \lambda_4 \mathcal{L}_{SE}^{distill} + \lambda_5 \mathcal{L}_{SC}^{distill}, \\
\mathcal{L}_{total} &= \mathcal{L}_{centroid} +
\lambda_6 \mathcal{L}_{FUTR} + \mathcal{L}_{distill},
\end{align}
where $\{ \lambda_i \mid i \in [1, 6] \}$ are fixed loss coefficients that balance the contribution of each component in our unified training framework. This teacher-student distillation framework enables the student to benefit from the richer supervision available to the teacher, ultimately leading to improved state-change recognition performance by only relying on the current video clip.

  \section{Experiments}

\subsection{Dataset Overview}
\paragraph{\textbf{CholecT50}} 
The public dataset~\citep{nwoye2021rendezvous} consists of 50 laparoscopic cholecystectomy videos sampled at 1 fps, annotated with multi-label surgical action triplets comprising 6 instruments, 10 verbs, and 15 target anatomical structures across 7 procedural phases. We follow the RDV splits~\citep{nwoye2022data}: 35 training, 5 validation, and 10 test videos. The videos correspond to raw video recordings from the Cholec80 dataset~\citep{twinanda2016endonet}. We use this dataset at two stages: (1) state-change pretraining with verb annotations, and (2) evaluation of downstream tasks.

\paragraph{\textbf{GraSP}} 
The public dataset~\citep{ayobi2025pixel} contains 13 robot-assisted radical prostatectomy videos with annotations for 11 phases, 21 procedural steps, surgical actions, instrument presence, and segmentation masks. The dataset provides Fold1 and Fold2 for training and a separate test set. We combine both folds for training and reserve 2 videos for validation (6 training, 5 test videos). Videos are available as pre-extracted frames at 1 fps. We use this dataset for both state-change pretraining (using step annotations) and downstream evaluation tasks.

\paragraph{\textbf{CholecTrack20}} 
The public dataset~\citep{nwoye2023cholectrack20} provides 20 laparoscopic cholecystectomy videos with tool-tracking annotations and surgical event labels including bleeding, smoke, occlusion, and lens fudging. The dataset splits into 10 training, 2 validation, and 8 test videos. We use this dataset exclusively for downstream evaluation of surgical event anticipation tasks. Since some videos in CholecTrack20 also appear in CholecT50, we remove these overlapping videos from all splits (train, validation, and test) to prevent data leakage.

\paragraph{\textbf{MultiBypass140}} This public dataset~\citep{Lavanchy2024} consists of 140 laparoscopic Roux-en-Y gastric bypass (LRYGB) surgery videos with multi-centric phase and step annotations from two medical centers: University Hospital of Strasbourg, France (Strasbourg) and Inselspital, Bern University Hospital, Switzerland (Bern). Each center contributes 40 training videos, 10 validation videos, and 20 test videos. The dataset contains 12 phases and 46 steps, making it a challenging benchmark for evaluating cross-procedure transfer. We use this dataset to evaluate how state-change representations pretrained on laparoscopic cholecystectomy transfer to gastric bypass procedures through fine-tuning. While the procedures share common instruments (\textit{grasper}, \textit{hook}), actions (\textit{dissect}, \textit{grasp}), and some anatomical targets (\textit{omentum}, \textit{abdominal wall cavity}), gastric bypass introduces new anatomical structures (\textit{stomach}, \textit{small-bowel}, \textit{colon}, \textit{spleen}) and new instruments (\textit{stapler}, \textit{needle-driver}). We use fold 0 across both centers for training and evaluation, reporting results across multiple random seeds for downstream future prediction tasks.

\begin{table*}[!htp]
    \centering
    \caption{Data Distribution for State Change Experiments on CholecT50 and GraSP datasets.}
    \label{tab:sc_data_dist}
    \setlength{\tabcolsep}{15pt}
    \resizebox{\textwidth}{!}{%
    \begin{tabular}{lcccccccccc}
    \toprule
        \multirow{2}{*}{\textbf{Split}} &
        \multicolumn{4}{c}{\textbf{CholecT50}} &
        \multicolumn{4}{c}{\textbf{GraSP}} \\
        \cmidrule(lr){2-5} 
        \cmidrule(lr){6-9} 
        & $\textbf{Continuity}$ & $\textbf{Discontinuity}$ & $\textbf{Onset}$ & $\textbf{Background}$
        & $\textbf{Continuity}$ & $\textbf{Discontinuity}$ & $\textbf{Onset}$ & $\textbf{Background}$ \\
    \midrule
        Train & $7851$ & $2255$ & $2380$ & $8883$ & $2045$ & $1150$ & $1147$ & $3191$ \\
        Val & $733$ & $234$ & $255$ & $847$ & $250$ & $178$ & $177$ & $427$ \\
        Test & $2363$ & $654$ & $691$ & $2613$ & $1336$ & $684$ & $679$ & $2015$ \\
    \bottomrule
    \end{tabular}
    }
\end{table*}

\subsection{Video Clip Sampling Algorithm}
To train our state-change classification model, we extract video clips from fine-grained datasets using Algorithm~\ref{alg:state_change_clips}. This procedure works with datasets containing fine-grained supervised annotations, such as action triplets in CholecT50~\citep{nwoye2021rendezvous} or step labels in GraSP~\citep{ayobi2025pixel}. The frame-wise label matrix $L_v$ represents one-hot encoded class labels for each frame. 

For CholecT50, we extract video clips directly from raw Cholec80 videos at 25fps using timestamps generated in step~4 of Algorithm~\ref{alg:state_change_clips}. We set clip duration $d=3$, sampling interval $t_s=10$, and buffer $t_b=100$ to learn state-change embeddings for the 10 verb classes. For GraSP, we form video clips by collecting pre-extracted frames (1fps) specified by the generated timestamps. We use parameters $d=4$, $t_s=90$, and $t_b=50$ to capture state transitions across the 21 step classes. 

To reduce label jitter during timestamp selection, we use a stability frame window of $w_f=3$ on the 1 fps annotation timeline. For onset and discontinuity, a candidate timestamp is retained only when the source label remains stable for the preceding $w_f$ frames and the destination label remains stable for the subsequent $w_f$ frames. For continuity and background, timestamps are sampled only from temporally stable regions that lie at least $w_f$ frames away from any onset or discontinuity. This filtering is particularly important for CholecT50, where verb labels are fine-grained and short-lived fluctuations may otherwise introduce spurious transitions. 
For example, an onset of the verb \emph{cut} at timestamp $t$ is accepted only if \emph{cut} is absent for the preceding $w_f$ frames and present for the following $w_f$ frames on the 1 fps label timeline.


In contrast, GraSP step labels are substantially coarser and more temporally stable, so label jitter is much less pronounced in that setting. If a candidate timestamp does not satisfy the stability condition, or if multiple label changes occur within the same local neighborhood, it is discarded rather than assigned an ambiguous target. This ensures that each constructed state-change clip pair is anchored to a temporally stable and well-defined transition event.

For transparency, we additionally report the effect of this timestamp-selection procedure. On CholecT50, the stability and overlap filtering retains 3,303 onset timestamps and 1,705 discontinuity timestamps, while discarding 281 onset candidates and 69 discontinuity candidates that otherwise satisfy the stability criterion but fall within the exclusion window of an already selected transition. These intermediate timestamp-level statistics are not expected to match the final per-split clip counts, since the latter are obtained only after clip construction, validity checks, temporal aggregation, and split assignment. For GraSP, timestamp-level transition counts are substantially more asymmetric due to the sparse step-event encoding, particularly for discontinuity events; therefore, we report the final clip-level onset, discontinuity, and continuity counts for that dataset, as these more directly reflect the supervision used for training and evaluation.

Our approach seamlessly handles both 25fps raw videos and pre-extracted frame formats for state-change pretraining. The resulting clip distributions across training, validation, and test splits are detailed in Table~\ref{tab:sc_data_dist}.
Since state-change labels are assigned independently for each class and then aggregated within a clip, the reported categories are not mutually exclusive. In particular, \textit{Background} denotes clips containing at least one class with no state change ($0 \rightarrow 0$), rather than clips that are entirely background-only; consequently, a single clip may contribute to both the background count and one or more transition categories.

\subsection{SFPBench: Surgical Future Prediction Benchmark}
\label{task_desc}
We construct \textbf{SFPBench}, a comprehensive surgical future prediction benchmark that addresses the lack of unified evaluation frameworks for surgical anticipation tasks. Our benchmark covers both short-term anticipation (seconds) and long-term forecasting (minutes or end of video) across diverse surgical scenarios, as shown in Figure~\ref{fig:task_overview}. Long-term tasks utilize the same clip sampling methodology described above, while short-term tasks require custom data construction approaches specific to each task. SFPBench unifies evaluation across multiple temporal scales and prediction types, enabling systematic comparison of anticipation methods. We describe each anticipation task and its data construction process.

\begin{algorithm}[ht]
\caption{State Change Clip Sampling}\label{alg:state_change_clips}
\KwData{Frame-wise label matrix $L_v$ for each video $v$; stability frame window $w_f$; stride $t_s$ for label 1; stride $t_b$ for label 0; clip duration $d$}
\KwResult{Clips and metadata for state change learning}
\ForEach{video $v$}{
    \tcp{1. Sample onset and discontinuity}
    \ForEach{frame timestamp $t$ and class $c$}{
        \textbf{Onset (0→1):} if $L_v[t-w_f:t-1,c]=0$ and $L_v[t:t+w_f-1,c]=1$, mark $t$ as onset.\;
        \textbf{Discontinuity (1→0):} if $L_v[t-w_f:t-1,c]=1$ and $L_v[t:t+w_f-1,c]=0$, mark $t$ as discontinuity.\;
        Exclude $t$ if within $w_f$ frames of another timestamps.\;
    }
    
    \tcp{2. Sample continuity and background}
    \For{label $x \in \{1,0\}$}{
        Use stride $t_s$ if $x=1$ (continuity), $t_b$ if $x=0$ (background).\;
        \textbf{Continuity (1→1):} In each window, find timestamp $t$ where $L_v[t-w_f:t+w_f,c]=1$ for any $c$, and $t$ is not within $w_f$ of any transition.\;
        \textbf{Background (0→0):} In each window, find timestamp $t$ where $L_v[t-w_f:t+w_f,c]=0$ for any $c$, and $t$ is not within $w_f$ of any transition.\;
    }

    \tcp{3. Merge and deduplicate}
    Merge timestamps, removing overlaps $\le w_f$ frames.\;

    \tcp{4. Generate clips}
    \ForEach{timestamp $t$}{
        Current clip: $[t-d-1, t-1]$.\;
        Future clip: $[t, t+d]$.\;
        Save valid, non-overlapping clip metadata.\;
    }
}
\end{algorithm}

\paragraph{\textbf{SFP-I: Remaining Surgery Duration (RSD) Prediction}} 
Accurate prediction of remaining surgery duration is fundamental to perioperative care, enabling optimal operating room utilization, precise anesthesia management, and improved workflow planning. This critical long-term forecasting challenge was first addressed by RSDNet~\citep{twinanda2018rsdnet}. To evaluate whether state-change representations improve temporal reasoning for surgery duration estimation, we apply our pretrained models to RSD prediction using the same video clips from state-change pretraining on CholecT50 and GraSP datasets.

To assess cross-procedure transfer learning, we additionally evaluate cholecystectomy-pretrained models fine-tuned for RSD prediction on MultiBypass140~\citep{Lavanchy2024}. We create 8-frame clips from consecutive frames using a sliding window with stride 20: starting from frame 0, each clip contains 8 consecutive frames, and the next clip starts 20 frames later. This yields 11,899/3,433/6,521 clips (train/val/test) for Strasbourg and 7,555/2,207/4,308 for Bern across fold 0. We follow RSDNet's regression framework, normalizing RSD values at each timestep $t$ to the range $[0, 20]$.

\paragraph{\textbf{SFP-II: Phase Transition Prediction}} 
This task predicts upcoming phase transitions within the next 2, 3, or 5 minutes, enabling proactive workflow management and timely preparation for procedural changes. We use the same video clips from state-change pretraining on CholecT50 and GraSP datasets. For cross-procedure transfer learning on MultiBypass140, we use the same 8-frame clips and preprocessing described in the RSD task. Following established protocols~\citep{rivoir2020rethinking,yuan2022anticipation}, we assign label 0 to clips where the current phase continues, and for other clips, the label represents the remaining time in minutes until the next transition, truncated at $p$ minutes. Ground truth values range within $[0, p]$, where 0 indicates an ongoing phase and $p$ indicates no transition within $p$ minutes.

\paragraph{\textbf{SFP-III: Step Transition Prediction}} 
This task extends phase transition prediction to finer-grained procedural steps, enabling more precise workflow anticipation. We apply our approach to step transitions using the same video clips from state-change pretraining on GraSP dataset. For cross-procedure transfer learning on MultiBypass140, we use the same 8-frame clips and preprocessing described in the RSD task. Label construction follows the same methodology as phase transition prediction.

\paragraph{\textbf{SFP-IV: Cystic Triplet Anticipation}} 
Anticipating surgical action triplets involving critical anatomical structures—\textit{cystic duct}, \textit{cystic artery}, \textit{cystic plate}, and \textit{cystic pedicle}—is crucial for surgical assistance systems. These triplets represent high-risk interactions during \textit{calot triangle dissection} and \textit{clipping and cutting} phases where surgical precision directly impacts patient safety. Predicting these critical actions before they occur enables real-time guidance and early warning systems to improve surgical outcomes.

\begin{table}[ht]
\centering
\begin{tabular}{|l|l|}
\hline
\textbf{\textlangle{}instrument, verb, target}\textrangle{} & \textbf{\textlangle{}instrument, verb, target}\textrangle{} \\
\hline
hook, dissect, cystic-duct & clipper, clip, cystic-duct \\
hook, dissect, cystic-plate & clipper, clip, cystic-artery \\
hook, dissect, cystic-artery & scissors, cut, cystic-duct \\
scissors, cut, cystic-artery & \\
\hline
\end{tabular}
\caption{List of surgical action triplet classes that interact directly with critical anatomical structures - \textit{cystic-duct}, \textit{cystic-artery}, \textit{cystic-plate} with deformable actions such as \textit{dissect}, \textit{clip}, and \textit{cut}.}
\label{tab:cystic_triplets}
\end{table}

We construct this short-term anticipation task by sampling 3-second video clips from CholecT50~\citep{nwoye2021rendezvous} at 1-sec, 3-sec, and 5-sec anticipation horizons before each cystic-structure-related triplet occurrence (Table~\ref{tab:cystic_triplets}).
This multi-horizon sampling approach is inspired by the EPIC-KITCHENS~\citep{Damen2021PAMI} anticipation framework. Given a video clip, the model predicts which surgical action triplet will occur next. Our dataset comprises 1,328 training, 192 validation, and 484 test video clips.

\paragraph{\textbf{SFP-V: Cholec Event Anticipation}} 
This task focuses on anticipating surgical events (e.g., bleeding) and visual challenges (e.g., smoke, occlusion) from CholecTrack20~\citep{nwoye2023cholectrack20}. These events are critical intraoperative complications that require timely surgical intervention.
We construct this short-term anticipation task by sampling 3-second video clips at anticipation horizons of 1-sec, 3-sec, and 5-sec before event occurrence.
Our dataset comprises 1,094 training, 218 validation, and 521 test video clips.

\subsection{Data Setup}
We standardize all video frames to a $224 \times 224$ spatial resolution regardless of their original dimensions. For training, both teacher and student models receive identical RandAugment transformations with magnitude $7$ and $4$ layers applied to input images. When working with raw video clips, we use the \textit{decord} library for on-the-fly frame extraction at $25$ fps during training, using uniform sampling with fixed intervals which is set to $8$ for video clip of length $8$. 

\subsection{Implementation Details}
\label{impl_details}
We implement SurgFUTR in PyTorch and use Vision-Transformer~\citep{dosovitskiy2020image} (ViT-S) small variant as our primary model $\phi_v$. 
\paragraph{\textbf{Backbone}} We use a ViT-S model pretrained using video masked autoencoding setup VideoMAEv2~\citep{wang2023videomaev2}. The model weights are obtained from distillation from ViT-giant to ViT-small, provided by the mmaction2~\citep{2020mmaction2} framework. For our experiments, we set the clip length $T = 8$, with spatial dimensions $H = W = 224$ and patch size $P = 16$. The VideoMAEv2 pretrained ViT uses a default tubelet size of $2$, which groups adjacent spatio-temporal tubes of shape $P \times P$ into single tokens. This configuration results in $784$ total spatio-temporal tokens for a video of length $T$, with the ViT-S backbone producing features of dimension $d = 384$.
\paragraph{\textbf{Clustering}} We set the number of centroids $K=25$ by default. The number of sinkhorn iterations $n_i$ is set to $3$ as in CrOC~\citep{stegmuller2023croc}. The temperature values for marginals is set to $1$ and the $\lambda$ in Equation~\ref{sinkhorn} is set to $1$.
\paragraph{\textbf{State Graph}} The projection layer $\phi_1$ is a single-layer MLP mapping features from $d=384$ to the $d_1=256$. During state graph construction, the temperature $\tau_1$ and similarity threshold $\theta$ parameters are set to $0.05$ and $0.02$ respectively. 
We use 4 attention heads in the single GATv2 layer of the state graph.
\paragraph{\textbf{State Decoder}} The feature dimension $d_2$ is set to $128$ using a single-layer MLP $\phi_2$. A second single-layer MLP $\phi_3$ then converts state features to per-class embeddings $\mathbf{s}^{+}_t \in \mathbb{R}^{N_c \times d_2}$, where $N_c$ is the number of classes: $N_c = 10$ for verb classes in CholecT50 and $N_c = 21$ for step classes in GraSP. Each class is then associated with $N_s = 4$ state-change labels: continuity, discontinuity, onset, and background.

For CholecT50~\citep{nwoye2021rendezvous}, we extract $3$-second video clips and predict states $\delta = 3$ seconds into the future. The momentum parameter for EMA in SurgFUTR-Lite is set to $0.004$.
For GraSP~\citep{ayobi2025pixel}, we construct video clips from pre-extracted frames using $T=4$ frames with the same prediction horizon of $\delta = 3$ seconds. We use $T=4$ frames to maintain temporal consistency with CholecT50's $3$-second clips, accounting for VideoMAEv2's internal temporal downsampling by a factor of two.

\paragraph{\textbf{State-Change Pretraining}} 
\label{sc_pretraining}
To construct the centroid transition matrix based on Algorithm~\ref{alg:transition_matrix}, we set $k = 3$ for the top-$k$ selection. 
The graph module $\mathcal{G}_Q$ in ActDyn module consists of a single GATv2-based graph layer with 4 attention heads and ELU activation, followed by layer normalization and a linear projection.
For computing the Wasserstein distance (Earth Mover's Distance), we use the \textit{geomloss}\footnote{\url{https://www.kernel-operations.io/geomloss}}~\citep{feydy2019interpolating} library with parameters $p = 2$ and $blur = 0.01$. The temperature $\tau_2$ in Equation~\ref{loss_sc_distill} is set to $0.3$. For loss weighting, we set $\lambda_1 = 1$, $\lambda_2 = 1$, $\lambda_3 = 0.5$ for $\mathcal{L}_{centroid}$; $\lambda_4 = 1$, $\lambda_5 = 1$ for $\mathcal{L}_{distill}$; and $\lambda_6 = 0.7$ for the future feature prediction loss $\mathcal{L}_{FUTR}$.

We train SurgFUTR for $40$ epochs with batch sizes of $32$ for the teacher model and $16$ for teacher-student distillation. 
The base learning rate is $1 \times 10^{-4}$. In the first 5 epochs, the learning rate is gradually increased from $1 \times 10^{-7}$ to the base learning rate, after which cosine annealing is applied with a minimum learning rate of $1 \times 10^{-6}$.
We employ AdamW optimizer with weight decay $0.05$ and apply gradient clipping with maximum norm $1.0$. The teacher model is trained on video clips at timesteps $t$ and $t+\delta$ for $40$ epochs. During distillation, we freeze the teacher and train only the student model with the ActDyn module. Training is conducted in mixed-precision using the mmaction2 framework on up to $2$ NVIDIA A40 or A100 GPUs. Hyperparameters are tuned on the validation set, and experiments are run with multiple random seeds.

\paragraph{\textbf{SFPBench}} 
For RSD (SFP-I), phase (SFP-II), and step (SFP-III) transition prediction, we use the same training settings as state-change pretraining except for the learning rate, which is set to $3 \times 10^{-4}$.
For cystic-triplet anticipation (SFP-IV), we use the same optimizer configuration as RSD but adjust the cosine annealing minimum learning rate from $1 \times 10^{-8}$ to $1 \times 10^{-6}$ and set the initial learning rate to $2 \times 10^{-4}$.
For event anticipation (SFP-V) on CholecTrack20~\cite{nwoye2023cholectrack20}, we train for 20 epochs with 1 warmup epoch followed by cosine annealing (minimum learning rate $1 \times 10^{-7}$, initial learning rate $2 \times 10^{-4}$).
To address class imbalance across all classification tasks, we apply inverse frequency class weighting.

\subsection{Evaluation metrics}
\paragraph{\textbf{State-Change Pretraining}}
We evaluate state-change recognition using mean average precision (mAP) and F1-score across all test clips. mAP assesses ranking quality across precision-recall thresholds, while F1-score measures recognition performance at a fixed threshold. We report performance for three states: continuity, discontinuity, and onset. The background state is excluded from reporting as it consistently achieves $>95\%$ performance due to its high prevalence in the data. Final results report the mean performance averaged over these three states.

\paragraph{\textbf{SFPBench}} 
We use task-appropriate metrics following established practices. For regression-based long-term forecasting tasks (remaining surgery duration, phase and step transitions), we report mean absolute error (MAE in minutes) following prior work~\citep{twinanda2018rsdnet,rivoir2020rethinking,yuan2022anticipation}. For classification-based short-term anticipation tasks, we report mAP, F1-score, and accuracy for comprehensive performance assessment.

        

\section{Results}
\begin{table*}[h]
    \centering
    \caption{Model performance comparison on CholecT50 for verb state-change recognition. Results are reported per state (continuity, discontinuity, onset) and their mean. Entries are mean ± standard deviation across $3$ random runs. Higher values indicate better performance.}
    \label{tab:results:quantitative:sc_verb_cholect50}
    \setlength{\tabcolsep}{6pt}
    \resizebox{\textwidth}{!}{%
    \begin{tabular}{lllcccccccc}
    \toprule
        \multirow{2}{*}{\textbf{Method}} &
        \multirow{2}{*}{\textbf{Backbone}} &
        \multirow{2}{*}{\textbf{Initialization}} &
        \multicolumn{4}{c}{\textbf{mAP (\%)}} &
        \multicolumn{4}{c}{\textbf{F1-Score (\%)}} \\
        \cmidrule(lr){4-7} 
        \cmidrule(lr){8-11} 
        & & & $\textbf{Continuity}$ & $\textbf{Discontinuity}$ & $\textbf{Onset}$ & $\textbf{mean}$ &
        $\textbf{Continuity}$ & $\textbf{Discontinuity}$ & $\textbf{Onset}$ & $\textbf{mean}$ \\
        \midrule
        VideoMAEv2 & ViT-S & Random & $63.1\pm\scriptstyle{0.2}$ & $7.8\pm\scriptstyle{0.3}$ & 
        $8.9\pm\scriptstyle{0.3}$ & 
        $26.6\pm\scriptstyle{0.2}$ &
        $62.7\pm\scriptstyle{0.8}$ & $11.7\pm\scriptstyle{1.0}$ &
        $14.2\pm\scriptstyle{0.5}$ & $29.5\pm\scriptstyle{0.4}$ \\
        VideoMAEv2 & ViT-S & Kinetics-400 & $75.2 \pm \scriptstyle{0.8}$ & $13.5 \pm \scriptstyle{0.4}$ &
        $14.7 \pm \scriptstyle{0.2}$ & 
        $34.4 \pm \scriptstyle{0.1}$ & 
        $69.5 \pm \scriptstyle{1.6}$ & $21.1 \pm \scriptstyle{0.6}$ &
        $20.2 \pm \scriptstyle{0.8}$ & $36.9 \pm \scriptstyle{0.5}$ \\
        VideoMAEv2 & ViT-S & Phase & $74.9 \pm \scriptstyle{1.2}$ & $13.4 \pm \scriptstyle{0.9}$ &
        $15.0 \pm \scriptstyle{0.1}$ & 
        $34.4 \pm \scriptstyle{0.1}$ & 
        $68.1 \pm \scriptstyle{0.0}$ & $21.6 \pm \scriptstyle{0.2}$ &
        $20.8 \pm \scriptstyle{0.2}$ & $36.9 \pm \scriptstyle{0.1}$ \\
        MoCoV2 & ResNet50 & Cholec80 & $74.1 \pm \scriptstyle{1.5}$ & $11.9 \pm \scriptstyle{0.7}$ &
        $13.7 \pm \scriptstyle{0.2}$ & 
        $33.2 \pm \scriptstyle{0.3}$ & 
        $68.9 \pm \scriptstyle{2.7}$ & $18.0 \pm \scriptstyle{1.0}$ &
        $21.6 \pm \scriptstyle{1.2}$ & $36.2 \pm \scriptstyle{0.9}$ \\
        DINO & ResNet50 & Cholec80 & $74.8 \pm \scriptstyle{0.6}$ & $10.9 \pm \scriptstyle{0.7}$ &
        $13.3 \pm \scriptstyle{0.5}$ & 
        $33.0 \pm \scriptstyle{0.1}$ & 
        $70.3 \pm \scriptstyle{2.1}$ & $17.0 \pm \scriptstyle{1.0}$ &
        $21.3 \pm \scriptstyle{0.3}$ & $36.2 \pm \scriptstyle{0.6}$ \\
        EndoViT & ViT-B & Endo700k & $59.7 \pm \scriptstyle{1.8}$ & $5.7 \pm \scriptstyle{0.3}$ &
        $7.0 \pm \scriptstyle{0.1}$ & 
        $24.1 \pm \scriptstyle{0.6}$ & 
        $61.8 \pm \scriptstyle{0.5}$ & $8.5 \pm \scriptstyle{0.4}$ &
        $12.1 \pm \scriptstyle{0.3}$ & $27.5 \pm \scriptstyle{0.3}$ \\
        SurgeNetXL & CaFormerS18 & SurgeNetXL & $74.9 \pm \scriptstyle{0.8}$ & $11.2 \pm \scriptstyle{1.1}$ &
        $12.8 \pm \scriptstyle{0.2}$ & 
        $33.0 \pm \scriptstyle{0.2}$ & 
        $69.5 \pm \scriptstyle{1.2}$ & $17.4 \pm \scriptstyle{1.7}$ &
        $20.9 \pm \scriptstyle{1.0}$ & $35.9 \pm \scriptstyle{0.7}$ \\
        \midrule
        SurgFUTR-Lite & ViT-S & Kinetics-400 & 
        $76.1 \pm \scriptstyle{1.0}$ & 
        $14.1 \pm \scriptstyle{0.7}$ &
        $16.5 \pm \scriptstyle{0.8}$ & 
        $35.6 \pm \scriptstyle{0.7}$ & 
        $68.8 \pm \scriptstyle{2.8}$ & 
        $20.1 \pm \scriptstyle{0.6}$ &
        $21.3 \pm \scriptstyle{1.4}$ & 
        $36.7 \pm \scriptstyle{1.2}$ \\
        SurgFUTR-S & ViT-S & Kinetics-400 & $76.1 \pm \scriptstyle{1.0}$ & $14.3 \pm \scriptstyle{0.6}$ &
        $14.9 \pm \scriptstyle{1.2}$ & 
        $35.1 \pm \scriptstyle{0.6}$ & 
        $69.2 \pm \scriptstyle{0.6}$ & $20.6 \pm \scriptstyle{0.2}$ &
        $20.0 \pm \scriptstyle{0.5}$ & $36.6 \pm \scriptstyle{0.3}$ \\
        \rowcolor{cyan!15}
        SurgFUTR-TS & ViT-S & Kinetics-400 & $\textbf{77.8} \pm \scriptstyle{1.0}$ & $\textbf{14.9} \pm \scriptstyle{0.8}$ &
        $\textbf{16.6} \pm \scriptstyle{0.2}$ & 
        $\textbf{36.4} \pm \scriptstyle{0.5}$ & 
        $\textbf{71.4} \pm \scriptstyle{1.7}$ & $\textbf{22.3} \pm \scriptstyle{0.4}$ &
        $\textbf{22.4} \pm \scriptstyle{0.1}$ & $\textbf{38.7} \pm \scriptstyle{0.5}$ \\
        
    \bottomrule
    \end{tabular}
    }
\end{table*}

In this section, we evaluate the transfer learning effectiveness of our state-change pretraining approach. Our central hypothesis is that models pretrained on state-change classification will demonstrate superior performance on downstream surgical anticipation tasks compared to baselines without state-change pretraining. We begin by exploring the baseline models selected for comparison and the reasoning behind these choices. 

We first present results on the state-change recognition pretraining task (continuity, discontinuity, onset detection) on both CholecT50 and GraSP datasets to establish the quality of our learned representations. We then evaluate downstream transfer performance across all five tasks in the \textbf{SFPBench} benchmark, comparing our state-change pretrained SurgFUTR variants against baselines without explicit state-change modeling. 
Finally, we assess cross-procedure transfer learning by transferring models pretrained on cholecystectomy procedures to gastric bypass procedures.
This comprehensive evaluation demonstrates how state-change pretraining enhances transfer learning for surgical anticipation both within and across different surgical procedures.

We conduct ablation studies to validate the importance of each component in SurgFUTR. We also provide qualitative analysis through visualizations of learned state representations, demonstrating how our model captures meaningful surgical state transitions.

\subsection{Baselines}
Comprehensive model evaluation requires carefully chosen baselines that establish meaningful performance comparisons across different architectural choices and initialization strategies. Table~\ref{tab:results:quantitative:sc_verb_cholect50}-\ref{tab:results:quantitative:sc_steps_grasp} presents our systematic comparison spanning multiple method-backbone-initialization combinations.

\textbf{VideoMAEv2 Variants:} Our primary baseline uses VideoMAEv2~\citep{wang2023videomaev2} with ViT-S backbone across three initialization strategies: \textit{Random} initialization provides a lower bound without prior knowledge; \textit{Kinetics-400} leverages general video understanding; and \textit{Phase} extends Kinetics-400 weights with surgical phase recognition pretraining, establishing domain-specific video representations.



\textbf{Self-Supervised Learning Models:} We evaluate SSL approaches with varying model scales and pretraining data sizes.
\textit{Small-scale models:} MoCoV2 and DINO, both using ResNet50 backbones pretrained on Cholec80 surgical images (weights from~\citep{ramesh2023dissecting}). These models demonstrate how general SSL techniques perform when trained specifically on surgical images, providing domain-adapted visual representations without explicit temporal modeling.
\textit{Large-scale models:} EndoViT~\citep{batic2024endovit} (ViT-B with MAE pretraining on Endo700k surgical images; weights from HuggingFace) and SurgeNetXL~\citep{jaspers2025scaling} (CaFormerS18 with DINO pretraining on $\sim$4.7M surgical video frames; weights from the open-source GitHub repository). Among SurgeNetXL variants, we select the procedure-agnostic version for consistency with other SSL baselines. These surgical foundation models have demonstrated effectiveness across multiple tasks including phase recognition, action triplets, and semantic segmentation.
Since all SSL models (MoCoV2, DINO, EndoViT, SurgeNetXL) were originally designed for single-frame analysis, we adapt them to video-level tasks by computing per-frame features and stacking them temporally.

\textbf{Proposed Methods:} 
Our proposed SurgFUTR variants share ViT-S backbones and Kinetics-400 initialization but employ different learning strategies: direct future feature prediction without explicit state modeling (Lite), student-only (S), and teacher-student distillation (TS).

For fair comparison, all models use identical fine-tuning protocols on downstream tasks as described in Section~\ref{impl_details}. The key difference is in the initialization: all our SurgFUTR variants are initialized from state-change pretrained weights (trained on CholecT50 or GraSP verb/step annotations), while baseline models are initialized directly from their original pretrained weights (Kinetics-400 for VideoMAEv2, Cholec80 for MoCoV2/DINO, Endo700k for EndoViT, etc.) without any state-change pretraining stage.

\subsection{State-Change Recognition Pretraining}

\subsubsection{Results on CholecT50}
Table~\ref{tab:results:quantitative:sc_verb_cholect50} presents performance on the state-change recognition pretraining task for \textit{verb} classes in CholecT50 dataset~\citep{nwoye2021rendezvous}. We report mAP and F1-score averaged across $3$ random seeds to provide a holistic assessment across different methods, backbones, and initialization strategies.

Predicting continuity consistently outperforms discontinuity and onset across all models. This is expected since continuity ($1-1$) indicates an action present in both timesteps $t$ and $t+\delta$, requiring no identification of missing cues. 
In contrast, discontinuity and onset are both challenging tasks with comparable difficulty levels, as evidenced by their similar performance ranges across all models. Both involve detecting temporal changes across video clips but require different types of temporal reasoning. Discontinuity detection requires identifying semantic cues present in the current clip but absent in the future clip—a task that demands proper future context rather than relying solely on current features. Onset detection, similar to anticipation, seeks cues that emerge in the future clip but are not captured in the current timestep. The comparable performance between these two tasks suggests that both forward and backward temporal reasoning present similar challenges for state-change recognition.

\textbf{Impact of Initialization:} VideoMAEv2 with ViT-S backbone demonstrates clear benefits from proper initialization. \textit{Random} initialization performs poorly (26.6\% mAP), while \textit{Kinetics-400} initialization improves performance by $7.8$pp in mAP and $7.4$pp in F1-Score. 
\textit{Phase} pretraining provides minimal additional benefit over \textit{Kinetics-400} in overall performance, with mean mAP unchanged at 34.4\%. However, \textit{Phase} pretraining achieves the best onset mAP (15.0\%) among all baselines, suggesting that surgical domain knowledge from phase labels can enhance specific aspects of temporal change detection despite limited overall gains.

\begin{table*}[h]
    \centering
    \caption{Model performance comparison on GraSP for step state-change recognition. Results are reported per state (continuity, discontinuity, onset) and their mean. Entries are mean ± standard deviation across $3$ random runs. Higher values indicate better performance.}
    \label{tab:results:quantitative:sc_steps_grasp}
    \setlength{\tabcolsep}{6pt}
    \resizebox{\textwidth}{!}{%
    \begin{tabular}{lllcccccccc}
    \toprule
        \multirow{2}{*}{\textbf{Method}} &
        \multirow{2}{*}{\textbf{Backbone}} &
        \multirow{2}{*}{\textbf{Initialization}} &
        \multicolumn{4}{c}{\textbf{mAP (\%)}} &
        \multicolumn{4}{c}{\textbf{F1-Score (\%)}} \\
        \cmidrule(lr){4-7} 
        \cmidrule(lr){8-11} 
        & & & $\textbf{Continuity}$ & $\textbf{Discontinuity}$ & $\textbf{Onset}$ & $\textbf{mean}$ &
        $\textbf{Continuity}$ & $\textbf{Discontinuity}$ & $\textbf{Onset}$ & $\textbf{mean}$ \\
        \midrule
        VideoMAEv2 & ViT-S & Random & $19.9\pm\scriptstyle{0.9}$ & $5.0\pm\scriptstyle{0.2}$ & 
        $5.2\pm\scriptstyle{0.1}$ & 
        $10.1\pm\scriptstyle{0.3}$ &
        $18.1\pm\scriptstyle{0.3}$ & $9.4\pm\scriptstyle{0.3}$ &
        $9.8\pm\scriptstyle{0.6}$ & $12.4\pm\scriptstyle{0.1}$ \\
        VideoMAEv2 & ViT-S & Kinetics-400 & $36.8 \pm \scriptstyle{0.3}$ & $10.6 \pm \scriptstyle{0.5}$ &
        $12.6 \pm \scriptstyle{0.6}$ & 
        $20.0 \pm \scriptstyle{0.2}$ & 
        $37.0 \pm \scriptstyle{0.5}$ & $16.4 \pm \scriptstyle{5.4}$ &
        $17.7 \pm \scriptstyle{4.1}$ & $23.7 \pm \scriptstyle{3.3}$ \\
        MoCoV2 & ResNet50 & Cholec80 & $34.1 \pm \scriptstyle{1.0}$ & $\textbf{13.0} \pm \scriptstyle{0.9}$ &
        $13.0 \pm \scriptstyle{0.4}$ & 
        $20.0 \pm \scriptstyle{0.0}$ & 
        $36.7 \pm \scriptstyle{3.4}$ & $17.7 \pm \scriptstyle{3.5}$ &
        $17.9 \pm \scriptstyle{3.9}$ & $24.0 \pm \scriptstyle{3.5}$ \\
        DINO & ResNet50 & Cholec80 & $33.9 \pm \scriptstyle{0.7}$ & $12.9 \pm \scriptstyle{0.8}$ &
        $13.2 \pm \scriptstyle{1.0}$ & 
        $20.0 \pm \scriptstyle{0.6}$ & 
        $36.7 \pm \scriptstyle{2.3}$ & 
        $\textbf{20.6} \pm \scriptstyle{1.4}$ &
        $17.8 \pm \scriptstyle{5.1}$ & 
        $24.0 \pm \scriptstyle{4.1}$ \\
        EndoViT & ViT-B & Endo700k & $19.7 \pm \scriptstyle{1.8}$ & $06.1 \pm \scriptstyle{0.6}$ &
        $06.2 \pm \scriptstyle{0.8}$ & 
        $10.7 \pm \scriptstyle{1.1}$ & 
        $20.1 \pm \scriptstyle{2.1}$ & 
        $10.7 \pm \scriptstyle{0.9}$ &
        $10.4 \pm \scriptstyle{1.2}$ & 
        $13.7 \pm \scriptstyle{1.4}$ \\
        SurgeNetXL & CaFormerS18 & SurgeNetXL & $20.0 \pm \scriptstyle{11.8}$ & $06.7 \pm \scriptstyle{2.1}$ &
        $07.1 \pm \scriptstyle{2.4}$ & 
        $11.3 \pm \scriptstyle{5.3}$ & 
        $20.3 \pm \scriptstyle{9.6}$ & 
        $11.8 \pm \scriptstyle{2.1}$ &
        $11.8 \pm \scriptstyle{2.7}$ & 
        $14.6 \pm \scriptstyle{4.7}$ \\
        \midrule
        SurgFUTR-Lite & ViT-S & Kinetics-400 & $37.9 \pm \scriptstyle{1.2}$ & $12.3 \pm \scriptstyle{1.2}$ &
        $\textbf{14.9} \pm \scriptstyle{1.1}$ & 
        $21.7 \pm \scriptstyle{1.1}$ & 
        $34.8 \pm \scriptstyle{0.9}$ & $17.4 \pm \scriptstyle{1.2}$ &
        $\textbf{21.4} \pm \scriptstyle{0.9}$ & $24.5 \pm \scriptstyle{0.8}$ \\
        SurgFUTR-S & ViT-S & Kinetics-400 & $36.8 \pm \scriptstyle{2.5}$ & $10.1 \pm \scriptstyle{0.9}$ &
        $12.4 \pm \scriptstyle{0.9}$ & 
        $19.8 \pm \scriptstyle{0.5}$ & 
        $36.1 \pm \scriptstyle{1.0}$ & $15.2 \pm \scriptstyle{3.7}$ &
        $17.3 \pm \scriptstyle{3.5}$ & $22.9 \pm \scriptstyle{2.7}$ \\
        \rowcolor{cyan!15}
        SurgFUTR-TS & ViT-S & Kinetics-400 & $\textbf{38.8} \pm \scriptstyle{1.8}$ & $12.7 \pm \scriptstyle{1.3}$ &
        $14.2 \pm \scriptstyle{0.8}$ & 
        $\textbf{21.9} \pm \scriptstyle{0.5}$ & 
        $\textbf{37.8} \pm \scriptstyle{1.6}$ & $20.2 \pm \scriptstyle{1.1}$ &
        $21.1 \pm \scriptstyle{0.8}$ & $\textbf{26.4} \pm \scriptstyle{0.3}$ \\
        
    \bottomrule
    \end{tabular}
    }
\end{table*}

\textbf{Backbone and Method Comparison:} Recent SSL methods (MoCoV2, DINO) with ResNet50 backbones pretrained on Cholec80 show competitive performance, achieving comparable results to bigger model like EndoViT. Surprisingly, EndoViT with ViT-B backbone performs worse than VideoMAEv2 with \textit{Random} initialization, despite being pretrained on extensive surgical data. The ResNet50-based SSL models still fall short of VideoMAEv2 with Kinetics-400 initialization, with maximum gaps of $1.2$pp in mAP and $0.7$pp in F1-Score. While MoCoV2 and DINO improve onset F1-Score over other baselines, they underperform in discontinuity detection.

\textbf{SurgFUTR Variants:} Our proposed methods all use ViT-S backbones with \textit{Kinetics-400} initialized weights but differ in their state-change learning approach. SurgFUTR-Lite, with direct future feature prediction, achieves strong performance (35.6\% mAP, 36.7\% F1-Score) that surpasses all baselines, demonstrating the effectiveness of future context modeling for state-change recognition. SurgFUTR-S, incorporating explicit state representation without future context, achieves slightly lower performance (35.1\% mAP, 36.6\% F1-Score), showing modest improvement in discontinuity detection ($0.2$pp over SurgFUTR-Lite in mAP) but notably lower onset performance ($1.6$pp drop in mAP), suggesting that future context is particularly important for anticipating state changes.


SurgFUTR-TS, our complete framework combining state representation with future context prediction, achieves the best overall performance with $2.0$pp improvement in mAP (36.4\%) over the strongest baseline and $0.8$pp over SurgFUTR-Lite. It demonstrates consistent gains across all state categories compared to the best baseline: $2.6$pp in continuity, $1.4$pp in discontinuity, and $1.6$pp in onset mAP, with corresponding F1-Score improvements of $1.9$pp, $1.2$pp, and $2.2$pp respectively. Our complete state-change learning framework with state representation, state graph, and centroid transition modeling through the ActDyn module achieves the best performance for learning state-change features.

\subsubsection{Results on GraSP}
To assess generalizability of our state-change formulation beyond CholecT50, we evaluate SurgFUTR on the GraSP~\citep{ayobi2025pixel} dataset, which captures robot-assisted radical prostatectomy procedures with distinct anatomical structures, instruments, and workflows. This cross-domain evaluation tests whether our state-change formulation transfers across different surgical procedures.

For GraSP, we adapt our state-change pretraining to utilize surgical steps rather than action triplets. These steps represent higher-level procedural concepts that encompass collections of individual surgical action triplets, providing a coarser but more structured temporal segmentation. This adaptation demonstrates the flexibility of our state-change formulation to work with different levels of surgical task granularity.

Table~\ref{tab:results:quantitative:sc_steps_grasp} presents the results, reporting mean average precision and F1-scores averaged across 3 random seeds following the same evaluation protocol as CholecT50. Similar to CholecT50, continuity detection significantly outperforms discontinuity and onset across all methods, reflecting the inherent difficulty of detecting temporal changes in surgical workflows.

\textbf{Impact of Initialization:} VideoMAEv2 with ViT-S backbone shows dramatic improvement from proper initialization on GraSP. \textit{Random} initialization yields poor performance (10.1\% mAP), while \textit{Kinetics-400} initialization provides substantial gains of $9.9$pp in mAP and $11.3$pp in F1-Score, demonstrating even stronger benefits than observed on CholecT50.

\textbf{Backbone and Method Comparison:} The ResNet50-based SSL methods (MoCoV2, DINO) pretrained on Cholec80 achieve competitive performance, matching VideoMAEv2 with Kinetics-400 in mean mAP (20.0\%) and F1-Score ($\sim$24\%). Notably, these surgical domain-adapted models show particular strength in discontinuity and onset detection, outperforming the general video-pretrained VideoMAEv2. In contrast, SurgeNetXL and the larger vision transformer models EndoViT continue to underperform significantly, achieving only 11.3\% and 10.7\% mean mAP respectively, despite their extensive surgical pretraining. This suggests that model architecture and training methodology may be more critical than dataset size for cross-domain surgical transfer.

\textbf{SurgFUTR Variants:} Our proposed methods demonstrate consistent improvements over baselines. SurgFUTR-TS achieves the best overall performance across all models with 21.9\% mAP and 26.4\% F1-Score, delivering $1.9$pp mAP and $2.4$pp F1-Score improvements over the best baseline. It excels particularly in continuity detection (38.8\% mAP, 37.8\% F1-Score) while maintaining competitive performance in discontinuity and onset categories.
SurgFUTR-Lite shows strong performance with 21.7\% mAP and 24.5\% F1-Score, closely approaching SurgFUTR-TS performance while demonstrating the effectiveness of direct future feature prediction. SurgFUTR-S achieves comparable baseline performance (19.8\% mAP) but shows trade-offs across state categories compared to both other variants. This validates the generalizability of our state-change formulation across different surgical procedures and annotation granularities.

\subsection{Downstream evaluation on SFPBench}

\paragraph{\textbf{Task-specific baselines}}
In addition to the general self-supervised and foundation-pretrained models, we evaluate a set of task-specific baselines from prior work on surgical anticipation and remaining surgery duration (RSD) estimation. These baselines play two roles in our study. The first group consists of anticipation-pretrained models that learn future-aware visual-temporal representations and can be transferred to different downstream tasks. The second group consists of models developed specifically for duration estimation. For all baselines, we retain the original backbone family and standard initialization as closely as possible, while adapting supervision to the label space available in each dataset.

\paragraph{\textbf{Anticipation-pretrained baselines}}
We include the Anticipative Video Transformer (AVT)~\citep{girdhar2021anticipative}, RULSTM~\citep{furnari2020rolling}, and SlowFast RULSTM (SF-RULSTM)~\citep{osman2021slowfast} as transferable anticipation-pretrained baselines. AVT uses a ViT-B backbone initialized from ImageNet-21K and is trained in two stages: first for causal feature forecasting and next-action anticipation, and then fine-tuned for the downstream task. RULSTM and SF-RULSTM use InceptionV3 backbones initialized from ImageNet-1K and are trained in the same pretrain-then-transfer manner. Anticipation supervision is defined over verb labels for CholecT50 and step labels for GraSP. SF-RULSTM extends RULSTM with slow and fast temporal streams to capture complementary temporal dynamics. We use these models for both short-term and long-term downstream evaluation because their training objective is anticipation-based rather than tied to a specific endpoint such as phase recognition or RSD estimation.

\begin{table*}[h]
    \centering
    \caption{Comparison transparency summary for the main baselines and SurgFUTR variants. Checkmarks indicate the downstream tasks used for evaluation. Abbreviations: PT Obj. = pretraining objective; VMAE = video masked autoencoder; PR = phase recognition; C-SSL = contrastive self-supervised learning; SurgFPT = surgical foundation model pretraining; ANTR = action anticipation; SC PT = state-change pretraining; TS = teacher-student distillation. Dataset-specific clip lengths are described in the implementation details.}
    \label{tab:comparison_transparency}
    \setlength{\tabcolsep}{14pt}
    \resizebox{\textwidth}{!}{%
    \begin{tabular}{lll l ccccc}
    \toprule
        \multirow{2}{*}{\textbf{Method}} &
        \multirow{2}{*}{\textbf{Backbone}} &
        \multirow{2}{*}{\textbf{Initialization}} &
        \multirow{2}{*}{\textbf{PT Obj.}} &
        \multicolumn{5}{c}{\textbf{Evaluation Scope}} \\
        \cmidrule(lr){5-9}
        & & & &
        \textbf{SFP-I} &
        \textbf{SFP-II} &
        \textbf{SFP-III} &
        \textbf{SFP-IV} &
        \textbf{SFP-V} \\
    \midrule

    \multicolumn{9}{l}{\textit{General SSL / Foundation Pretraining}} \\
    VideoMAEv2 & ViT-S & Random & VMAE & \checkmark & \checkmark & \checkmark & \checkmark & \checkmark \\
    VideoMAEv2 & ViT-S & Kinetics-400 & VMAE & \checkmark & \checkmark & \checkmark & \checkmark & \checkmark \\
    VideoMAEv2 & ViT-S & Phase & PR & \checkmark & \checkmark & \checkmark & \checkmark & \checkmark \\
    MoCoV2 & ResNet50 & Cholec80 & C-SSL & \checkmark & \checkmark & \checkmark & \checkmark & \checkmark \\
    DINO & ResNet50 & Cholec80 & C-SSL & \checkmark & \checkmark & \checkmark & \checkmark & \checkmark \\
    EndoViT & ViT-B & Endo700k & SurgFPT & \checkmark & \checkmark & \checkmark & \checkmark & \checkmark \\
    SurgeNetXL & CaFormerS18 & SurgeNetXL & SurgFPT & \checkmark & \checkmark & \checkmark & \checkmark & \checkmark \\

    \midrule
    \multicolumn{9}{l}{\textit{Task-Specific Baselines}} \\
    TimeLSTM & ResNet152 & ImageNet-1K & -- & \checkmark & -- & -- & -- & -- \\
    TransLocal & ResNet101 & ImageNet-1K & -- & \checkmark & -- & -- & -- & -- \\
    RSDNet & ResNet152 & ImageNet-1K & -- & \checkmark & -- & -- & -- & -- \\
    BD-Net & DenseNet169 & ImageNet-1K & -- & \checkmark & -- & -- & -- & -- \\
    AVT & ViT-B & ImageNet-21K & ANTR & \checkmark & \checkmark & \checkmark & \checkmark & \checkmark \\
    RULSTM & InceptionV3 & ImageNet-1K & ANTR & \checkmark & \checkmark & \checkmark & \checkmark & \checkmark \\
    SF-RULSTM & InceptionV3 & ImageNet-1K & ANTR & \checkmark & \checkmark & \checkmark & \checkmark & \checkmark \\

    \midrule
    \multicolumn{9}{l}{\textit{Our Method (State-Change Pretraining)}} \\
    SurgFUTR-Lite & ViT-S & State-Change & SC PT & \checkmark & \checkmark & \checkmark & \checkmark & \checkmark \\
    SurgFUTR-S & ViT-S & State-Change & SC PT & \checkmark & \checkmark & \checkmark & \checkmark & \checkmark \\
    SurgFUTR-TS & ViT-S & State-Change & SC PT + TS & \checkmark & \checkmark & \checkmark & \checkmark & \checkmark \\

    \bottomrule
    \end{tabular}
    }
\end{table*}

\paragraph{\textbf{RSD-specific baselines}}
To complement these transferable anticipation baselines, we also evaluate methods whose primary design is aligned with remaining-duration estimation. TimeLSTM~\citep{aksamentov2017deep} and RSDNet~\citep{twinanda2018rsdnet} are implemented with ResNet152 backbones initialized from ImageNet-1K, TransLocal~\citep{loukas2024prediction} uses a ResNet101 backbone initialized from ImageNet-1K, and BD-Net~\citep{wu2023nonlinear} uses a DenseNet169 backbone initialized from ImageNet-1K. Unlike AVT and the RULSTM family, these methods are included as direct RSD baselines, since their modeling assumptions are centered on surgical progress and remaining-time prediction. Accordingly, they are evaluated only for the RSD task.

\paragraph{\textbf{Scope of comparison}}
We restrict the task-specific comparison to baselines whose supervision requirements and temporal assumptions are compatible with the annotation protocol of the present benchmarks. Methods that require additional supervision beyond the available labels, such as explicit instrument instance annotations, interaction cues, or bounding-box-level supervision, are therefore excluded~\citep{yuan2022anticipation}. We also do not include CataNet~\citep{marafioti2021catanet} separately, as its formulation is closely related to the BD-Net family~\citep{wu2023nonlinear}; accordingly, we retain BD-Net as the representative baseline for RSD prediction task in this category. Similarly, we exclude SWAG~\citep{boels2025swag} from direct comparison because its design relies on densely sampled long-range temporal context, operating over input sequences spanning approximately 24 minutes of video, whereas our setting is clip-based, with inputs limited to 8 frames in CholecT50 and 4 frames in GraSP. Under such conditions, a faithful reproduction of the SWAG protocol would not be comparable. Since SWAG is itself built upon an AVT-based anticipation backbone, the AVT baseline included in our study provides the closest comparable approximation to its underlying anticipation strategy under our input regime. Finally, we do not include the Bayesian anticipation model of Rivoir et al.~\citep{rivoir2020rethinking}, as it is formulated for sparse surgical instrument anticipation using an AlexNet-based visual representation, which is both architecturally outdated relative to the transformer-based AVT baseline and mismatched to our label space, particularly for GraSP where instrument anticipation is not defined. Altogether, these choices ensure that the downstream evaluation reflects differences in representation learning and temporal modeling under a shared supervision setting, rather than differences induced by unequal annotation requirements or incompatible input assumptions.


\paragraph{\textbf{SFP-I: Remaining Surgery Duration (RSD) Prediction}} 
Tables~\ref{tab:results:quantitative:rsd_only_ct50} and~\ref{tab:results:quantitative:rsd_only_grasp} present RSD prediction results on CholecT50 and GraSP datasets respectively, evaluated using mean absolute error (MAE) where lower values indicate better performance.

\begin{table}[h]
    \centering
    \caption{Model transfer performance comparison on CholecT50: SFP-I (RSD) by MAE; lower is better.}
    \label{tab:results:quantitative:rsd_only_ct50}
    \setlength{\tabcolsep}{10pt}
    \resizebox{\columnwidth}{!}{%
    \begin{tabular}{lllc}
    \toprule
        \textbf{Method} &
        \textbf{Backbone} &
        \textbf{Initialization} & 
        \textbf{RSD} \\
        \midrule
        \multicolumn{4}{l}{\textit{General SSL / Foundation Pretraining}} \\
        VideoMAEv2 & ViT-S & Random & $2.051\pm\scriptstyle{0.097}$ \\
        VideoMAEv2 & ViT-S & Kinetics-400 &  $1.655\pm\scriptstyle{0.068}$ \\
        VideoMAEv2 & ViT-S & Phase  &  $1.536\pm\scriptstyle{0.095}$ \\
        MoCoV2 & ResNet50 & Cholec80 &  $1.515\pm\scriptstyle{0.040}$ \\
        DINO & ResNet50 & Cholec80 & $1.678\pm\scriptstyle{0.128}$ \\
        EndoViT & ViT-B & Endo700k & $2.091\pm\scriptstyle{0.049}$ \\
        SurgeNetXL & CaFormerS18 & SurgeNetXL & $2.090\pm\scriptstyle{0.047}$ \\

        \midrule
        \multicolumn{4}{l}{\textit{Task-Specific Baselines}} \\
        TimeLSTM & ResNet152 & ImageNet-1K & $1.951\pm\scriptstyle{0.084}$ \\
        TransLocal & ResNet101 & ImageNet-1K & $1.990\pm\scriptstyle{0.281}$ \\
        RSDNet & ResNet152 & ImageNet-1K & $2.041\pm\scriptstyle{0.038}$ \\
        BD-Net & DenseNet169 & ImageNet-1K & $1.708\pm\scriptstyle{0.115}$ \\
        AVT & ViT-B & ImageNet-21K & $1.475\pm\scriptstyle{0.124}$ \\
        RULSTM & InceptionV3 & ImageNet-1K & $1.570\pm\scriptstyle{0.020}$ \\
        SF-RULSTM & InceptionV3 & ImageNet-1K & $1.754\pm\scriptstyle{0.235}$ \\
        
        \midrule
        \multicolumn{4}{l}{\textit{Our Method (State-Change Pretraining)}} \\
        SurgFUTR-Lite & ViT-S & State-Change & $1.644\pm\scriptstyle{0.046}$ \\
        SurgFUTR-S & ViT-S & State-Change & $1.741\pm\scriptstyle{0.157}$ \\
        \rowcolor{cyan!15}
        SurgFUTR-TS & ViT-S & State-Change & $\textbf{1.465}\pm\scriptstyle{0.041}$ \\
        \bottomrule 
    \end{tabular}
    }
\end{table}

\textbf{Baseline Performance Across Datasets:}
Among the general SSL and foundation-model baselines, consistent trends emerge across both surgical domains, revealing the importance of architectural choice over model scale. On CholecT50, ResNet50-based SSL methods achieve strong performance: MoCoV2 (1.515 MAE) and DINO (1.678 MAE) substantially outperform larger ViT-based surgical foundation models EndoViT (2.091 MAE) and SurgeNetXL (2.090 MAE) by 38.0\% and 24.6\% respectively. On GraSP, this pattern persists with MoCoV2 (7.607 MAE) and DINO (7.684 MAE) outperforming EndoViT (8.023 MAE) and SurgeNetXL (8.046 MAE) by 5.5\% and 4.7\% respectively. Notably, the performance gap is more pronounced on CholecT50 (up to 38\% improvement) than GraSP (up to 5.5\%), suggesting that surgical foundation models struggle particularly with temporal forecasting in more complex procedures, despite their larger parameter counts and large scale pretraining.

\textbf{Initialization Effects:}
VideoMAEv2 demonstrates substantial sensitivity to initialization across both datasets. On CholecT50, Kinetics-400 (1.655 MAE) and Phase (1.536 MAE) initialization significantly outperform random initialization (2.051 MAE) by 19.3\% and 25.1\% respectively, with Phase pretraining providing an additional 7.2\% improvement over Kinetics-400. On GraSP, Kinetics-400 (7.301 MAE) provides 9.3\% improvement over random initialization (8.053 MAE). The consistent benefits of general video pretraining (Kinetics-400) across both datasets validate that natural video understanding provides a strong inductive bias for surgical temporal reasoning, while surgical phase-level supervision (Phase) further refines this capability on CholecT50.

\begin{table}[h]
    \centering
    \caption{Model transfer performance comparison on GraSP: SFP-I (RSD) by MAE; lower is better.}
    \label{tab:results:quantitative:rsd_only_grasp}
    \setlength{\tabcolsep}{10pt}
    \resizebox{\columnwidth}{!}{%
    \begin{tabular}{lllc}
    \toprule
        \textbf{Method} &
        \textbf{Backbone} &
        \textbf{Initialization} & 
        \textbf{RSD} \\
        \midrule
        \multicolumn{4}{l}{\textit{General SSL / Foundation Pretraining}} \\
        VideoMAEv2 & ViT-S & Random & $8.053\pm\scriptstyle{0.188}$ \\
        VideoMAEv2 & ViT-S & Kinetics-400 & $7.301\pm\scriptstyle{0.160}$ \\
        MoCoV2 & ResNet50 & Cholec80 & $7.607\pm\scriptstyle{0.886}$ \\
        DINO & ResNet50 & Cholec80 & $7.684\pm\scriptstyle{0.937}$ \\
        EndoViT & ViT-B & Endo700k & $8.023\pm\scriptstyle{0.213}$ \\
        SurgeNetXL & CaFormerS18 & SurgeNetXL & $8.046\pm\scriptstyle{0.518}$ \\

        \midrule
        \multicolumn{4}{l}{\textit{Task-Specific Baselines}} \\
        TimeLSTM & ResNet152 & ImageNet-1K & $8.214\pm\scriptstyle{0.256}$ \\
        TransLocal & ResNet101 & ImageNet-1K & $8.608\pm\scriptstyle{0.173}$ \\
        RSDNet & ResNet152 & ImageNet-1K & $8.133\pm\scriptstyle{0.273}$ \\
        BD-Net & DenseNet169 & ImageNet-1K & $8.349\pm\scriptstyle{0.310}$ \\
        AVT & ViT-B & ImageNet-21K & $7.231\pm\scriptstyle{0.186}$ \\
        RULSTM & InceptionV3 & ImageNet-1K & $7.469\pm\scriptstyle{0.510}$ \\
        SF-RULSTM & InceptionV3 & ImageNet-1K & $7.409\pm\scriptstyle{0.688}$ \\
        
        \midrule
        \multicolumn{4}{l}{\textit{Our Method (State-Change Pretraining)}} \\
        SurgFUTR-Lite & ViT-S & State-Change & $7.111\pm\scriptstyle{0.087}$ \\
        SurgFUTR-S & ViT-S & State-Change & $7.319\pm\scriptstyle{0.188}$ \\
        \rowcolor{cyan!15}
        SurgFUTR-TS & ViT-S & State-Change & $\textbf{7.071}\pm\scriptstyle{0.190}$ \\
        \bottomrule 
    \end{tabular}
    }
\end{table}

\textbf{Task-Specific Baselines:}
On CholecT50, the anticipation-pretrained baselines are consistently stronger than the RSD-specific baselines. AVT achieves the best task-specific result at 1.475 MAE, narrowly trailing SurgFUTR-TS by 0.7\%, while RULSTM is also competitive at 1.570 MAE. SF-RULSTM does not improve over RULSTM in this setting, yielding 1.754 MAE. Among the duration-oriented baselines, BD-Net is the strongest at 1.708 MAE, outperforming TimeLSTM (1.951 MAE), TransLocal (1.990 MAE), and RSDNet (2.041 MAE), but remaining weaker than both AVT and RULSTM. Overall, these results suggest that anticipation-based pretraining transfers more effectively to clip-level RSD prediction on CholecT50 than direct duration-oriented modeling.

On GraSP, the same overall trend is preserved, although the gap between task-specific baselines is narrower. AVT again provides the strongest task-specific result at 7.231 MAE, followed by SF-RULSTM at 7.409 MAE and RULSTM at 7.469 MAE. In contrast to CholecT50, the SlowFast variant is marginally better than RULSTM on GraSP, indicating that multi-rate temporal modeling may offer modest benefit in this domain. The direct RSD baselines remain uniformly weaker, with RSDNet at 8.133 MAE, TimeLSTM at 8.214 MAE, BD-Net at 8.349 MAE, and TransLocal at 8.608 MAE. Taken together, the GraSP results reinforce the same conclusion as CholecT50: anticipation-pretrained baselines provide a stronger transfer signal for RSD prediction than architectures optimized directly for remaining-duration estimation.

\textbf{SurgFUTR Performance Across Datasets:}
State-change pretrained variants consistently achieve best-in-class performance across both datasets. SurgFUTR-TS delivers optimal results on CholecT50 (1.465 MAE) and GraSP (7.071 MAE), outperforming the strongest general baselines—MoCoV2 at 1.515 MAE (3.3\% improvement) and Kinetics-400 VideoMAEv2 at 7.301 MAE (3.2\% improvement) respectively, as well as the strongest task-specific baseline AVT on both datasets (1.475 MAE on CholecT50 and 7.231 MAE on GraSP). SurgFUTR-Lite shows mixed behavior: it is worse than MoCoV2 on CholecT50 (1.644 MAE; +8.5\% error) but improves over Kinetics-400 on GraSP (7.111 MAE; 2.6\% better), while remaining below the best task-specific baseline on CholecT50 and only modestly ahead of it on GraSP. SurgFUTR-S is close to Kinetics-400 on GraSP (7.319 MAE; 0.25\% worse) but lags on CholecT50 (1.741 MAE; 18.8\% worse than TS). Overall, the consistent gains of SurgFUTR-TS indicate that explicit state-change modeling with centroid transition dynamics (ActDyn) enhances long-horizon temporal forecasting beyond SSL and general video pretraining, while also improving over dedicated task-specific baselines for RSD prediction.
Although the absolute RSD improvement on CholecT50 is modest in isolation, SurgFUTR-TS remains competitive on both datasets and performs favorably relative to both general and task-specific baselines under the same evaluation protocol.



We also contextualize the RSD comparisons in absolute terms. Following the same normalization convention as RSDNet, we set the RSD normalization factor to $s_{\text{norm}}=5$, which expresses the target in 5-minute units and keeps the regression range numerically compact. Under this scaling, an MAE difference of 1.0 corresponds to 5 minutes of absolute prediction error. On CholecT50, the improvement of SurgFUTR-TS over AVT (1.475 to 1.465) corresponds to approximately 0.05 minutes, i.e., about 3 seconds, while the improvement over MoCoV2 (1.515 to 1.465) corresponds to approximately 0.25 minutes, i.e., about 15 seconds, indicating only modest absolute gains. On GraSP, the margins are more substantial: the improvement over AVT (7.231 to 7.071) corresponds to approximately 0.80 minutes, i.e., about 48 seconds, and the improvement over Kinetics-400 VideoMAEv2 (7.301 to 7.071) corresponds to approximately 1.15 minutes, i.e., about 69 seconds. Overall, the absolute RSD gains are modest on CholecT50 but noticeably larger on GraSP.

\textbf{Transfer to MultiBypass140.} 
Table~\ref{tab:results:quantitative:rsd_only_mbp140_stras} and Table~\ref{tab:results:quantitative:rsd_only_mbp140_bern} present cross-procedure transfer learning results for RSD prediction on the two MultiBypass140 centers. SurgFUTR-TS achieves the best performance on both centers (Strasbourg: $2.043\pm0.015$ MAE; Bern: $1.594\pm0.025$ MAE), demonstrating effective transfer from cholecystectomy to gastric bypass procedures. Among baselines, general vision pretraining strategies show competitive performance: Phase initialization achieves $2.082\pm0.049$ MAE on Strasbourg and $1.644\pm0.006$ MAE on Bern, while Kinetics-400 achieves $2.098\pm0.029$ and $1.631\pm0.043$ respectively. 
\begin{table}[h]
    \centering
    \caption{Model transfer performance comparison on MultiBypass140 (Center: Strasbourg): SFP-I (RSD) by MAE; lower is better.}
    \label{tab:results:quantitative:rsd_only_mbp140_stras}
    \setlength{\tabcolsep}{3pt}
    \resizebox{\columnwidth}{!}{%
    \begin{tabular}{lllc}
    \toprule
        \textbf{Method} &
        \textbf{Backbone} &
        \textbf{Initialization} & 
        \textbf{RSD (Center: Strasbourg)} \\
        \midrule
        \multicolumn{4}{l}{\textit{General SSL / Foundation Pretraining}} \\
        VideoMAEv2 & ViT-S & Random & $2.582\pm\scriptstyle{0.144}$ \\
        VideoMAEv2 & ViT-S & Kinetics-400 & $2.098\pm\scriptstyle{0.029}$ \\
        VideoMAEv2 & ViT-S & Phase & $2.082\pm\scriptstyle{0.049}$ \\
        MoCoV2 & ResNet50 & Cholec80 & $2.152\pm\scriptstyle{0.002}$ \\
        DINO & ResNet50 & Cholec80 & $2.154\pm\scriptstyle{0.072}$ \\
        EndoViT & ViT-B & Endo700k & $2.743\pm\scriptstyle{0.015}$ \\
        SurgeNetXL & CaFormerS18 & SurgeNetXL & $2.242\pm\scriptstyle{0.035}$ \\

        \midrule
        \multicolumn{4}{l}{\textit{Task-Specific Baselines}} \\
        TimeLSTM & ResNet152 & ImageNet-1K & $2.492\pm\scriptstyle{0.070}$ \\
        TransLocal & ResNet101 & ImageNet-1K & $2.448\pm\scriptstyle{0.189}$ \\
        RSDNet & ResNet152 & ImageNet-1K & $2.379\pm\scriptstyle{0.055}$ \\
        BD-Net & DenseNet169 & ImageNet-1K & $2.348\pm\scriptstyle{0.033}$ \\
        AVT & ViT-B & ImageNet-21K & $2.105\pm\scriptstyle{0.029}$ \\
        RULSTM & InceptionV3 & ImageNet-1K & $2.210\pm\scriptstyle{0.048}$ \\
        SF-RULSTM & InceptionV3 & ImageNet-1K & $2.373\pm\scriptstyle{0.285}$ \\
        
        \midrule
        \multicolumn{4}{l}{\textit{Our Method (State-Change Pretraining)}} \\
        SurgFUTR-Lite & ViT-S & State-Change & $2.091\pm\scriptstyle{0.011}$ \\
        SurgFUTR-S & ViT-S & State-Change & $2.064\pm\scriptstyle{0.010}$ \\
        \rowcolor{cyan!15}
        SurgFUTR-TS & ViT-S & State-Change & $\textbf{2.043}\pm\scriptstyle{0.015}$ \\
        \bottomrule 
    \end{tabular}
    }
\end{table}
Self-supervised baselines (MoCov2, DINO) and surgical foundation models (EndoViT, SurgeNetXL) demonstrate weaker transfer, with EndoViT showing particularly poor generalization (Strasbourg: $2.743\pm0.015$ MAE; Bern: $2.040\pm0.036$ MAE). 

Among the task-specific baselines, anticipation-pretrained models again provide the strongest transfer performance across both centers. For the Strasbourg center, AVT is the best task-specific baseline at $2.105\pm0.029$ MAE, followed by RULSTM at $2.210\pm0.048$, whereas the direct RSD baselines remain weaker, with BD-Net performing best among them at $2.348\pm0.033$ MAE. A similar ordering is observed for the Bern center, where AVT achieves $1.636\pm0.027$ MAE and RULSTM attains $1.706\pm0.020$, while the strongest duration-oriented baseline, BD-Net, remains higher at $1.752\pm0.065$. SF-RULSTM does not improve over RULSTM on either center, yielding $2.373\pm0.285$ on Strasbourg and $1.804\pm0.235$ on Bern. Taken together, these results suggest that anticipation-pretrained representations transfer more effectively across procedures and clinical sites than architectures optimized directly for remaining-duration estimation.

Within SurgFUTR variants, the full teacher-student framework (TS) consistently outperforms lightweight variants (Lite, S) by $2.3\%$ and $1.0\%$ on Strasbourg, and $2.1\%$ and $2.4\%$ on Bern, validating that comprehensive state-change modeling enhances cross-procedure generalization.
This inter-center performance difference is consistent with the underlying characteristics of MultiBypass140 \citep{Lavanchy2024}. 
In particular, the published benchmark reports a substantially shorter average procedure duration for Bern than for Strasbourg (72 vs. 110 minutes), suggesting a meaningful difference in temporal workflow structure, remaining-duration range, and overall procedure progression between the two centers.

Notably, performance varies significantly across centers, with Bern showing lower absolute MAE values across all methods, suggesting center-specific characteristics influence temporal prediction difficulty despite identical preprocessing and training protocols.

\begin{table}[h]
    \centering
    \caption{Model transfer performance comparison on MultiBypass140 (Center: Bern): SFP-I (RSD) by MAE; lower is better.}
    \label{tab:results:quantitative:rsd_only_mbp140_bern}
    \setlength{\tabcolsep}{6pt}
    \resizebox{\columnwidth}{!}{%
    \begin{tabular}{lllc}
    \toprule
        \textbf{Method} &
        \textbf{Backbone} &
        \textbf{Initialization} & 
        \textbf{RSD (Center: Bern)} \\
        \midrule
        \multicolumn{4}{l}{\textit{General SSL / Foundation Pretraining}} \\
        VideoMAEv2 & ViT-S & Random & $1.946\pm\scriptstyle{0.007}$ \\
        VideoMAEv2 & ViT-S & Kinetics-400 & $1.631\pm\scriptstyle{0.043}$ \\
        VideoMAEv2 & ViT-S & Phase & $1.644\pm\scriptstyle{0.006}$ \\
        MoCoV2 & ResNet50 & Cholec80 & $1.673\pm\scriptstyle{0.052}$ \\
        DINO & ResNet50 & Cholec80 & $1.637\pm\scriptstyle{0.035}$ \\
        EndoViT & ViT-B & Endo700k & $2.040\pm\scriptstyle{0.036}$ \\
        SurgeNetXL & CaFormerS18 & SurgeNetXL & $1.921\pm\scriptstyle{0.266}$ \\

        \midrule
        \multicolumn{4}{l}{\textit{Task-Specific Baselines}} \\
        TimeLSTM & ResNet152 & ImageNet-1K & $1.924\pm\scriptstyle{0.155}$ \\
        TransLocal & ResNet101 & ImageNet-1K & $1.791\pm\scriptstyle{0.042}$ \\
        RSDNet & ResNet152 & ImageNet-1K & $1.866\pm\scriptstyle{0.013}$ \\
        BD-Net & DenseNet169 & ImageNet-1K & $1.752\pm\scriptstyle{0.065}$ \\
        AVT & ViT-B & ImageNet-21K & $1.636\pm\scriptstyle{0.027}$ \\
        RULSTM & InceptionV3 & ImageNet-1K & $1.706\pm\scriptstyle{0.020}$ \\
        SF-RULSTM & InceptionV3 & ImageNet-1K & $1.804\pm\scriptstyle{0.235}$ \\
        
        \midrule
        \multicolumn{4}{l}{\textit{Our Method (State-Change Pretraining)}} \\
        SurgFUTR-Lite & ViT-S & State-Change & $1.628\pm\scriptstyle{0.007}$ \\
        SurgFUTR-S & ViT-S & State-Change & $1.632\pm\scriptstyle{0.005}$ \\
        \rowcolor{cyan!15}
        SurgFUTR-TS & ViT-S & State-Change & $\textbf{1.594}\pm\scriptstyle{0.025}$ \\
        \bottomrule 
    \end{tabular}
    }
\end{table}


On the Strasbourg center, the improvement of SurgFUTR-TS over AVT (2.105 to 2.043) corresponds to approximately 0.31 minutes, i.e., about 18.6 seconds, while the improvement over VideoMAEv2-Phase (2.082 to 2.043) corresponds to approximately 0.20 minutes, i.e., about 11.7 seconds. On the Bern center, the improvement over AVT (1.636 to 1.594) corresponds to approximately 0.21 minutes, i.e., about 12.6 seconds, and the improvement over DINO (1.637 to 1.594) corresponds to approximately 0.22 minutes, i.e., about 12.9 seconds. Thus, the transfer gains are modest in absolute terms, but consistent across both centers.

\paragraph{\textbf{SFP-II: Phase Transition Prediction}} We evaluate model performance on phase transition prediction (Section~\ref{task_desc}), a critical surgical future prediction task. Tables~\ref{tab:results:quantitative:phtr_only_ct50} and~\ref{tab:results:quantitative:phtr_only_grasp} show mean absolute error (MAE) across 2, 3, and 5-minute future anticipation horizons on CholecT50 and GraSP respectively, with lower values indicating better performance.

\textbf{Baseline Performance Across Datasets:}
Among the general SSL and foundation-model baselines, ResNet50-based SSL methods consistently achieve the strongest baseline performance. On CholecT50, MoCoV2 (0.383 MAE) and DINO (0.386 MAE) significantly outperform other baselines, achieving at least 0.043 MAE improvement over VideoMAEv2 with Kinetics-400 (0.426 MAE). On GraSP, MoCoV2 (0.331 MAE) and DINO (0.344 MAE) maintain superior performance. Notably, surgical phase pretraining on CholecT50 (0.439 MAE) yields higher errors than SSL methods, while EndoViT and SurgeNetXL consistently underperform by at least 0.219 MAE on CholecT50 and 0.119 MAE on GraSP, suggesting that surgical foundation models struggle with temporal forecasting despite large scale pretraining.

\begin{table}[h]
    \centering
    \caption{Model transfer performance on CholecT50: SFP-II (phase transition) by MAE; lower is better.}
    \label{tab:results:quantitative:phtr_only_ct50}
    \setlength{\tabcolsep}{3pt}
    \resizebox{\columnwidth}{!}{%
    \begin{tabular}{@{}llcccc@{}}
    \toprule
        \multirow{2}{*}{\textbf{Method}} &
        \multirow{2}{*}{\textbf{Initialization}} &
        \multicolumn{4}{c}{\textbf{Phase Transition}} \\
        \cmidrule(lr){3-6}
        & & \textbf{2 min} & \textbf{3 min} & \textbf{5 min} & \textbf{mean} \\
        \midrule
        \multicolumn{4}{l}{\textit{General SSL / Foundation Pretraining}} \\
        VideoMAEv2 & Random &
        $0.368 \pm \scriptstyle{0.033}$ &
        $0.547 \pm \scriptstyle{0.042}$ &
        $0.944 \pm \scriptstyle{0.070}$ &
        $0.620 \pm \scriptstyle{0.048}$ \\
        VideoMAEv2 & Kinetics-400 & 
        $0.242 \pm \scriptstyle{0.008}$ &
        $0.375 \pm \scriptstyle{0.014}$ &
        $0.661 \pm \scriptstyle{0.019}$ &
        $0.426 \pm \scriptstyle{0.013}$ \\
        VideoMAEv2 & Phase  & 
        $0.245 \pm \scriptstyle{0.005}$ &
        $0.385 \pm \scriptstyle{0.006}$ &
        $0.688 \pm \scriptstyle{0.015}$ &
        $0.439 \pm \scriptstyle{0.009}$ \\
        MoCoV2 & Cholec80 & 
        $0.211 \pm \scriptstyle{0.001}$ &
        $0.331 \pm \scriptstyle{0.002}$ &
        $0.608 \pm \scriptstyle{0.009}$ &
        $0.383 \pm \scriptstyle{0.003}$ \\
        DINO & Cholec80 & 
        $0.213 \pm \scriptstyle{0.002}$ &
        $0.334 \pm \scriptstyle{0.002}$ &
        $0.613 \pm \scriptstyle{0.008}$ &
        $0.386 \pm \scriptstyle{0.003}$ \\
        EndoViT & Endo700k & 
        $0.357 \pm \scriptstyle{0.003}$ &
        $0.533 \pm \scriptstyle{0.005}$ &
        $0.925 \pm \scriptstyle{0.010}$ &
        $0.605 \pm \scriptstyle{0.004}$ \\
        SurgeNetXL & SurgeNetXL & 
        $0.332 \pm \scriptstyle{0.042}$ &
        $0.505 \pm \scriptstyle{0.061}$ &
        $0.889 \pm \scriptstyle{0.110}$ &
        $0.575 \pm \scriptstyle{0.071}$ \\

        \midrule
        \multicolumn{4}{l}{\textit{Task-Specific Baselines}} \\
        AVT & ImageNet-21K &
        $0.243 \pm \scriptstyle{0.023}$ &
        $0.363 \pm \scriptstyle{0.029}$ &
        $0.633 \pm \scriptstyle{0.047}$ &
        $0.413 \pm \scriptstyle{0.033}$ \\
        RULSTM & ImageNet-1K &
        $0.221 \pm \scriptstyle{0.027}$ &
        $0.344 \pm \scriptstyle{0.039}$ &
        $0.624 \pm \scriptstyle{0.052}$ &
        $0.396 \pm \scriptstyle{0.039}$ \\
        SF-RULSTM & ImageNet-1K &
        $0.264 \pm \scriptstyle{0.065}$ &
        $0.400 \pm \scriptstyle{0.088}$ &
        $0.712 \pm \scriptstyle{0.137}$ &
        $0.459 \pm \scriptstyle{0.097}$ \\
        
        \midrule
        \multicolumn{4}{l}{\textit{Our Method (State-Change Pretraining)}} \\
        SurgFUTR-Lite & State-Change & 
        $0.207 \pm \scriptstyle{0.004}$ &
        $0.322 \pm \scriptstyle{0.006}$ &
        $0.581 \pm \scriptstyle{0.018}$ &
        $0.370 \pm \scriptstyle{0.009}$ \\
        SurgFUTR-S & State-Change & 
        $0.214 \pm \scriptstyle{0.029}$ &
        $0.337 \pm \scriptstyle{0.045}$ &
        $0.609 \pm \scriptstyle{0.061}$ &
        $0.387 \pm \scriptstyle{0.045}$ \\
        \rowcolor{cyan!15}
        SurgFUTR-TS & State-Change & 
        $\textbf{0.196} \pm \scriptstyle{0.005}$ &
        $\textbf{0.305} \pm \scriptstyle{0.006}$ &
        $\textbf{0.557} \pm \scriptstyle{0.012}$ &
        $\textbf{0.353} \pm \scriptstyle{0.007}$ \\
        \bottomrule 
    \end{tabular}
    }
\end{table}

\textbf{Initialization Effects:} 
VideoMAEv2 demonstrates substantial sensitivity to initialization. On CholecT50, Kinetics-400 (0.426 MAE) and Phase (0.439 MAE) initialization significantly outperform random initialization (0.620 MAE) by 31.3\% and 29.2\% respectively. On GraSP, Kinetics-400 (0.385 MAE) provides 11.3\% improvement over random initialization (0.434 MAE). Interestingly, Phase pretraining offers marginal benefits over Kinetics-400 on CholecT50 (0.013 MAE difference), suggesting that phase-level annotations provide limited advantage for phase transition prediction compared to general video understanding.

\textbf{Task-Specific Baselines:}
The task-specific baselines further clarify the role of anticipation-oriented models for phase transition prediction. On CholecT50, RULSTM attains the strongest task-specific performance with a mean MAE of 0.396, followed by AVT at 0.413, while SF-RULSTM yields a higher error of 0.459. This pattern is consistent across all anticipation horizons, indicating that the simpler recurrent anticipation setup transfers more effectively to clip-level phase transition prediction than the SlowFast variant in this domain. On GraSP, the same family of models remains competitive, but the relative ordering shifts slightly: AVT achieves the lowest task-specific mean MAE at 0.312, followed closely by RULSTM at 0.316, whereas SF-RULSTM again trails at 0.349. Taken together, these results show that anticipation-oriented baselines provide strong task-specific comparisons for phase transition prediction across both datasets, with AVT and RULSTM yielding the most reliable performance, while the additional complexity of SF-RULSTM does not translate into consistent gains under the present transfer setting.


\textbf{SurgFUTR Performance Across Datasets:} 
State-change pretrained variants achieve the best performance across both datasets with dataset-specific patterns. On CholecT50 dataset, SurgFUTR-TS achieves the best overall performance (0.353 MAE), outperforming the strongest general baseline MoCoV2 by 7.8\% (0.030 MAE) and the strongest task-specific baseline RULSTM (0.396 MAE) by 10.9\%, while SurgFUTR-Lite (0.370 MAE) and SurgFUTR-S (0.387 MAE) show moderate improvements. On GraSP, all SurgFUTR variants demonstrate substantially stronger gains: SurgFUTR-TS (0.302 MAE), SurgFUTR-S (0.308 MAE), and SurgFUTR-Lite (0.316 MAE) outperform MoCoV2 by 8.8\%, 6.9\%, and 4.5\% respectively, while SurgFUTR-TS also improves over the strongest task-specific baseline AVT (0.312 MAE). The performance gains are most pronounced at longer prediction horizons (5 min): SurgFUTR-TS achieves 0.557 MAE vs. 0.608 MAE for MoCoV2 on CholecT50 (8.4\% improvement) and 0.519 MAE vs. 0.589 MAE on GraSP (11.9\% improvement), indicating that state-change pretraining enhances long-term temporal reasoning more effectively than SSL methods and task-specific anticipation baselines.

We also contextualize the phase transition results in absolute terms using the mean MAE across the 2, 3, and 5-minute horizons. Since the target is expressed directly in minutes until the next transition, an MAE difference of 0.1 corresponds to 6 seconds. On CholecT50, the mean MAE improvement of SurgFUTR-TS over the strongest task-specific baseline RULSTM (0.396 to 0.353) corresponds to approximately 2.58 seconds, while the improvement over AVT (0.413 to 0.353) corresponds to approximately 3.60 seconds; compared with the strongest general baseline MoCoV2 (0.383), the gain corresponds to approximately 1.80 seconds. On GraSP, the mean MAE improvement of SurgFUTR-TS over RULSTM (0.316 to 0.302) corresponds to approximately 0.84 seconds, while the improvement over AVT (0.312 to 0.302) corresponds to approximately 0.60 seconds; compared with the strongest general baseline MoCoV2 (0.331), the gain corresponds to approximately 1.74 seconds. Overall, the absolute gains for phase transition prediction remain modest, but they are more pronounced than those observed for RSD, particularly on CholecT50.

\begin{table}[h]
    \centering
    \caption{Model transfer performance on GraSP: SFP-II (phase transition) by MAE; lower is better.}
    \label{tab:results:quantitative:phtr_only_grasp}
    \setlength{\tabcolsep}{3pt}
    \resizebox{\columnwidth}{!}{%
    \begin{tabular}{@{}llcccc@{}}
    \toprule
        \multirow{2}{*}{\textbf{Method}} &
        \multirow{2}{*}{\textbf{Initialization}} &
        \multicolumn{4}{c}{\textbf{Phase Transition}} \\
        \cmidrule(lr){3-6}
        & & \textbf{2 min} & \textbf{3 min} & \textbf{5 min} & \textbf{mean} \\
        \midrule
        \multicolumn{4}{l}{\textit{General SSL / Foundation Pretraining}} \\
        VideoMAEv2 & Random &
        $0.275 \pm \scriptstyle{0.003}$ &
        $0.391 \pm \scriptstyle{0.005}$ &
        $0.637 \pm \scriptstyle{0.006}$ &
        $0.434 \pm \scriptstyle{0.004}$ \\
        VideoMAEv2 & Kinetics-400 & 
        $0.237 \pm \scriptstyle{0.001}$ &
        $0.341 \pm \scriptstyle{0.006}$ &
        $0.577 \pm \scriptstyle{0.010}$ &
        $0.385 \pm \scriptstyle{0.005}$ \\
        MoCoV2 & Cholec80 & 
        $0.196 \pm \scriptstyle{0.003}$ &
        $0.294 \pm \scriptstyle{0.004}$ &
        $0.503 \pm \scriptstyle{0.005}$ &
        $0.331 \pm \scriptstyle{0.004}$ \\
        DINO & Cholec80 & 
        $0.205 \pm \scriptstyle{0.006}$ &
        $0.305 \pm \scriptstyle{0.005}$ &
        $0.521 \pm \scriptstyle{0.007}$ &
        $0.344 \pm \scriptstyle{0.006}$ \\
        EndoViT & Endo700k & 
        $0.285 \pm \scriptstyle{0.002}$ &
        $0.403 \pm \scriptstyle{0.002}$ &
        $0.661 \pm \scriptstyle{0.004}$ &
        $0.450 \pm \scriptstyle{0.002}$ \\
        SurgeNetXL & SurgeNetXL & 
        $0.286 \pm \scriptstyle{0.005}$ &
        $0.409 \pm \scriptstyle{0.009}$ &
        $0.670 \pm \scriptstyle{0.007}$ &
        $0.455 \pm \scriptstyle{0.007}$ \\

        \midrule
        \multicolumn{4}{l}{\textit{Task-Specific Baselines}} \\
        AVT & ImageNet-21K &
        $0.191 \pm \scriptstyle{0.001}$ &
        $0.278 \pm \scriptstyle{0.000}$ &
        $0.467 \pm \scriptstyle{0.001}$ &
        $0.312 \pm \scriptstyle{0.001}$ \\
        RULSTM & ImageNet-1K &
        $0.187 \pm \scriptstyle{0.007}$ &
        $0.280 \pm \scriptstyle{0.009}$ &
        $0.481 \pm \scriptstyle{0.016}$ &
        $0.316 \pm \scriptstyle{0.011}$ \\
        SF-RULSTM & ImageNet-1K &
        $0.208 \pm \scriptstyle{0.027}$ &
        $0.310 \pm \scriptstyle{0.036}$ &
        $0.528 \pm \scriptstyle{0.057}$ &
        $0.349 \pm \scriptstyle{0.040}$ \\
        
        \midrule
        \multicolumn{4}{l}{\textit{Our Method (State-Change Pretraining)}} \\
        SurgFUTR-Lite & State-Change & 
        $0.202 \pm \scriptstyle{0.004}$ &
        $0.283 \pm \scriptstyle{0.007}$ &
        $0.463 \pm \scriptstyle{0.010}$ &
        $0.316 \pm \scriptstyle{0.007}$ \\
        SurgFUTR-S & State-Change & 
        $\textbf{0.185} \pm \scriptstyle{0.006}$ &
        $0.274 \pm \scriptstyle{0.008}$ &
        $0.465 \pm \scriptstyle{0.009}$ &
        $0.308 \pm \scriptstyle{0.008}$ \\
        \rowcolor{cyan!15}
        SurgFUTR-TS & State-Change & 
        $\textbf{0.185} \pm \scriptstyle{0.010}$ &
        $\textbf{0.269} \pm \scriptstyle{0.006}$ &
        $\textbf{0.451} \pm \scriptstyle{0.003}$ &
        $\textbf{0.302} \pm \scriptstyle{0.005}$ \\
        \bottomrule 
    \end{tabular}
    }
\end{table}

\textbf{Transfer to MultiBypass140:}
Tables~\ref{tab:results:quantitative:phtr_only_multibypass140_stras} and~\ref{tab:results:quantitative:phtr_only_multibypass140_bern} present phase transition prediction transfer results across temporal horizons (2, 3, 5 minutes). SurgFUTR-TS achieves the best overall performance on Strasbourg (mean MAE: $0.142\pm0.002$) and ties with SurgFUTR-S on Bern ($0.195\pm0.001$), demonstrating robust cross-procedure transfer for workflow anticipation. Self-supervised baselines (MoCov2, DINO) show surprisingly strong transfer on Strasbourg (mean MAE: $0.157$/$0.156$), outperforming general vision pretraining strategies like Kinetics-400 ($0.222\pm0.030$) and Phase ($0.239\pm0.005$). However, on Bern, the performance hierarchy shifts: SurgFUTR variants maintain superior performance while self-supervised methods degrade to $0.242$/$0.240$ MAE. Surgical foundation models struggle significantly, with EndoViT achieving poor transfer on both centers ($0.332$ MAE) and SurgeNetXL showing center-dependent performance (Strasbourg: $0.161$ vs. Bern: $0.380$ MAE). 
Within SurgFUTR variants, TS and S achieve comparable performance, both outperforming Lite by $6.7\%$ on Strasbourg and $8.0\%$ on Bern.
\begin{table}[h]
    \centering
    \caption{Model transfer performance on MultiBypass140 (Center: Strasbourg): SFP-II (phase transition) by MAE; lower is better.}
    \label{tab:results:quantitative:phtr_only_multibypass140_stras}
    \setlength{\tabcolsep}{3pt}
    \resizebox{\columnwidth}{!}{%
    \begin{tabular}{@{}llcccc@{}}
    \toprule
        \multirow{2}{*}{\textbf{Method}} &
        \multirow{2}{*}{\textbf{Initialization}} &
        \multicolumn{4}{c}{\textbf{Phase Transition (Center: Strasbourg)}} \\
        \cmidrule(lr){3-6}
        & & \textbf{2 min} & \textbf{3 min} & \textbf{5 min} & \textbf{mean} \\
        \midrule
        \multicolumn{4}{l}{\textit{General SSL / Foundation Pretraining}} \\
        VideoMAEv2 & Random &
        $0.175 \pm \scriptstyle{0.001}$ &
        $0.253 \pm \scriptstyle{0.001}$ &
        $0.423 \pm \scriptstyle{0.001}$ &
        $0.284 \pm \scriptstyle{0.001}$ \\
        VideoMAEv2 & Kinetics-400 & 
        $0.133 \pm \scriptstyle{0.018}$ &
        $0.197 \pm \scriptstyle{0.027}$ &
        $0.339 \pm \scriptstyle{0.045}$ &
        $0.222 \pm \scriptstyle{0.030}$ \\
        VideoMAEv2 & Phase & 
        $0.141 \pm \scriptstyle{0.004}$ &
        $0.214 \pm \scriptstyle{0.006}$ &
        $0.363 \pm \scriptstyle{0.005}$ &
        $0.239 \pm \scriptstyle{0.005}$ \\
        MoCoV2 & Cholec80 & 
        $0.090 \pm \scriptstyle{0.000}$ &
        $0.138 \pm \scriptstyle{0.000}$ &
        $0.243 \pm \scriptstyle{0.002}$ &
        $0.157 \pm \scriptstyle{0.000}$ \\
        DINO & Cholec80 & 
        $0.090 \pm \scriptstyle{0.000}$ &
        $0.137 \pm \scriptstyle{0.000}$ &
        $0.241 \pm \scriptstyle{0.003}$ &
        $0.156 \pm \scriptstyle{0.001}$ \\
        EndoViT & Endo700k & 
        $0.202 \pm \scriptstyle{0.004}$ &
        $0.295 \pm \scriptstyle{0.005}$ &
        $0.499 \pm \scriptstyle{0.008}$ &
        $0.332 \pm \scriptstyle{0.006}$ \\
        SurgeNetXL & SurgeNetXL & 
        $0.093 \pm \scriptstyle{0.003}$ &
        $0.142 \pm \scriptstyle{0.005}$ &
        $0.248 \pm \scriptstyle{0.010}$ &
        $0.161 \pm \scriptstyle{0.006}$ \\

        \midrule
        \multicolumn{4}{l}{\textit{Task-Specific Baselines}} \\
        AVT & ImageNet-21K &
        $0.091 \pm \scriptstyle{0.006}$ &
        $0.136 \pm \scriptstyle{0.007}$ &
        $0.234 \pm \scriptstyle{0.009}$ &
        $0.154 \pm \scriptstyle{0.007}$ \\
        RULSTM & ImageNet-1K &
        $0.084 \pm \scriptstyle{0.001}$ &
        $0.128 \pm \scriptstyle{0.002}$ &
        $0.228 \pm \scriptstyle{0.003}$ &
        $0.146 \pm \scriptstyle{0.002}$ \\
        SF-RULSTM & ImageNet-1K &
        $0.138 \pm \scriptstyle{0.094}$ &
        $0.203 \pm \scriptstyle{0.128}$ &
        $0.346 \pm \scriptstyle{0.202}$ &
        $0.229 \pm \scriptstyle{0.142}$ \\
        
        \midrule
        \multicolumn{4}{l}{\textit{Our Method (State-Change Pretraining)}} \\
        SurgFUTR-Lite & State-Change & 
        $0.090 \pm \scriptstyle{0.002}$ &
        $0.135 \pm \scriptstyle{0.002}$ &
        $0.236 \pm \scriptstyle{0.004}$ &
        $0.153 \pm \scriptstyle{0.003}$ \\
        SurgFUTR-S & State-Change & 
        $0.085 \pm \scriptstyle{0.001}$ &
        $0.127 \pm \scriptstyle{0.001}$ &
        $\textbf{0.220} \pm \scriptstyle{0.001}$ &
        $0.144 \pm \scriptstyle{0.000}$ \\
        \rowcolor{cyan!15}
        SurgFUTR-TS & State-Change & 
        $\textbf{0.082} \pm \scriptstyle{0.001}$ &
        $\textbf{0.125} \pm \scriptstyle{0.003}$ &
        $\textbf{0.220} \pm \scriptstyle{0.003}$ &
        $\textbf{0.142} \pm \scriptstyle{0.002}$ \\
        \bottomrule 
    \end{tabular}
    }
\end{table}
The task-specific baselines further show that cross-center transfer affects anticipation models differently. On Strasbourg, RULSTM is the strongest task-specific baseline with a mean MAE of 0.146, followed closely by AVT at 0.154, while SF-RULSTM is notably weaker at 0.229 and exhibits substantially larger variance. On Bern, the ordering shifts: AVT becomes the strongest task-specific baseline at 0.211 mean MAE, whereas RULSTM degrades to 0.273 and SF-RULSTM further declines to 0.328. Overall, these results indicate that anticipation-based baselines remain relevant comparators for phase transition transfer, but their robustness to center-specific variation differs considerably, with AVT showing the most stable behavior across the two sites.

Notably, prediction difficulty increases with anticipation horizon across all methods, with 5-minute predictions showing $2.4\times$ higher MAE than 2-minute predictions on average, and Bern demonstrating higher absolute MAE values than Strasbourg across all horizons and methods. Unlike in the RSD setting, where Bern yielded lower absolute MAE, this pattern suggests that center-specific characteristics affect the two downstream tasks differently. In particular, while shorter average procedure duration in Bern may simplify remaining-time regression, phase transition prediction depends more strongly on the temporal structure and predictability of upcoming workflow changes within the anticipation horizon.

\begin{table}[h]
    \centering
    \caption{Model transfer performance on MultiBypass140 (Center: Bern): SFP-II (phase transition) by MAE; lower is better.}
    \label{tab:results:quantitative:phtr_only_multibypass140_bern}
    \setlength{\tabcolsep}{3pt}
    \resizebox{\columnwidth}{!}{%
    \begin{tabular}{@{}llcccc@{}}
    \toprule
        \multirow{2}{*}{\textbf{Method}} &
        \multirow{2}{*}{\textbf{Initialization}} &
        \multicolumn{4}{c}{\textbf{Phase Transition (Center: Bern)}} \\
        \cmidrule(lr){3-6}
        & & \textbf{2 min} & \textbf{3 min} & \textbf{5 min} & \textbf{mean} \\
        \midrule
        \multicolumn{4}{l}{\textit{General SSL / Foundation Pretraining}} \\
        VideoMAEv2 & Random &
        $0.182 \pm \scriptstyle{0.002}$ &
        $0.263 \pm \scriptstyle{0.003}$ &
        $0.438 \pm \scriptstyle{0.007}$ &
        $0.295 \pm \scriptstyle{0.004}$ \\
        VideoMAEv2 & Kinetics-400 & 
        $0.163 \pm \scriptstyle{0.010}$ &
        $0.239 \pm \scriptstyle{0.012}$ &
        $0.406 \pm \scriptstyle{0.026}$ &
        $0.269 \pm \scriptstyle{0.016}$ \\
        VideoMAEv2 & Phase & 
        $0.164 \pm \scriptstyle{0.000}$ &
        $0.244 \pm \scriptstyle{0.002}$ &
        $0.411 \pm \scriptstyle{0.008}$ &
        $0.273 \pm \scriptstyle{0.003}$ \\
        MoCoV2 & Cholec80 & 
        $0.141 \pm \scriptstyle{0.003}$ &
        $0.215 \pm \scriptstyle{0.007}$ &
        $0.370 \pm \scriptstyle{0.012}$ &
        $0.242 \pm \scriptstyle{0.007}$ \\
        DINO & Cholec80 & 
        $0.139 \pm \scriptstyle{0.002}$ &
        $0.212 \pm \scriptstyle{0.006}$ &
        $0.370 \pm \scriptstyle{0.010}$ &
        $0.240 \pm \scriptstyle{0.006}$ \\
        EndoViT & Endo700k & 
        $0.202 \pm \scriptstyle{0.003}$ &
        $0.296 \pm \scriptstyle{0.003}$ &
        $0.499 \pm \scriptstyle{0.004}$ &
        $0.332 \pm \scriptstyle{0.003}$ \\
        SurgeNetXL & SurgeNetXL & 
        $0.238 \pm \scriptstyle{0.001}$ &
        $0.344 \pm \scriptstyle{0.005}$ &
        $0.560 \pm \scriptstyle{0.003}$ &
        $0.380 \pm \scriptstyle{0.003}$ \\

        \midrule
        \multicolumn{4}{l}{\textit{Task-Specific Baselines}} \\
        AVT & ImageNet-21K &
        $0.130 \pm \scriptstyle{0.012}$ &
        $0.188 \pm \scriptstyle{0.012}$ &
        $0.316 \pm \scriptstyle{0.018}$ &
        $0.211 \pm \scriptstyle{0.014}$ \\
        RULSTM & ImageNet-1K &
        $0.171 \pm \scriptstyle{0.006}$ &
        $0.245 \pm \scriptstyle{0.008}$ &
        $0.402 \pm \scriptstyle{0.009}$ &
        $0.273 \pm \scriptstyle{0.007}$ \\
        SF-RULSTM & ImageNet-1K &
        $0.208 \pm \scriptstyle{0.053}$ &
        $0.295 \pm \scriptstyle{0.065}$ &
        $0.482 \pm \scriptstyle{0.093}$ &
        $0.328 \pm \scriptstyle{0.070}$ \\

        \midrule
        \multicolumn{4}{l}{\textit{Our Method (State-Change Pretraining)}} \\
        SurgFUTR-Lite & State-Change & 
        $0.132 \pm \scriptstyle{0.001}$ &
        $0.190 \pm \scriptstyle{0.001}$ &
        $0.312 \pm \scriptstyle{0.001}$ &
        $0.212 \pm \scriptstyle{0.000}$ \\
        SurgFUTR-S & State-Change & 
        $0.118 \pm \scriptstyle{0.005}$ &
        $\textbf{0.173} \pm \scriptstyle{0.007}$ &
        $\textbf{0.293} \pm \scriptstyle{0.012}$ &
        $0.195 \pm \scriptstyle{0.008}$ \\
        \rowcolor{cyan!15}
        SurgFUTR-TS & State-Change & 
        $\textbf{0.117} \pm \scriptstyle{0.000}$ &
        $\textbf{0.173} \pm \scriptstyle{0.001}$ &
        $0.295 \pm \scriptstyle{0.002}$ &
        $\textbf{0.195} \pm \scriptstyle{0.001}$ \\
        \bottomrule 
    \end{tabular}
    }
\end{table}

Contextualizing these MultiBypass140 phase transition results in absolute terms provides a clearer sense of their magnitude. On Strasbourg center, the mean MAE improvement of SurgFUTR-TS over RULSTM (0.146 to 0.142) corresponds to approximately 0.24 seconds, over AVT (0.154 to 0.142) to approximately 0.72 seconds, and over DINO (0.156 to 0.142) to approximately 0.84 seconds. On Bern center, the improvement over AVT (0.211 to 0.195) corresponds to approximately 0.96 seconds, and over DINO (0.240 to 0.195) to approximately 2.70 seconds, while remaining essentially on par with SurgFUTR-S in mean MAE. Overall, the absolute gains on MultiBypass140 are modest but consistent, and are larger on Bern than on Strasbourg.

\paragraph{\textbf{SFP-III: Step Transition Prediction in GraSP}}
We analyze future prediction capability on step transition prediction, which enables surgical AI anticipation at finer granularity compared to coarse phase-level predictions. As GraSP~\citep{ayobi2025pixel} provides step annotations, we investigate how state-change pretraining impacts step transition forecasting.

Table~\ref{tab:results:quantitative:step_only_grasp} reports model performance in mean absolute error (MAE) for step transition prediction on GraSP.

\textbf{Baseline Performance:} 
Among the general SSL and foundation-model baselines, surgical domain-adapted SSL models achieve the strongest performance, with MoCoV2 (0.274 MAE) and DINO (0.278 MAE) leading this group.
Interestingly, EndoViT (0.282 MAE) and SurgeNetXL (0.283 MAE) show competitive results despite their larger scale, performing within 2.9\% and 3.3\% of the best SSL baseline. VideoMAEv2 variants show mixed performance, with Random initialization (0.286 MAE) and Kinetics-400 (0.338 MAE) spanning a wide 23.4\% range. Overall, baseline methods cluster within a narrow 0.064 MAE range (0.274--0.338 MAE), suggesting that step transition prediction poses relatively uniform difficulty across different pretraining strategies.

\begin{table}[h]
    \centering
    \caption{Model transfer performance on GraSP: SFP-III (step transition) by MAE; lower is better.}
    \label{tab:results:quantitative:step_only_grasp}
    \setlength{\tabcolsep}{3pt}
    \resizebox{\columnwidth}{!}{%
    \begin{tabular}{@{}llcccc@{}}
    \toprule
        \multirow{2}{*}{\textbf{Method}} &
        \multirow{2}{*}{\textbf{Initialization}} &
        \multicolumn{4}{c}{\textbf{Step Transition}} \\
        \cmidrule(lr){3-6}
        & & \textbf{2 min} & \textbf{3 min} & \textbf{5 min} & \textbf{mean} \\
        \midrule
        \multicolumn{4}{l}{\textit{General SSL / Foundation Pretraining}} \\
        VideoMAEv2 & Random &
        $0.182 \pm \scriptstyle{0.002}$ &
        $0.258 \pm \scriptstyle{0.003}$ &
        $0.418 \pm \scriptstyle{0.005}$ &
        $0.286 \pm \scriptstyle{0.003}$ \\
        VideoMAEv2 & Kinetics-400 & 
        $0.208 \pm \scriptstyle{0.011}$ &
        $0.305 \pm \scriptstyle{0.008}$ &
        $0.500 \pm \scriptstyle{0.022}$ &
        $0.338 \pm \scriptstyle{0.013}$ \\
        MoCoV2 & Cholec80 & 
        $0.164 \pm \scriptstyle{0.001}$ &
        $0.244 \pm \scriptstyle{0.001}$ &
        $0.414 \pm \scriptstyle{0.001}$ &
        $0.274 \pm \scriptstyle{0.001}$ \\
        DINO & Cholec80 & 
        $0.166 \pm \scriptstyle{0.001}$ &
        $0.248 \pm \scriptstyle{0.003}$ &
        $0.419 \pm \scriptstyle{0.005}$ &
        $0.278 \pm \scriptstyle{0.003}$ \\
        EndoViT & Endo700k & 
        $0.173 \pm \scriptstyle{0.002}$ &
        $0.253 \pm \scriptstyle{0.002}$ &
        $0.422 \pm \scriptstyle{0.003}$ &
        $0.282 \pm \scriptstyle{0.002}$ \\
        SurgeNetXL & SurgeNetXL & 
        $0.176 \pm \scriptstyle{0.001}$ &
        $0.254 \pm \scriptstyle{0.002}$ &
        $0.419 \pm \scriptstyle{0.003}$ &
        $0.283 \pm \scriptstyle{0.002}$ \\

        \midrule
        \multicolumn{4}{l}{\textit{Task-Specific Baselines}} \\
        AVT & ImageNet-21K &
        $0.162 \pm \scriptstyle{0.001}$ &
        $0.234 \pm \scriptstyle{0.001}$ &
        $0.384 \pm \scriptstyle{0.006}$ &
        $0.260 \pm \scriptstyle{0.002}$ \\
        RULSTM & ImageNet-1K &
        $0.157 \pm \scriptstyle{0.005}$ &
        $0.237 \pm \scriptstyle{0.005}$ &
        $0.406 \pm \scriptstyle{0.010}$ &
        $0.267 \pm \scriptstyle{0.007}$ \\
        SF-RULSTM & ImageNet-1K &
        $0.165 \pm \scriptstyle{0.005}$ &
        $0.243 \pm \scriptstyle{0.005}$ &
        $0.410 \pm \scriptstyle{0.009}$ &
        $0.273 \pm \scriptstyle{0.005}$ \\
        
        \midrule
        \multicolumn{4}{l}{\textit{Our Method (State-Change Pretraining)}} \\
        SurgFUTR-Lite & State-Change & 
        $0.167 \pm \scriptstyle{0.003}$ &
        $0.236 \pm \scriptstyle{0.004}$ &
        $0.383 \pm \scriptstyle{0.005}$ &
        $0.262 \pm \scriptstyle{0.004}$ \\
        SurgFUTR-S & State-Change & 
        $0.155 \pm \scriptstyle{0.012}$ &
        $0.228 \pm \scriptstyle{0.015}$ &
        $0.384 \pm \scriptstyle{0.022}$ &
        $0.256 \pm \scriptstyle{0.016}$ \\
        \rowcolor{cyan!15}
        SurgFUTR-TS & State-Change & 
        $\textbf{0.149} \pm \scriptstyle{0.001}$ &
        $\textbf{0.222} \pm \scriptstyle{0.002}$ &
        $\textbf{0.377} \pm \scriptstyle{0.003}$ &
        $\textbf{0.249} \pm \scriptstyle{0.002}$ \\
        \bottomrule 
    \end{tabular}
    }
\end{table}

\textbf{Initialization Effects:} 
Step transition prediction reveals unexpected initialization patterns for VideoMAEv2. Random initialization (0.286 MAE) outperforms Kinetics-400 initialization (0.338 MAE) by 15.4\% (0.052 MAE), a counterintuitive result suggesting that general video pretraining on natural scenes introduces domain biases that hinder fine-grained surgical step prediction, where surgical-specific temporal patterns dominate. In contrast, surgical domain-adapted methods (MoCoV2, DINO with Cholec80) demonstrate the critical importance of surgical-specific pretraining, outperforming Kinetics-400 by at least 17.8\% (0.060 MAE). This finding highlights a key limitation of general vision foundation models for granular surgical workflow understanding.

\textbf{Task-Specific Baselines:}
The task-specific baselines remain competitive for step transition prediction and provide a stronger comparison than most general pretraining strategies. AVT achieves the best task-specific performance with a mean MAE of 0.260, followed closely by RULSTM at 0.267 and SF-RULSTM at 0.273. Compared with the strongest general baseline, MoCoV2 at 0.274, AVT yields a modest improvement while RULSTM remains in a similar range. This indicates that task-specific anticipation training is beneficial at the finer step-transition level, although the gap over strong surgical SSL pretraining is smaller than in the phase transition setting. The comparatively narrow spread among AVT, RULSTM, SF-RULSTM, and the best SSL baselines further suggests that fine-grained step forecasting is a challenging regime in which both anticipation-specific modeling and surgical-domain pretraining contribute meaningfully.


\textbf{SurgFUTR Performance:} 
State-change pretrained variants demonstrate clear and consistent improvements across all baselines. SurgFUTR-TS delivers the strongest performance (0.249 MAE), achieving 9.1\% improvement over the best general baseline MoCoV2 (0.274 MAE) and 26.3\% improvement over Kinetics-400 VideoMAEv2 (0.338 MAE), while also improving over the strongest task-specific baseline AVT (0.260 MAE). SurgFUTR-S (0.256 MAE) shows intermediate performance with 6.6\% improvement over MoCoV2 and also remains slightly ahead of AVT, while SurgFUTR-Lite (0.262 MAE) achieves 4.4\% improvement but remains marginally below AVT. Critically, performance gains are consistent across all prediction horizons: at 5 minutes, SurgFUTR-TS achieves 0.377 MAE vs. 0.414 MAE for MoCoV2 (8.9\% improvement), demonstrating that state-change pretraining captures fine-grained temporal dependencies essential for long-term step transition forecasting. The progressive improvement from Lite to S to TS (4.4\% → 6.6\% → 9.1\%) validates that each component of our teacher-student framework contributes meaningfully to fine-grained surgical workflow anticipation beyond both general SSL baselines and task-specific anticipation baselines.

Looking at the results in absolute terms, SurgFUTR-TS also maintains a consistent advantage on GraSP step transition prediction. Using the mean MAE across the 2, 3, and 5-minute horizons, the improvement over AVT (0.260 to 0.249) corresponds to approximately 0.66 seconds, the improvement over RULSTM (0.267 to 0.249) to approximately 1.08 seconds, and the improvement over the strongest general baseline MoCoV2 (0.274 to 0.249) to approximately 1.50 seconds. This indicates that the benefit of state-change pretraining extends to step transition prediction as well.

\textbf{Transfer to MultiBypass140:}
Tables~\ref{tab:results:quantitative:step_only_multibypass140_stras} and~\ref{tab:results:quantitative:step_only_multibypass140_bern} present step transition prediction transfer results across temporal horizons (2, 3, 5 minutes). SurgFUTR variants achieve better cross-procedure transfer performance, with SurgFUTR-S leading on Strasbourg (0.073 MAE) and SurgFUTR-TS achieving best results on Bern (0.110 MAE). Notably, SurgFUTR-S (student-only with state-change pretraining) slightly outperforms the full teacher-student framework (TS) on Strasbourg, suggesting that for fine-grained step transitions, the core state representation learning through Sinkhorn clustering may be more critical than explicit future prediction modeling. 
\begin{table}[h]
    \centering
    \caption{Model transfer performance on MultiBypass140 (Center: Strasbourg): SFP-III (step transition) by MAE; lower is better.}
    \label{tab:results:quantitative:step_only_multibypass140_stras}
    \setlength{\tabcolsep}{3pt}
    \resizebox{\columnwidth}{!}{%
    \begin{tabular}{@{}llcccc@{}}
    \toprule
        \multirow{2}{*}{\textbf{Method}} &
        \multirow{2}{*}{\textbf{Initialization}} &
        \multicolumn{4}{c}{\textbf{Step Transition (Center: Strasbourg)}} \\
        \cmidrule(lr){3-6}
        & & \textbf{2 min} & \textbf{3 min} & \textbf{5 min} & \textbf{mean} \\
        \midrule
        \multicolumn{4}{l}{\textit{General SSL / Foundation Pretraining}} \\
        VideoMAEv2 & Random &
        $0.080 \pm \scriptstyle{0.002}$ &
        $0.121 \pm \scriptstyle{0.004}$ &
        $0.163 \pm \scriptstyle{0.004}$ &
        $0.121 \pm \scriptstyle{0.003}$ \\
        VideoMAEv2 & Kinetics-400 & 
        $0.116 \pm \scriptstyle{0.010}$ &
        $0.171 \pm \scriptstyle{0.012}$ &
        $0.251 \pm \scriptstyle{0.002}$ &
        $0.180 \pm \scriptstyle{0.006}$ \\
        VideoMAEv2 & Phase & 
        $0.123 \pm \scriptstyle{0.018}$ &
        $0.171 \pm \scriptstyle{0.014}$ &
        $0.252 \pm \scriptstyle{0.000}$ &
        $0.182 \pm \scriptstyle{0.010}$ \\
        MoCoV2 & Cholec80 & 
        $0.050 \pm \scriptstyle{0.003}$ &
        $0.074 \pm \scriptstyle{0.004}$ &
        $0.117 \pm \scriptstyle{0.006}$ &
        $0.081 \pm \scriptstyle{0.004}$ \\
        DINO & Cholec80 & 
        $0.051 \pm \scriptstyle{0.003}$ &
        $0.075 \pm \scriptstyle{0.004}$ &
        $0.119 \pm \scriptstyle{0.007}$ &
        $0.082 \pm \scriptstyle{0.004}$ \\
        EndoViT & Endo700k & 
        $0.079 \pm \scriptstyle{0.001}$ &
        $0.121 \pm \scriptstyle{0.002}$ &
        $0.167 \pm \scriptstyle{0.004}$ &
        $0.122 \pm \scriptstyle{0.003}$ \\
        SurgeNetXL & SurgeNetXL & 
        $0.061 \pm \scriptstyle{0.005}$ &
        $0.093 \pm \scriptstyle{0.008}$ &
        $0.130 \pm \scriptstyle{0.010}$ &
        $0.095 \pm \scriptstyle{0.008}$ \\

        \midrule
        \multicolumn{4}{l}{\textit{Task-Specific Baselines}} \\
        AVT & ImageNet-21K &
        $0.053 \pm \scriptstyle{0.001}$ &
        $0.075 \pm \scriptstyle{0.002}$ &
        $0.115 \pm \scriptstyle{0.003}$ &
        $0.081 \pm \scriptstyle{0.002}$ \\
        RULSTM & ImageNet-1K &
        $0.045 \pm \scriptstyle{0.001}$ &
        $0.068 \pm \scriptstyle{0.001}$ &
        $0.109 \pm \scriptstyle{0.000}$ &
        $0.074 \pm \scriptstyle{0.000}$ \\
        SF-RULSTM & ImageNet-1K &
        $0.080 \pm \scriptstyle{0.030}$ &
        $0.119 \pm \scriptstyle{0.044}$ &
        $0.162 \pm \scriptstyle{0.049}$ &
        $0.120 \pm \scriptstyle{0.041}$ \\

        \midrule
        \multicolumn{4}{l}{\textit{Our Method (State-Change Pretraining)}} \\
        SurgFUTR-Lite & State-Change & 
        $0.055 \pm \scriptstyle{0.000}$ &
        $0.079 \pm \scriptstyle{0.000}$ &
        $0.115 \pm \scriptstyle{0.000}$ &
        $0.083 \pm \scriptstyle{0.000}$ \\
        \rowcolor{cyan!15}
        SurgFUTR-S & State-Change & 
        $\textbf{0.047} \pm \scriptstyle{0.001}$ &
        $\textbf{0.068} \pm \scriptstyle{0.001}$ &
        $\textbf{0.104} \pm \scriptstyle{0.002}$ &
        $\textbf{0.073} \pm \scriptstyle{0.001}$ \\
        SurgFUTR-TS & State-Change & 
        $0.048 \pm \scriptstyle{0.000}$ &
        $0.070 \pm \scriptstyle{0.001}$ &
        $0.106 \pm \scriptstyle{0.001}$ &
        $0.075 \pm \scriptstyle{0.001}$ \\
        \bottomrule 
    \end{tabular}
    }
\end{table}

Self-supervised baselines demonstrate strong transfer capabilities: MoCoV2 (0.081/0.132 MAE) and DINO (0.082/0.133 MAE) outperform all other baselines on both centers. Surprisingly, VideoMAEv2 with random initialization achieves competitive performance (0.121/0.132 MAE), significantly outperforming Kinetics-400 (0.180/0.196 MAE) and Phase (0.182/0.179 MAE) initialization by 32.8\%/32.7\% and 33.5\%/26.3\% respectively. This counterintuitive pattern reinforces findings from in-domain step prediction: general video pretraining introduces biases detrimental to fine-grained surgical workflow understanding. 
\begin{table}[h]
    \centering
    \caption{Model transfer performance on MultiBypass140 (Center: Bern): SFP-III (step transition) by MAE; lower is better.}
    \label{tab:results:quantitative:step_only_multibypass140_bern}
    \setlength{\tabcolsep}{3pt}
    \resizebox{\columnwidth}{!}{%
    \begin{tabular}{@{}llcccc@{}}
    \toprule
        \multirow{2}{*}{\textbf{Method}} &
        \multirow{2}{*}{\textbf{Initialization}} &
        \multicolumn{4}{c}{\textbf{Step Transition (Center: Bern)}} \\
        \cmidrule(lr){3-6}
        & & \textbf{2 min} & \textbf{3 min} & \textbf{5 min} & \textbf{mean} \\
        \midrule
        \multicolumn{4}{l}{\textit{General SSL / Foundation Pretraining}} \\
        VideoMAEv2 & Random &
        $0.087 \pm \scriptstyle{0.001}$ &
        $0.144 \pm \scriptstyle{0.001}$ &
        $0.165 \pm \scriptstyle{0.001}$ &
        $0.132 \pm \scriptstyle{0.001}$ \\
        VideoMAEv2 & Kinetics-400 & 
        $0.126 \pm \scriptstyle{0.000}$ &
        $0.198 \pm \scriptstyle{0.004}$ &
        $0.263 \pm \scriptstyle{0.005}$ &
        $0.196 \pm \scriptstyle{0.003}$ \\
        VideoMAEv2 & Phase & 
        $0.124 \pm \scriptstyle{0.002}$ &
        $0.181 \pm \scriptstyle{0.002}$ &
        $0.231 \pm \scriptstyle{0.004}$ &
        $0.179 \pm \scriptstyle{0.003}$ \\
        MoCoV2 & Cholec80 & 
        $0.083 \pm \scriptstyle{0.007}$ &
        $0.140 \pm \scriptstyle{0.010}$ &
        $0.173 \pm \scriptstyle{0.013}$ &
        $0.132 \pm \scriptstyle{0.010}$ \\
        DINO & Cholec80 & 
        $0.085 \pm \scriptstyle{0.004}$ &
        $0.141 \pm \scriptstyle{0.010}$ &
        $0.172 \pm \scriptstyle{0.016}$ &
        $0.133 \pm \scriptstyle{0.009}$ \\
        EndoViT & Endo700k & 
        $0.084 \pm \scriptstyle{0.000}$ &
        $0.144 \pm \scriptstyle{0.000}$ &
        $0.167 \pm \scriptstyle{0.000}$ &
        $0.132 \pm \scriptstyle{0.000}$ \\
        SurgeNetXL & SurgeNetXL & 
        $0.083 \pm \scriptstyle{0.001}$ &
        $0.142 \pm \scriptstyle{0.000}$ &
        $0.165 \pm \scriptstyle{0.002}$ &
        $0.130 \pm \scriptstyle{0.001}$ \\

        \midrule
        \multicolumn{4}{l}{\textit{Task-Specific Baselines}} \\
        AVT & ImageNet-21K &
        $0.082 \pm \scriptstyle{0.002}$ &
        $0.130 \pm \scriptstyle{0.002}$ &
        $0.158 \pm \scriptstyle{0.003}$ &
        $0.124 \pm \scriptstyle{0.002}$ \\
        RULSTM & ImageNet-1K &
        $0.101 \pm \scriptstyle{0.011}$ &
        $0.160 \pm \scriptstyle{0.012}$ &
        $0.189 \pm \scriptstyle{0.009}$ &
        $0.150 \pm \scriptstyle{0.011}$ \\
        SF-RULSTM & ImageNet-1K &
        $0.101 \pm \scriptstyle{0.009}$ &
        $0.160 \pm \scriptstyle{0.011}$ &
        $0.192 \pm \scriptstyle{0.010}$ &
        $0.151 \pm \scriptstyle{0.010}$ \\
        
        \midrule
        \multicolumn{4}{l}{\textit{Our Method (State-Change Pretraining)}} \\
        SurgFUTR-Lite & State-Change & 
        $0.079 \pm \scriptstyle{0.000}$ &
        $0.127 \pm \scriptstyle{0.000}$ &
        $0.152 \pm \scriptstyle{0.000}$ &
        $0.119 \pm \scriptstyle{0.000}$ \\
        SurgFUTR-S & State-Change & 
        $0.070 \pm \scriptstyle{0.001}$ &
        $\textbf{0.117} \pm \scriptstyle{0.001}$ &
        $\textbf{0.144} \pm \scriptstyle{0.001}$ &
        $0.111 \pm \scriptstyle{0.001}$ \\
        \rowcolor{cyan!15}
        SurgFUTR-TS & State-Change & 
        $\textbf{0.070} \pm \scriptstyle{0.000}$ &
        $\textbf{0.117} \pm \scriptstyle{0.001}$ &
        $\textbf{0.144} \pm \scriptstyle{0.001}$ &
        $\textbf{0.110} \pm \scriptstyle{0.000}$ \\
        \bottomrule 
    \end{tabular}
    }
\end{table}

The task-specific baselines show a pronounced center-dependent shift in relative performance. On the Strasbourg center, RULSTM is the strongest task-specific baseline with a mean MAE of 0.074, outperforming AVT (0.081) and substantially surpassing SF-RULSTM (0.120). In contrast, on the Bern center, AVT becomes the strongest task-specific baseline at 0.124 mean MAE, while both RULSTM and SF-RULSTM degrade to 0.150 and 0.151, respectively. This reversal mirrors the behavior observed for phase transition transfer and suggests that the recurrent anticipation baselines are more sensitive to center-specific temporal structure, whereas AVT provides more stable task-specific transfer across sites.

Surgical foundation models show mixed transfer: EndoViT performs poorly (0.122/0.132 MAE), while SurgeNetXL demonstrates better adaptation (0.095/0.130 MAE). SurgFUTR variants achieve improvements over the best general and task-specific baselines: SurgFUTR-S outperforms the strongest task-specific baseline RULSTM on Strasbourg (0.074 MAE) by a small margin and the strongest task-specific baseline AVT on Bern (0.124 MAE), while also improving over MoCoV2 by 9.9\%/15.9\% on Strasbourg/Bern, while SurgFUTR-TS achieves modest gains over the strongest task-specific baselines on both centers, in addition to 7.4\%/16.7\% improvements respectively over MoCoV2. The results suggest that state-change pretraining combined with clustering-based state representations provides effective temporal cues for step transition prediction across both medical centers.

In absolute terms, the transfer results on MultiBypass140 step transition prediction remain encouraging across both centers. At Strasbourg, SurgFUTR-S achieves the lowest mean MAE, improving over RULSTM from 0.074 to 0.073 (about 0.06 seconds) and over both AVT and MoCoV2 from 0.081 to 0.073 (about 0.48 seconds). At Bern, SurgFUTR-TS attains the best mean MAE, reducing error relative to AVT from 0.124 to 0.110 (about 0.84 seconds) and relative to the strongest general baseline SurgeNetXL from 0.130 to 0.110 (about 1.20 seconds). These results show that the transfer benefit of state-change pretraining extends consistently across both centers.

\paragraph{\textbf{SFP-IV: Cystic-Triplet Anticipation in CholecT50}}
Short-term anticipation aims to foresee the next \(1\!-\!5\) seconds of surgical activity to enable timely assistance and safer decision-making (Figure~\ref{fig:task_overview}). For SFP-IV, we focus on anticipating action triplets that may deform critical structures during laparoscopic cholecystectomy, specifically targeting cystic duct and cystic artery interactions listed in Table~\ref{tab:cystic_triplets}. We evaluate model performance using mAP, F1, and accuracy metrics averaged across \(1\!-\!5\) second prediction horizons, with results shown in Table~\ref{tab:results:quantitative:cyantr_results}.

\begin{table}[h]
    \centering
    \caption{Model performance comparison on CholecT50: SFP-IV (cystic triplet anticipation) by mAP, F1, and Acc; higher is better.}
    \label{tab:results:quantitative:cyantr_results}
    \setlength{\tabcolsep}{8pt}
    \resizebox{\columnwidth}{!}{%
    \begin{tabular}{llcccc}
        \toprule
        {\textbf{Method}} & {\textbf{Initialization}} & \textbf{mAP (\%)} & \textbf{F1 (\%)} & \textbf{Acc (\%)} \\
        \midrule
        \multicolumn{4}{l}{\textit{General SSL / Foundation Pretraining}} \\
        VideoMAEv2 & Random & $26.28$ & $6.55$ & $14.29$ \\
        VideoMAEv2 & Kinetics-400    & $37.49$ & $29.95$ & $32.61$ \\
        VideoMAEv2 & Phase   & $41.98$ & $29.67$ & $34.38$ \\
        MoCoV2     & Cholec80 & $40.54$ & $37.39$ & $36.37$ \\
        DINO       & Cholec80 & $43.97$ & $35.90$ & $38.91$ \\
        EndoViT    & Endo700k & $19.58$ & $2.58$ & $14.29$ \\
        SurgeNetXL & SurgeNetXL & $22.62$  & $6.55$ & $14.29$ \\

        \midrule
        \multicolumn{4}{l}{\textit{Task-Specific Baselines}} \\
        AVT & ImageNet-21K & $44.50$ & $26.77$ & $35.46$ \\
        RULSTM & ImageNet-1K & $45.48$ & $41.41$ & $43.60$ \\
        SF-RULSTM & ImageNet-1K & $39.96$ & $34.96$ & $36.41$ \\
        
        \midrule
        \multicolumn{4}{l}{\textit{Our Method (State-Change Pretraining)}} \\
        SurgFUTR-Lite & State-Change & $44.57$ & $36.08$ & $39.33$ \\
        SurgFUTR-S    & State-Change & $40.76$ & $29.27$ & $33.70$ \\
        \rowcolor{cyan!15}
        SurgFUTR-TS   & State-Change & $\textbf{50.29}$ & $\textbf{43.91}$ & $\textbf{45.61}$ \\
        \bottomrule 
    \end{tabular}
    }
\end{table}

\textbf{Baseline Performance:} ResNet50-based SSL methods achieve the strongest baseline performance, with DINO reaching 43.97\% mAP and 38.91\% Acc, while MoCoV2 delivers competitive performance (40.54\% mAP) with the highest baseline F1-Score (37.39\%). VideoMAEv2 with Phase pretraining (41.98\% mAP, 29.67\% F1, 34.38\% Acc) shows strong results, approaching SSL method performance. In stark contrast, despite their larger scale and extensive surgical pretraining, EndoViT (19.58\% mAP) and SurgeNetXL (22.62\% mAP) dramatically underperform, achieving results comparable to or worse than random initialization, highlighting significant transfer limitations for anticipation tasks.

\textbf{Initialization Effects:} VideoMAEv2 shows substantial performance differences across initialization strategies. \textit{Random} initialization yields poor results (26.28\% mAP, 6.55\% F1, 14.29\% Acc), while \textit{Kinetics-400} achieves 37.49\% mAP, 29.95\% F1, and 32.61\% Acc—representing substantial improvements of 11.21pp, 23.40pp, and 18.32pp respectively. \textit{Phase} pretraining provides additional gains over Kinetics-400 with 4.49pp mAP improvement (41.98\%), demonstrating the value of surgical domain knowledge for anticipation tasks where understanding surgical workflow is critical.

\textbf{Task-Specific Baselines:} The task-specific baselines provide a clearer picture for short-term cystic-triplet anticipation. RULSTM now emerges as the strongest task-specific baseline across all three aggregate metrics, achieving 45.48\% mAP, 41.41\% F1, and 43.60\% accuracy, while AVT remains competitive in mAP at 44.50\% and SF-RULSTM trails behind at 39.96\% mAP, 34.96\% F1, and 36.41\% accuracy. Unlike the earlier trend where different task-specific baselines favored different metrics, the updated results show that RULSTM provides the most balanced performance within this baseline family. This suggests that recurrent anticipation-based temporal modeling is a particularly strong comparator for fine-grained short-term cystic-triplet forecasting, making the gains of SurgFUTR-TS more meaningful in this setting.


\textbf{SurgFUTR Performance:} Our state-change pretrained variants achieve strong performance, with SurgFUTR-TS demonstrating clear superiority across all metrics. SurgFUTR-Lite (44.57\% mAP, 36.08\% F1, 39.33\% Acc) outperforms all general baselines in mAP and Acc, with 0.60pp mAP improvement over DINO (43.97\%), though it remains slightly below MoCoV2 in F1-Score (36.08\% vs 37.39\%) and below RULSTM in all three aggregate metrics (45.48\% mAP, 41.41\% F1, 43.60\% Acc). SurgFUTR-S (40.76\% mAP, 29.27\% F1, 33.70\% Acc) shows competitive performance but falls short of both SurgFUTR-Lite and top baselines across all metrics. SurgFUTR-TS delivers the best results with 50.29\% mAP, 43.91\% F1, and 45.61\% Acc, achieving substantial improvements of 6.32pp mAP over DINO, 6.52pp F1 over MoCoV2, and 6.70pp Acc over DINO, while also improving over the strongest task-specific baseline RULSTM by 4.81pp in mAP, 2.50pp in F1, and 2.01pp in accuracy—surpassing all baselines across every metric.
This represents the largest performance gains observed across all evaluated tasks, demonstrating particularly strong benefits of state-change pretraining for short-term surgical anticipation where temporal state modeling is crucial.

\begin{figure}[!ht]
\centering
    \includegraphics[width=1.00\columnwidth]{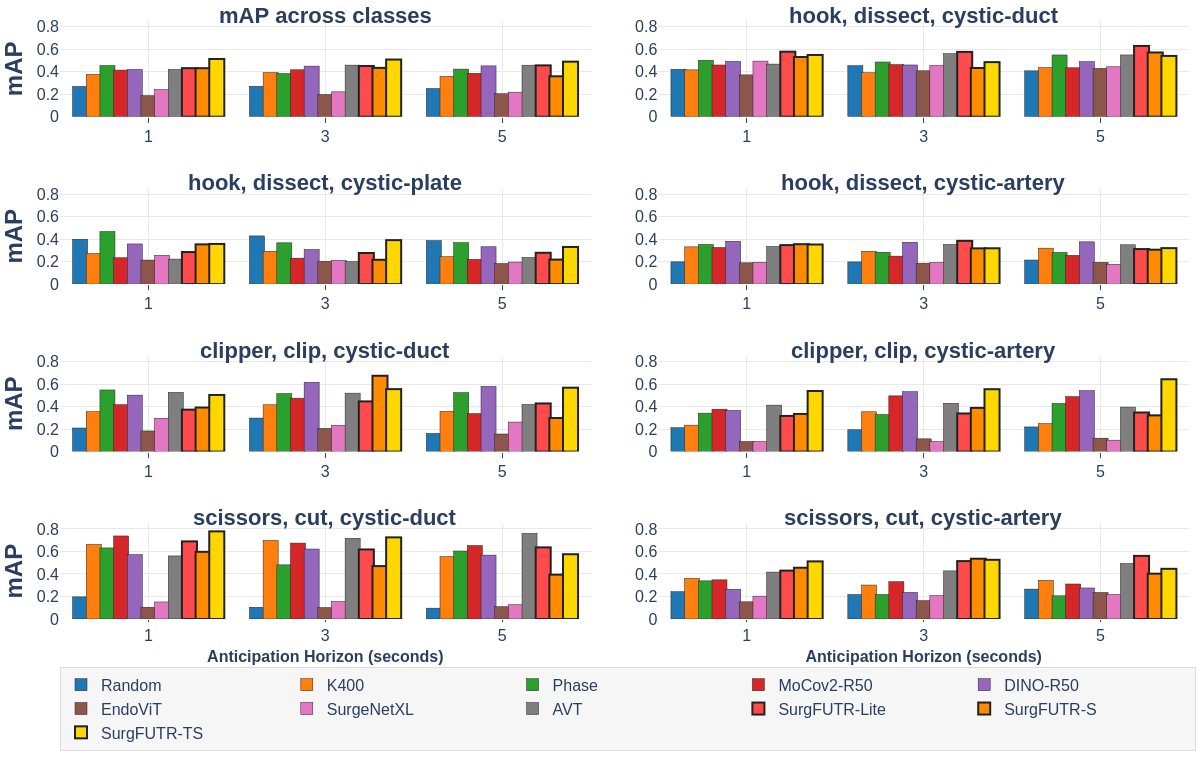} 
    \caption{Performance on cystic structure-related surgical action triplets in CholecT50 across baselines and SurgFUTR variants.}
    \label{fig:cyantr_detailed}
\end{figure}
Figure~\ref{fig:cyantr_detailed} further breaks down cystic-triplet anticipation on CholecT50 by triplet class and prediction horizon using AP. The class-wise trends are more heterogeneous than the aggregate results alone suggest. SurgFUTR-TS shows its clearest advantage on $\langle$\textit{clipper, clip, cystic-artery}$\rangle$, where it achieves the highest AP at all three horizons (0.54/0.55/0.64 at 1s/3s/5s). It is also strongest on $\langle$\textit{hook, dissect, cystic-plate}$\rangle$ at 3s (0.39), although Phase pretraining remains better on the same class at 1s and 5s (0.47 and 0.37, respectively). 
SurgFUTR-Lite is consistently strongest for $\langle$\textit{hook, dissect, cystic-duct}$\rangle$ across 1s, 3s, and 5s (0.58/0.58/0.63); notably, at the 1s horizon all SurgFUTR variants outperform the non-SurgFUTR baselines on this class, with Lite/TS/S reaching 0.58/0.55/0.53 AP versus 0.50 for the strongest competing baseline (Phase). SurgFUTR-Lite also gives the best 5s result for $\langle$\textit{scissors, cut, cystic-artery}$\rangle$ (0.56).
SurgFUTR-S is strongest for $\langle$\textit{clipper, clip, cystic-duct}$\rangle$ at 3s (0.67) and for $\langle$\textit{scissors, cut, cystic-artery}$\rangle$ at 3s (0.54). At the same time, several triplets are better captured by other baselines: SF-RULSTM is strongest for $\langle$\textit{hook, dissect, cystic-artery}$\rangle$ at all three horizons (0.39/0.37/0.43), RULSTM is strongest for $\langle$\textit{scissors, cut, cystic-duct}$\rangle$ at 1s and 3s (0.83/0.78), and AVT or SF-RULSTM lead the same class at 5s (0.76/0.77). 

A plausible interpretation is that the benefit of state-change pretraining depends on whether the learned temporal state provides predictive information beyond the current visual appearance. This is most evident for triplets such as $\langle$\textit{clipper, clip, cystic-artery}$\rangle$, where SurgFUTR-TS is strongest across all horizons, and $\langle$\textit{hook, dissect, cystic-duct}$\rangle$, where SurgFUTR variants remain consistently strong and all three outperform the non-SurgFUTR baselines at 1s. At the same time, the class-wise breakdown shows that strong anticipation baselines remain advantageous for some other triplets: for example, SF-RULSTM is strongest for $\langle$\textit{hook, dissect, cystic-artery}$\rangle$, while RULSTM performs particularly well for $\langle$\textit{scissors, cut, cystic-duct}$\rangle$. This suggests that the advantage of state-change pretraining is not uniform across all triplets, but is most pronounced when the future label depends on temporal state evolution rather than primarily on immediately visible local cues. The horizon-dependent variation further supports this view, indicating that the contribution of the learned state representation differs across both triplet type and anticipation interval.

\begin{table}[h]
    \centering
    \caption{Model performance comparison on CholecTrack20: SFP-V (event anticipation) by mAP, F1, and Acc; higher is better.}
    \label{tab:results:quantitative:ct3es_results}
    \setlength{\tabcolsep}{8pt}
    \resizebox{\columnwidth}{!}{%
    \begin{tabular}{llcccc}
        \toprule
        {\textbf{Method}} & {\textbf{Initialization}} & \textbf{mAP (\%)} & \textbf{F1 (\%)} & \textbf{Acc (\%)} \\
        \midrule
        \multicolumn{4}{l}{\textit{General SSL / Foundation Pretraining}} \\
        VideoMAEv2 & Random & $34.61$ & $24.31$ & $33.33$ \\
        VideoMAEv2 & Kinetics-400    & $43.86$ & $32.78$ & $42.56$ \\
        VideoMAEv2 & Phase   & $44.13$ & $36.06$ & $44.38$ \\
        MoCoV2     & Cholec80 & $42.12$ & $37.98$ & $42.43$ \\
        DINO       & Cholec80 & $40.36$ & $37.74$ & $37.73$ \\
        EndoViT    & Endo700k & $34.21$ & $24.31$ & $33.33$ \\
        SurgeNetXL & SurgeNetXL & $35.52$  & $24.31$ & $33.33$ \\

        \midrule
        \multicolumn{4}{l}{\textit{Task-Specific Baselines}} \\
        AVT & ImageNet-21K & $40.72$ & $29.38$ & $33.48$ \\
        RULSTM & ImageNet-1K & $38.32$ & $35.60$ & $35.64$ \\
        SF-RULSTM & ImageNet-1K & $43.02$ & $39.18$ & $42.86$ \\
        
        \midrule
        \multicolumn{4}{l}{\textit{Our Method (State-Change Pretraining)}} \\
        SurgFUTR-Lite & State-Change & $43.04$ & $35.83$ & $40.25$ \\
        SurgFUTR-S    & State-Change & $42.09$ & $29.82$ & $43.18$ \\
        \rowcolor{cyan!15}
        SurgFUTR-TS   & State-Change & $\textbf{46.12}$ & $\textbf{41.40}$ & $\textbf{46.48}$ \\
        \bottomrule 
    \end{tabular}
    }
\end{table}

\paragraph{\textbf{SFP-V: Cholec Event Anticipation in CholecTrack20}}
CholecTrack20~\citep{nwoye2023cholectrack20} focuses on anticipating surgical event such as bleeding and visual challenges (smoke, occlusion) within \(1\!-\!5\) second horizons during laparoscopic cholecystectomy. Table~\ref{tab:results:quantitative:ct3es_results} reports mAP, F1, and accuracy averaged across prediction horizons.

\textbf{Baseline Performance:} \textit{Phase} pretraining emerges as the strongest baseline (44.13\% mAP, 36.06\% F1, 44.38\% Acc), closely followed by Kinetics-400 (43.86\% mAP, 32.78\% F1, 42.56\% Acc). ResNet50-based SSL methods show solid performance, with MoCoV2 achieving the highest baseline F1-Score (37.98\%) alongside competitive mAP (42.12\%) and accuracy (42.43\%), while DINO reaches 40.36\% mAP and 37.74\% F1. EndoViT (34.21\% mAP) and SurgeNetXL (35.52\% mAP) again dramatically underperform, failing to exceed even \textit{Random} initialization results (34.61\% mAP), confirming their limited transfer capabilities for anticipation tasks.

\textbf{Initialization Effects:} VideoMAEv2 shows consistent benefits from domain-relevant initialization. \textit{Random} initialization achieves 34.61\% mAP, while \textit{Kinetics-400} provides substantial improvement to 43.86\% mAP (9.25pp gain). \textit{Phase} pretraining achieves the best VideoMAEv2 performance with 44.13\% mAP and notably strong accuracy (44.38\%), demonstrating surgical domain knowledge benefits for event anticipation.

\textbf{Task-Specific Baselines:} The task-specific baselines remain competitive for short-term event anticipation, but their relative strengths vary across evaluation metrics. SF-RULSTM attains the strongest overall task-specific performance, with 43.02\% mAP, 39.18\% F1, and 42.86\% accuracy, thereby outperforming both AVT and RULSTM on all three aggregate measures. AVT remains competitive in mAP at 40.72\%, while RULSTM achieves stronger F1 and accuracy than AVT despite its lower mAP. This suggests that temporal anticipation-specific architectures transfer effectively to event prediction, with the multi-scale formulation of SF-RULSTM offering a favorable inductive bias in this setting. At the same time, the best task-specific baseline does not surpass the strongest \textit{Phase}-initialized model, suggesting that phase-aware surgical pretraining remains highly effective for short-horizon event anticipation.


\textbf{SurgFUTR Performance:} Our variants show mixed results across different architectures. SurgFUTR-Lite achieves competitive performance (43.04\% mAP, 35.83\% F1, 40.25\% Acc) that approaches the best general baselines, falling just 1.09pp short of Phase in mAP and 2.15pp below MoCoV2 in F1 (37.98\%), while remaining comparable to the task-specific baselines and slightly exceeding SF-RULSTM in mAP. SurgFUTR-S shows degraded performance (42.09\% mAP, 29.82\% F1, 43.18\% Acc), with notably lower F1-Score indicating challenges for models trained only with current video clips without teacher-based distillation. SurgFUTR-TS delivers the best overall performance with 46.12\% mAP, 41.40\% F1, and 46.48\% Acc, achieving 1.99pp mAP improvement over the strongest general baseline (\textit{Phase}), 3.42pp F1 improvement over MoCoV2, and 2.10pp Acc improvement over Phase, while also surpassing the strongest task-specific baseline SF-RULSTM across all three metrics, demonstrating effective transfer of state-change representations to event anticipation tasks.

\begin{figure}[!ht]
\centering
    \includegraphics[width=1.00\columnwidth]{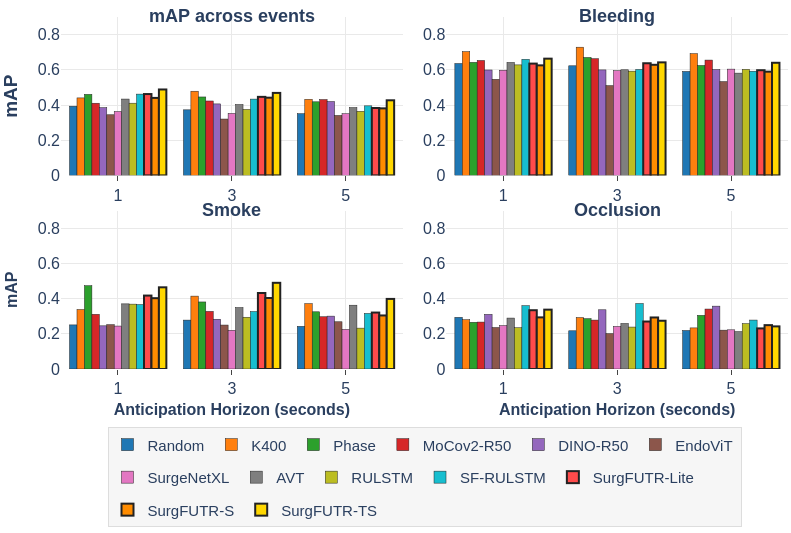} 
    \caption{Event-specific and averaged results in the CholecTrack20 dataset across all baselines and SurgFUTR variants.}
    \label{fig:ct3es_detailed}
\end{figure}

\begin{table*}[h]
    \centering
    \caption{Ablation on impact of loss components for SurgFUTR-TS model on CholecT50 dataset.}
    \label{tab:results:ablation:loss_fn}
    \setlength{\tabcolsep}{8pt}
    \resizebox{\textwidth}{!}{%
    \begin{tabular}{ccccccccccc}
        \toprule
        \multicolumn{3}{c}{\textbf{Components}} & 
        \multicolumn{4}{c}{\textbf{mAP (\%)}} & 
        \multicolumn{4}{c}{\textbf{F1-Score (\%)}} \\
        \cmidrule(lr){4-7} 
        \cmidrule(lr){8-11} 
        $\mathcal{L}^{distill}_{SC}$ & $\mathcal{L}^{distill}_{SE}$ & $\mathcal{L}_{FUTR}$ &
        $\textbf{Continuity}$ & $\textbf{Discontinuity}$ & $\textbf{Onset}$ & $\textbf{mean}$
        & $\textbf{Continuity}$ & $\textbf{Discontinuity}$ & $\textbf{Onset}$ & $\textbf{mean}$ \\
        \midrule
        &  &  $\checkmark$ & $76.48$ & $14.31$ & $13.74$ & $34.84$ & $68.45$ & $21.63$ & $19.78$ & $36.62$ \\
        & $\checkmark$ &  & $77.77$ & $12.71$ & $15.58$ & $35.35$ & $73.15$ & $20.28$ & $\textbf{23.72}$ & $39.05$ \\
        $\checkmark$ & &  & $77.49$ & $13.34$ & $16.18$ & $35.67$ & $\textbf{73.74}$ & $20.19$ & $23.37$ & $39.10$ \\
        $\checkmark$ & $\checkmark$ & & $77.86$ & $14.38$ & $16.38$ & $36.20$ & $71.88$ & $21.57$ & $23.06$ & $38.84$ \\
        $\checkmark$ &  & $\checkmark$ & $77.37$ & $14.43$ & $14.64$ & $35.48$ & $72.24$ & $21.04$ & $21.67$ & $38.31$ \\
        & $\checkmark$ & $\checkmark$ & $76.87$ & $13.93$ & $14.84$ & $35.21$ & $69.38$ & $19.61$ & $19.04$ & $36.01$ \\
        $\checkmark$ & $\checkmark$ & $\checkmark$ & $\textbf{78.77}$ & $\textbf{15.39}$ & $\textbf{16.62}$ & $\textbf{36.93}$ & $72.73$ & $\textbf{22.45}$ & $22.30$ & $\textbf{39.16}$ \\
        \bottomrule
    \end{tabular}
    }
\end{table*}

Figure~\ref{fig:ct3es_detailed} shows class-wise AP performance across CholecTrack20~\citep{nwoye2023cholectrack20} surgical events/visual challenges and anticipation horizons using AP. Performance patterns reveal task-specific strengths rather than universal dominance. For \textit{bleeding} detection, VideoMAEv2 with \textit{Kinetics-400} initialization leads across all horizons (0.70/0.73/0.69 at 1s/3s/5s vs. 0.66/0.64/0.64 for SurgFUTR-TS), with MoCov2-R50 also showing competitive performance (0.65/0.66/0.66). In contrast to the earlier aggregate comparison, the detailed breakdown shows that SurgFUTR-TS achieves top performance on \textit{smoke} detection at all three horizons (0.46/0.49/0.40), outperforming Phase (0.47/0.38/0.33) as well as the task-specific anticipation baselines AVT (0.37/0.35/0.36), RULSTM (0.37/0.29/0.23), and SF-RULSTM (0.37/0.33/0.32). For \textit{occlusion} detection, task-specific and self-supervised baselines prevail: SF-RULSTM leads at 1s and 3s (0.36/0.37), while DINO-R50 leads at longer horizons (0.36 at 5s) and MoCov2-R50 remains competitive, all outperforming SurgFUTR-TS (0.34/0.27/0.24). Among SurgFUTR variants, TS demonstrates the most balanced performance across events and horizons, particularly at shorter anticipation windows (1s/3s), where it maintains competitive or leading scores on \textit{smoke} and \textit{bleeding} detection.

The event-wise pattern suggests that the contribution of state-change pretraining is event-dependent rather than uniform. SurgFUTR-TS is strongest for \textit{smoke} at all three horizons, whereas \textit{bleeding} is better handled by Kinetics-400 pretraining and \textit{occlusion} is better handled by SF-RULSTM at 1s/3s and DINO-R50 at 5s. One possible explanation is that, for \textit{bleeding}, visually predictive cues may already be present in the preceding clip, allowing strong general video pretraining to anticipate the event effectively without relying as much on explicit state-change modeling. By contrast, \textit{smoke} may exhibit greater short-term temporal variation across horizons, making the temporally structured representation learned through state-change pretraining more useful. Overall, the results support a more selective interpretation of the method: SurgFUTR-TS is particularly effective for some event classes, but not uniformly dominant across the entire benchmark.


\subsection{Ablation Studies}

\begin{figure}[h]
\centering
    \includegraphics[width=1.00\columnwidth]{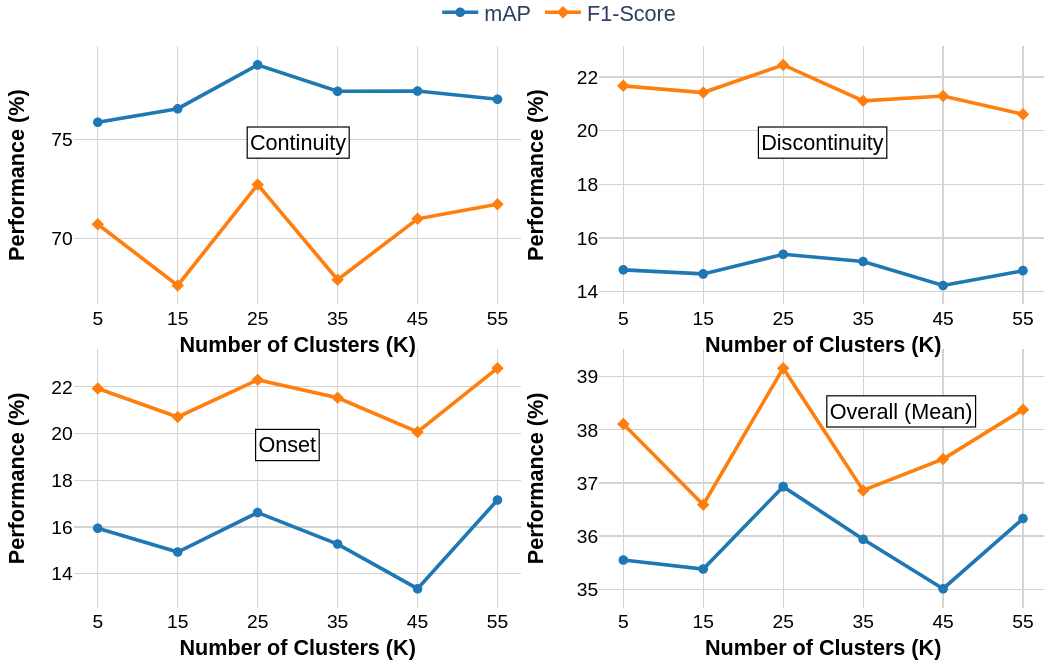} 
    \caption{Ablation on impact of cluster size (K) on state-change prediction performance in CholecT50 dataset.}
    \label{fig:num_clusters_ablation}
\end{figure}

\paragraph{\textbf{(a) Cluster size (K)}}
We test cluster counts \(K \in \{5, 15, 25, 35, 45, 55\}\) during teacher-student pretraining. Figure~\ref{fig:num_clusters_ablation} shows \(K = 25\) achieves optimal performance (36.93\% mAP, 39.16\% F1-Score), providing an effective balance between granularity and generalization. Performance degrades at \(K = 15\) (F1: 36.59\%) and \(K = 45\) (mAP: 35.01\%), suggesting that too few clusters oversimplify surgical state representations while excessive clusters introduce fragmentation that hinders feature learning. 
Interestingly, at \(K = 15\), continuity detection exhibits divergent metric behavior: mAP increases to 76.56\% while F1 drops to 67.64\% (vs. 70.73\% at \(K=5\)), indicating improved ranking quality but degraded precision-recall balance---likely due to insufficient cluster granularity causing ambiguous decision boundaries for continuous state predictions. Notably, very low cluster counts (\(K = 5\): 35.55\% mAP) maintain competitive performance by capturing coarse-grained state changes, while very high counts (\(K = 55\): 36.33\% mAP) partially recover through increased representational capacity, though neither matches the discriminative-generalization trade-off achieved at \(K = 25\). The consistent peak across both continuity (78.77\% mAP) and transition events (discontinuity: 15.39\%, onset: 16.62\%) at \(K = 25\) validates this cluster granularity as optimal for modeling diverse surgical state-change patterns.

\begin{figure}[h]
\centering
    \includegraphics[width=1.00\columnwidth]{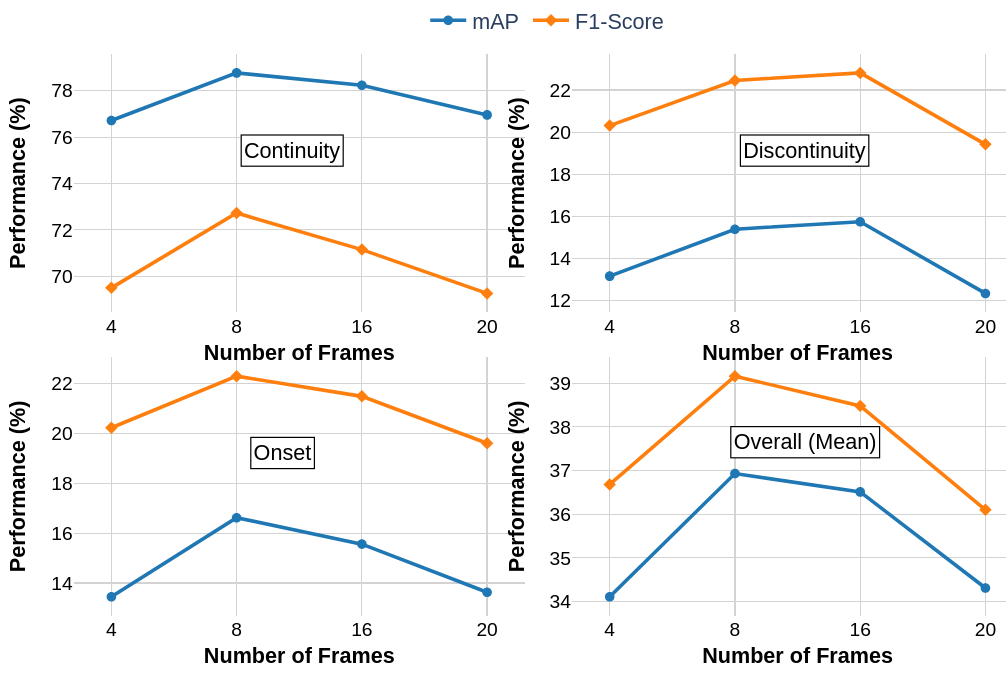} 
    \caption{Ablation on number of frames for State Change prediction in the CholecT50 dataset in SurgFUTR-TS.}
    \label{fig:num_frames_ablation}
\end{figure}

\paragraph{\textbf{(b) Impact of number of frames in the video clip}}
Figure~\ref{fig:num_frames_ablation} shows how frame sampling affects temporal modeling. Performance varies significantly across configurations: 4 frames (baseline) achieves 34.11\% mAP; 8 frames yields optimal mAP (36.93\%) with trade-offs in F1-Score; 16 frames balances both metrics (36.51\% mAP, 38.48\% F1); while 20 frames shows degraded performance. The 8-frame configuration provides the best mAP while maintaining efficiency, indicating that moderate temporal sampling captures sufficient context for surgical state transitions.

\paragraph{\textbf{(c) Impact of losses in SurgFUTR-TS}}
Table~\ref{tab:results:ablation:loss_fn} shows the contribution of different loss components. Each component adds distinct value: $\mathcal{L}^{distill}_{SC}$ improves continuity prediction and overall performance by transferring knowledge about state transitions; $\mathcal{L}^{distill}_{SE}$ enhances onset prediction through better state representations; and $\mathcal{L}_{FUTR}$ provides the most significant boost across all metrics by enabling future state anticipation. The full model with all three losses achieves optimal performance, indicating that these components work synergistically to enhance state-change prediction capabilities.

\paragraph{\textbf{(d) Impact of components in SurgFUTR}}
We perform component-wise ablations using a progressive removal strategy on the CholecT50 dataset, since the modules in SurgFUTR are structurally dependent rather than independent. Starting from the full SurgFUTR-TS model, we remove \emph{ActDyn}, then the \emph{State Graph}, and finally the \emph{State Encoder}, following the hierarchy of the proposed state modeling pipeline. In this way, the ablations measure how performance changes as the model is gradually simplified, from removing explicit state-transition modeling, to removing graph-based state reasoning, and finally to removing the clustered state representation itself.

\begin{table}[h]
    \centering
    \caption{Progressive ablation of SurgFUTR components on CholecT50 for long-horizon anticipation, evaluated using RSD and phase transition prediction. Lower is better.}
    \label{tab:results:ablation_rsd_phase}
    \setlength{\tabcolsep}{2pt}
    \resizebox{\columnwidth}{!}{%
    \begin{tabular}{lccccc}
        \toprule
        \multirow{2}{*}{\textbf{Method}} &
        \multirow{2}{*}{\textbf{RSD}} &
        \multicolumn{4}{c}{\textbf{Phase Transition}} \\
        \cmidrule(lr){3-6}
        & & \textbf{2 min} & \textbf{3 min} & \textbf{5 min} & \textbf{mean} \\
        \midrule
        \rowcolor{cyan!15}
        SurgFUTR-TS &
        $\textbf{1.465}\pm\scriptstyle{0.041}$ &
        $\textbf{0.196}\pm\scriptstyle{0.005}$ &
        $\textbf{0.305}\pm\scriptstyle{0.006}$ &
        $\textbf{0.557}\pm\scriptstyle{0.012}$ &
        $\textbf{0.353}\pm\scriptstyle{0.007}$ \\

        \quad $\hookrightarrow$ w/o ActDyn &
        $1.607\pm\scriptstyle{0.037}$ &
        $0.272\pm\scriptstyle{0.033}$ &
        $0.410\pm\scriptstyle{0.043}$ &
        $0.687\pm\scriptstyle{0.024}$ &
        $0.457\pm\scriptstyle{0.033}$ \\

        \quad $\hookrightarrow$ w/o State Graph &
        $1.651\pm\scriptstyle{0.006}$ &
        $0.276\pm\scriptstyle{0.051}$ &
        $0.394\pm\scriptstyle{0.049}$ &
        $0.681\pm\scriptstyle{0.071}$ &
        $0.450\pm\scriptstyle{0.054}$ \\

        \quad $\hookrightarrow$ w/o State Encoder &
        $1.603\pm\scriptstyle{0.104}$ &
        $0.297\pm\scriptstyle{0.041}$ &
        $0.420\pm\scriptstyle{0.030}$ &
        $0.706\pm\scriptstyle{0.013}$ &
        $0.474\pm\scriptstyle{0.028}$ \\
        \bottomrule
    \end{tabular}
    }
\end{table}

Table~\ref{tab:results:ablation_rsd_phase} reports a progressive ablation of SurgFUTR on long-horizon anticipation, namely RSD and phase transition prediction, which require forecasting workflow evolution several minutes ahead. Starting from the full SurgFUTR-TS model, we first remove \emph{ActDyn}, then remove the \emph{State Graph}, and finally remove the \emph{State Encoder}. This progressive setup follows the dependency among the modules, since ActDyn operates on graph-refined state representations, while the State Graph itself depends on the learned state encoding. The full SurgFUTR-TS model achieves the best performance on all metrics, with the lowest RSD ($1.465$) and the lowest mean phase transition error ($0.353$), showing that the complete design is most effective for modeling long-range temporal dependencies. Removing ActDyn already leads to a clear degradation, increasing RSD to $1.607$ and the mean phase transition error to $0.457$, which highlights the importance of explicitly modeling how surgical states evolve over time. Removing the State Graph further worsens RSD to $1.651$, suggesting that structured reasoning over latent surgical states is particularly important for capturing global workflow progression. Finally, removing the State Encoder increases the mean phase transition error to $0.474$, indicating that strong state representations provide the foundation for stable long-horizon transition prediction. Overall, the progressive performance drop confirms that the three components contribute hierarchically and complement each other in supporting robust long-range anticipation.

Table~\ref{tab:results:quantitative:cyantr_results_ablation} presents the same progressive ablation on short-term anticipation tasks, namely cystic triplet anticipation on CholecT50 and event anticipation on CholecTrack20, where the prediction horizon is only a few seconds. Here again, the full SurgFUTR-TS model performs best across both tasks, achieving $50.29/43.91/45.61$ on SFP-IV and $46.12/41.40/46.48$ on SFP-V in terms of mAP/F1/Acc. Removing ActDyn causes an immediate drop on both benchmarks, showing that explicit temporal transition modeling is beneficial even in the short-term setting. Removing the State Graph leads to a further decline, especially on SFP-V, indicating that short-term event anticipation also benefits from reasoning over interactions between latent states. Finally, removing the State Encoder further weakens performance, particularly in F1, suggesting that discriminative state encoding is important for recognizing subtle near-future changes. Taken together, these results show that the hierarchical design of SurgFUTR benefits not only long-range workflow forecasting but also short-term anticipation of fine-grained surgical actions and events.

\begin{table}[t]
    \centering
    \caption{Progressive ablation of SurgFUTR components on short-term anticipation tasks: SFP-IV cystic triplet anticipation on CholecT50 and SFP-V event anticipation on CholecTrack20. Performance is reported using mAP, F1, and accuracy; higher is better.}
    \label{tab:results:quantitative:cyantr_results_ablation}
    \setlength{\tabcolsep}{14pt}
    \resizebox{\columnwidth}{!}{%
    \begin{tabular}{lcccc}
        \toprule
        {\textbf{Method}} & \textbf{mAP (\%)} & \textbf{F1 (\%)} & \textbf{Acc (\%)} \\
        \midrule
        \multicolumn{4}{l}{\textit{SFP-IV Cystic Triplet Anticipation}} \\
        \rowcolor{cyan!15}
        SurgFUTR-TS   & $\textbf{50.29}$ & $\textbf{43.91}$ & $\textbf{45.61}$ \\
        \quad $\hookrightarrow$ w/o ActDyn & $46.43$ & $33.20$ & $36.74$ \\
        \quad $\hookrightarrow$ w/o State Graph & $43.88$ & $36.29$ & $38.52$ \\
        \quad $\hookrightarrow$ w/o State Encoder & $46.83$ & $32.89$ & $38.78$ \\

        \midrule
        \multicolumn{4}{l}{\textit{SFP-V Event Anticipation}} \\
        \rowcolor{cyan!15}
        SurgFUTR-TS   & $\textbf{46.12}$ & $\textbf{41.40}$ & $\textbf{46.48}$ \\
        \quad $\hookrightarrow$ w/o ActDyn & $45.16$ & $34.59$ & $43.24$ \\
        \quad $\hookrightarrow$ w/o State Graph & $43.36$ & $25.56$ & $38.15$ \\
        \quad $\hookrightarrow$ w/o State Encoder & $46.02$ & $31.06$ & $41.45$ \\
        
        \bottomrule 
    \end{tabular}
    }
\end{table}

\noindent{
\textbf{(e) Is binary state change enough?}
We further investigate whether a simpler binary state-change objective is sufficient, or whether a finer 4-class formulation is needed. In our default setup, state change between two consecutive clips is modeled using four classes: continuity ($1\!\rightarrow\!1$), discontinuity ($1\!\rightarrow\!0$), onset ($0\!\rightarrow\!1$), and background ($0\!\rightarrow\!0$). The binary variant is derived directly from this setup by collapsing continuity and background into a \emph{no-change} class ($0$), and discontinuity and onset into a \emph{change} class ($1$). This preserves the same temporal structure while removing the distinction between different transition types.

Table~\ref{tab:results:ablation_rsd_phase_2sc} compares our default 4-class state-change pretraining objective with a simpler binary formulation on long-horizon anticipation on CholecT50. The 4-class formulation consistently outperforms the binary variant on both RSD and phase transition prediction, reducing RSD from $1.627$ to $1.465$ and lowering the mean phase transition error from $0.439$ to $0.353$. The gains are consistent across all prediction horizons, with improvements at 2, 3, and 5 min, indicating that fine-grained supervision provides a stronger learning signal for modeling long-range workflow evolution. Unlike a binary change/no-change objective, the 4-class formulation forces the model to distinguish qualitatively different temporal situations: whether the workflow is stably continuing, undergoing a meaningful transition, entering an onset-like change region, or remaining in background context. In this way, the model is not only asked to detect that a change exists, but also to characterize the type and temporal role of that change. This richer supervision encourages the learned representation to capture multiple aspects of workflow evolution, including persistence, transition dynamics, and the structure of local temporal boundaries, rather than collapsing all variation into a single notion of change. As a result, the 4-class objective yields more coherent temporal representations and improves robustness for long-horizon forecasting.

\begin{table}[h]
    \centering
    \caption{Effect of state-change label granularity on long-horizon anticipation in CholecT50. Lower is better.}
    \label{tab:results:ablation_rsd_phase_2sc}
    \setlength{\tabcolsep}{6pt}
    \resizebox{\columnwidth}{!}{%
    \begin{tabular}{lccccc}
        \toprule
        \multirow{2}{*}{\textbf{SC-Target}} &
        \multirow{2}{*}{\textbf{RSD}} &
        \multicolumn{4}{c}{\textbf{Phase Transition}} \\
        \cmidrule(lr){3-6}
        & & \textbf{2 min} & \textbf{3 min} & \textbf{5 min} & \textbf{mean} \\
        \midrule
        \rowcolor{cyan!15}
        4-class &
        $\textbf{1.465}\pm\scriptstyle{0.041}$ &
        $\textbf{0.196}\pm\scriptstyle{0.005}$ &
        $\textbf{0.305}\pm\scriptstyle{0.006}$ &
        $\textbf{0.557}\pm\scriptstyle{0.012}$ &
        $\textbf{0.353}\pm\scriptstyle{0.007}$ \\

        2-class &
        $1.627\pm\scriptstyle{0.048}$ &
        $0.262\pm\scriptstyle{0.024}$ &
        $0.383\pm\scriptstyle{0.027}$ &
        $0.671\pm\scriptstyle{0.038}$ &
        $0.439\pm\scriptstyle{0.029}$ \\
        \bottomrule
    \end{tabular}
    }
\end{table}

Table~\ref{tab:results:quantitative:cy_ct_results_2sc} further evaluates the same design choice on short-term anticipation tasks. On SFP-IV cystic triplet anticipation, the 4-class formulation yields substantial gains across all metrics, improving mAP from $45.64$ to $50.29$, F1 from $29.50$ to $43.91$, and accuracy from $35.42$ to $45.61$. This suggests that, for fine-grained short-horizon anticipation, separating different forms of temporal change is especially beneficial: the model is encouraged not only to detect that an event boundary may be approaching, but also to distinguish whether the local visual dynamics correspond to stable continuation, transition, onset, or background context. Such distinctions are particularly relevant when anticipating safety-critical triplets, where short-term temporal cues can differ substantially in meaning even when all would be treated identically under a binary change/no-change target. On SFP-V event anticipation, the comparison is slightly more nuanced: the binary formulation obtains a marginally higher mAP ($47.00$ vs.\ $46.12$), but the 4-class formulation achieves better F1 and accuracy, reaching $41.40$ and $46.48$, respectively. This indicates that even when a binary objective can recover competitive retrieval performance, the 4-class objective produces more balanced event recognition by encouraging the model to encode a richer notion of temporal change instead of collapsing all dynamic behavior into a single positive class.} 
Taken together, these results suggest that fine-grained state-change supervision is particularly important for long-horizon anticipation and generally provides a more balanced trade-off on short-term surgical anticipation tasks, even if it is not uniformly superior on every individual metric, which is why we adopt the 4-class formulation as the default setting in SurgFUTR.

\begin{table}[h]
    \centering
    \caption{Effect of state-change label granularity on short-term anticipation. Results are reported on CholecT50 (SFP-IV) and CholecTrack20 (SFP-V); higher is better.}
    \label{tab:results:quantitative:cy_ct_results_2sc}
    \setlength{\tabcolsep}{17pt}
    \resizebox{\columnwidth}{!}{%
    \begin{tabular}{lcccc}
        \toprule
        {\textbf{SC-Target}} & \textbf{mAP (\%)} & \textbf{F1 (\%)} & \textbf{Acc (\%)} \\
        \midrule
        \multicolumn{4}{l}{\textit{SFP-IV Cystic Triplet Anticipation}} \\
        \rowcolor{cyan!15}
        4-class   & $\textbf{50.29}$ & $\textbf{43.91}$ & $\textbf{45.61}$ \\
        2-class & $45.64$ & $29.50$ & $35.42$ \\

        \midrule
        \multicolumn{4}{l}{\textit{SFP-V Event Anticipation}} \\
        \rowcolor{cyan!15}
        4-class   & $46.12$ & $\textbf{41.40}$ & $\textbf{46.48}$ \\
        2-class & $\textbf{47.00}$ & $38.88$ & $45.67$ \\
        \bottomrule 
    \end{tabular}
    }
\end{table}

\subsection{Qualitative results}
\begin{figure}[h]
\centering
    \includegraphics[width=1.0\columnwidth]{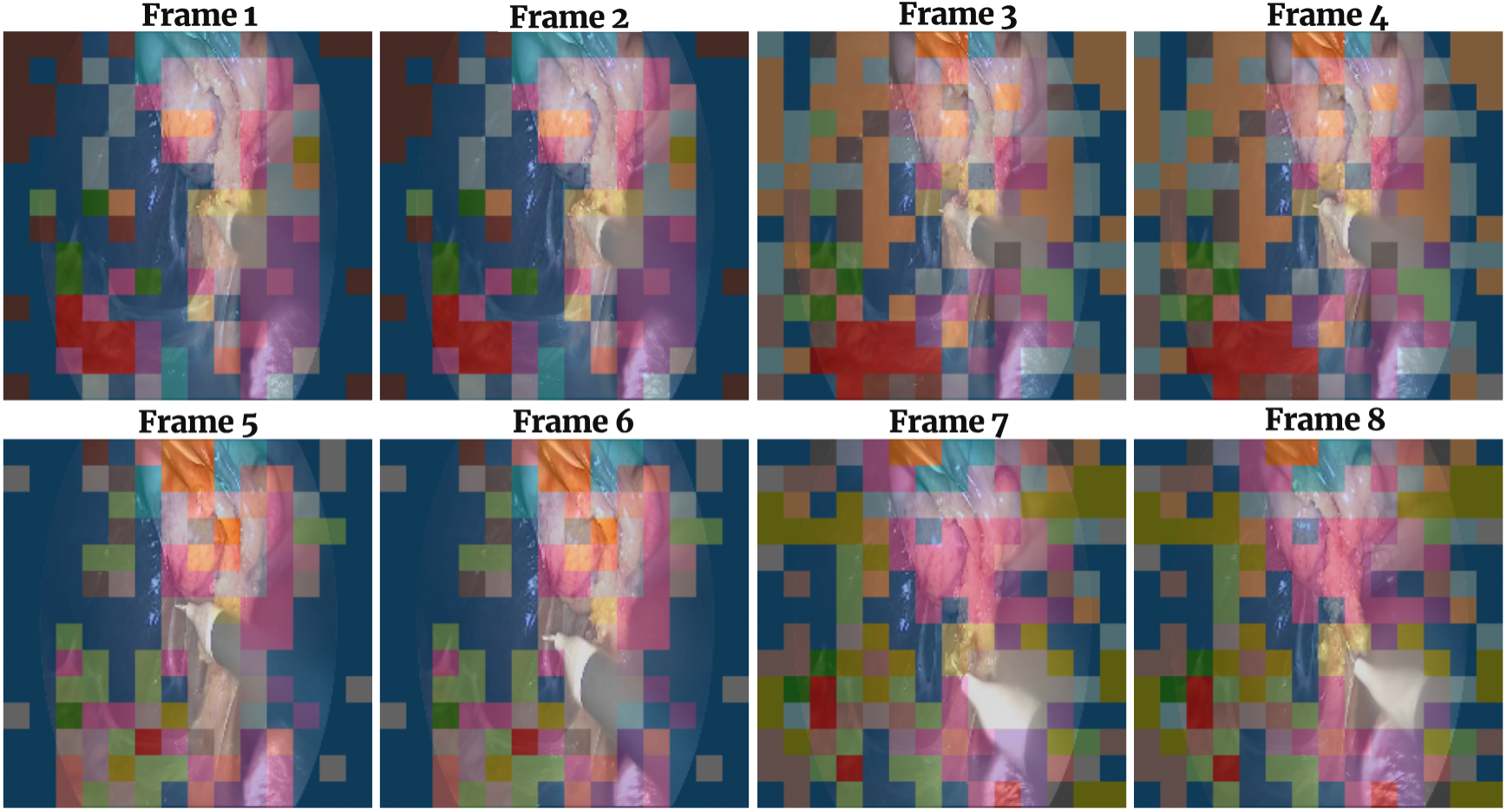} 
    \caption{Clustering visualization on CholecT50 validation videos. State-change pretraining produces emergent part-level clusters for \textit{hook} (tip and body at bottom-right) and \textit{grasper} (tip and body at top-middle).}
    \label{fig:qual_cluster1}
\end{figure}

\begin{figure}[h]
\centering
    \includegraphics[width=1.0\columnwidth]{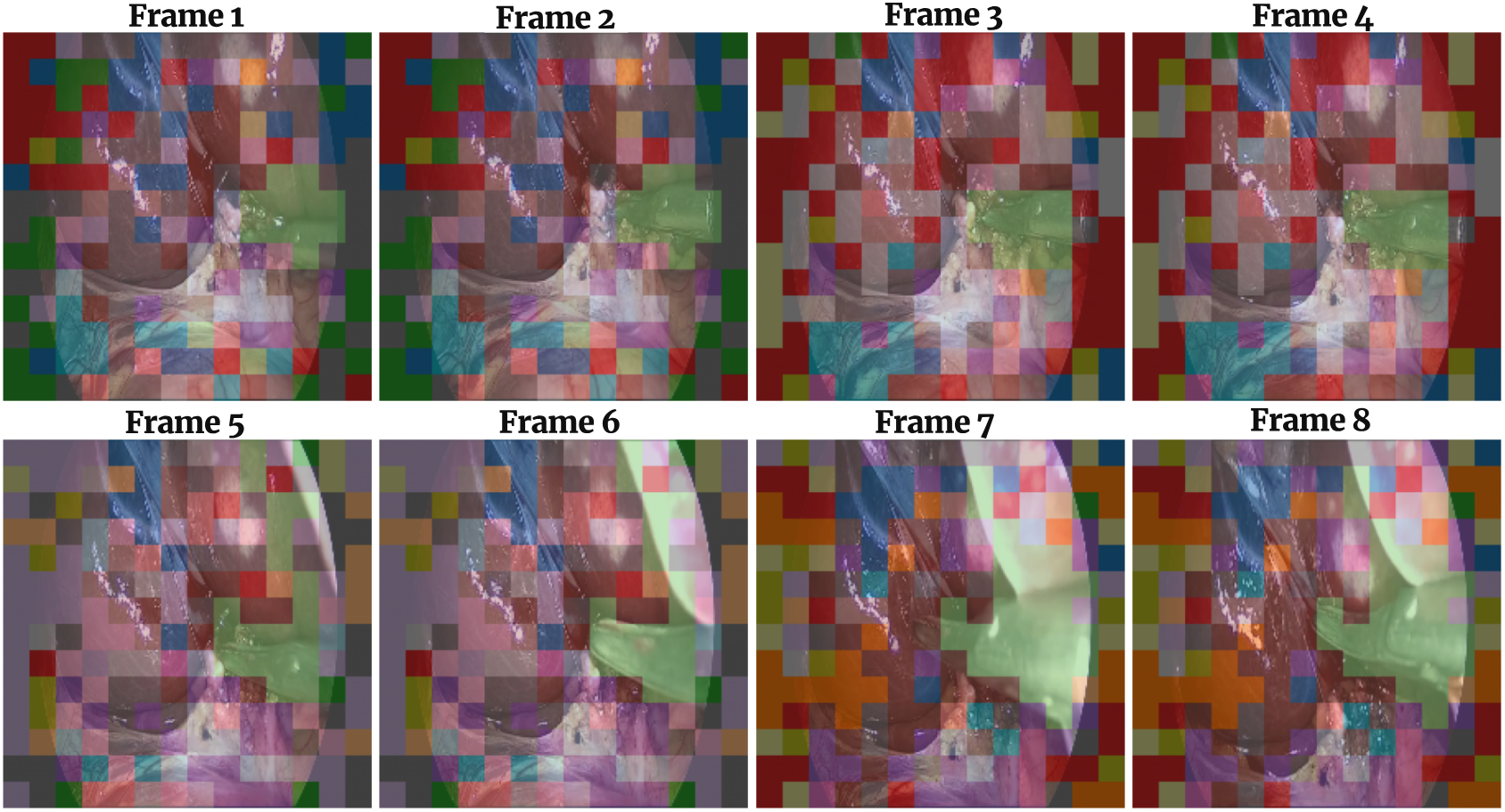} 
    \caption{Clustering visualization on CholecT50 validation videos. State-change pretraining produces emergent part-level clusters for \textit{scissor} components (tip and body at middle-right) and distinct tissue context groupings.}
    \label{fig:qual_cluster2}
\end{figure}

\begin{figure}[h]
\centering
    \includegraphics[width=1.0\columnwidth]{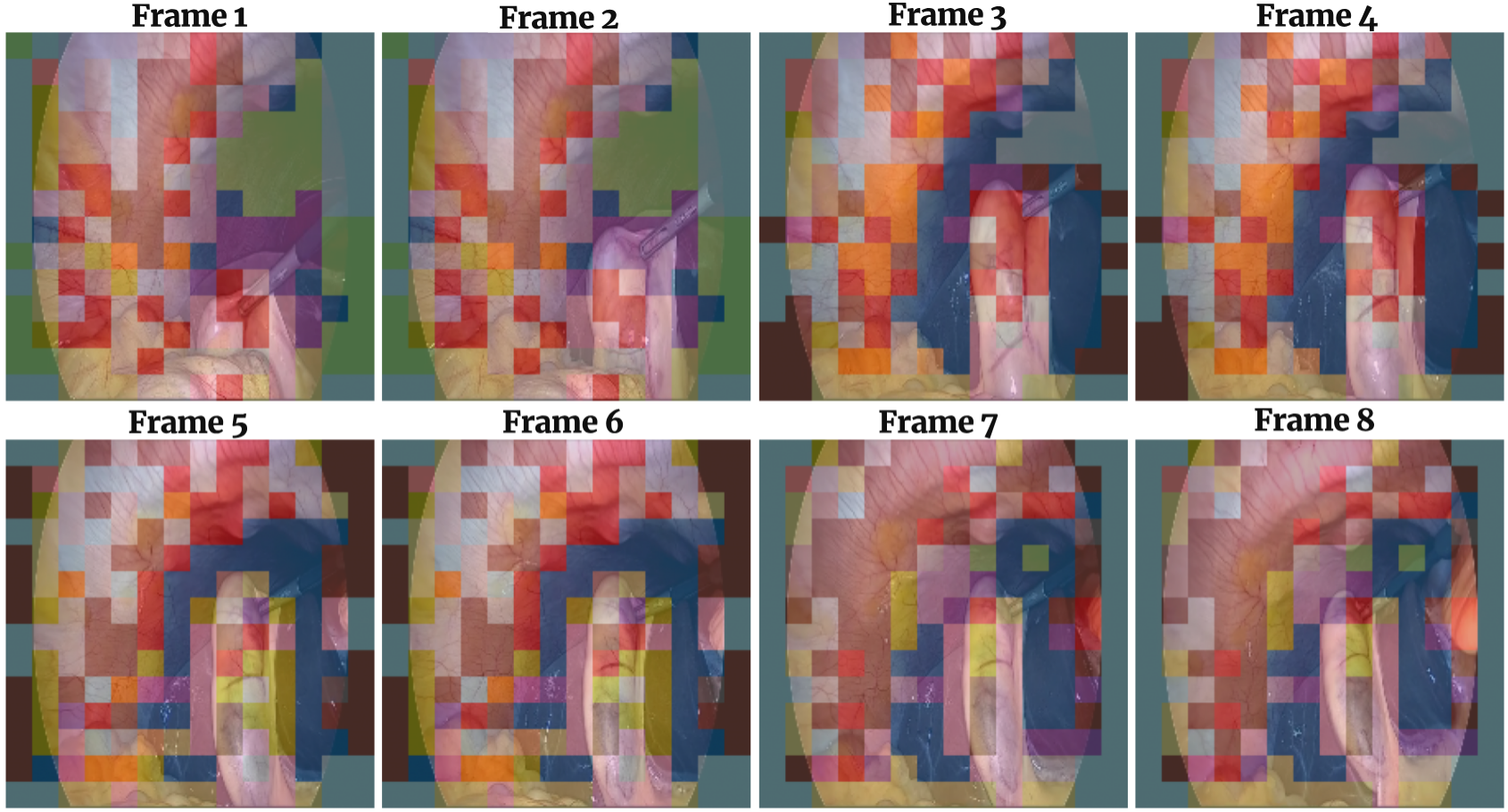} 
    \caption{Clustering visualization on CholecT50 validation videos. State-change pretraining produces emergent part-level clusters for \textit{grasper} components (tip and body at middle-right) and distinct tissue context groupings.}
    \label{fig:qual_cluster3}
\end{figure}

\begin{figure}[h]
\centering
    \includegraphics[width=1.0\columnwidth]{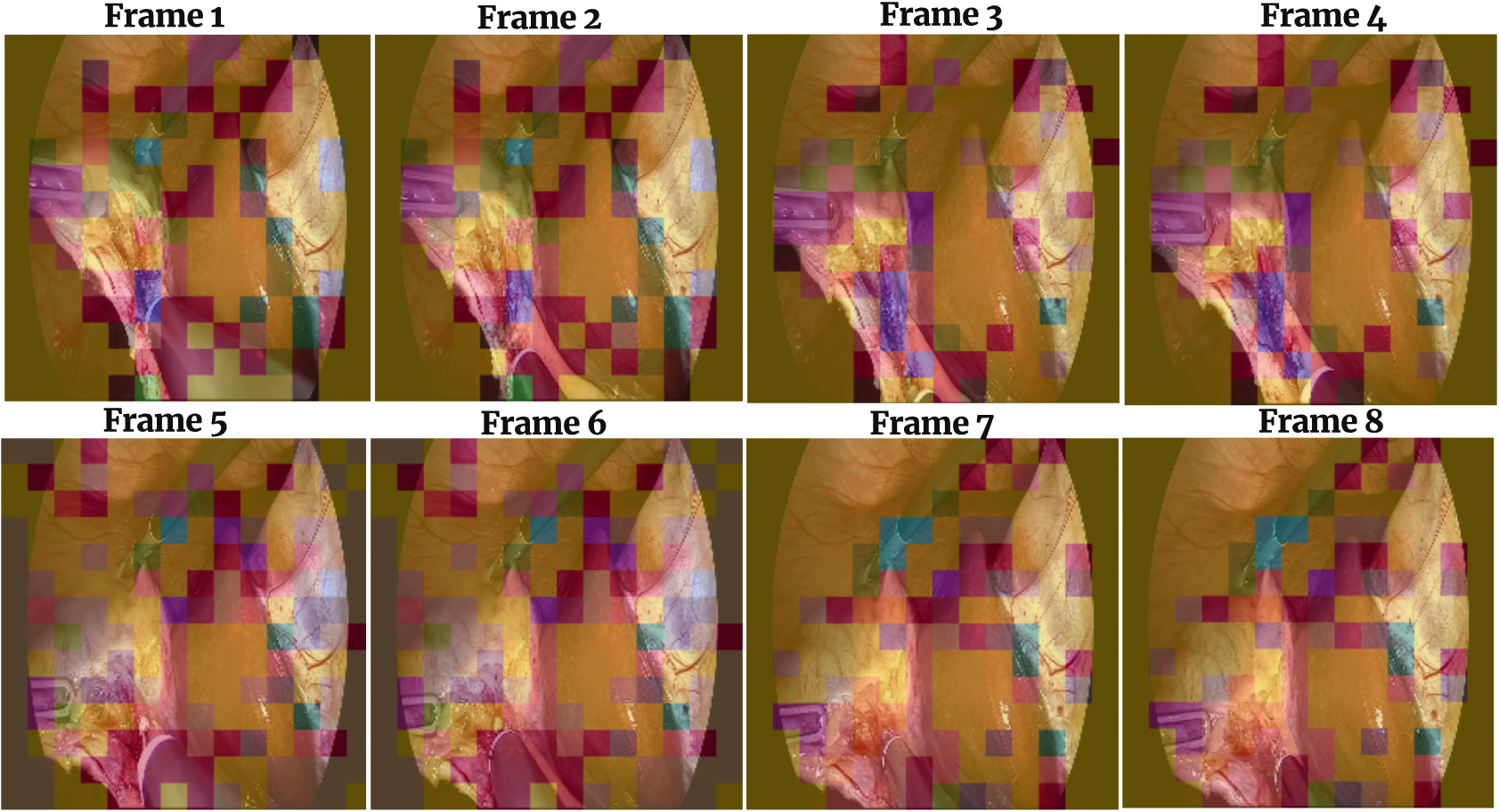} 
    \caption{Clustering visualization on CholecT50 validation videos. State-change pretraining produces instrument-specific clusters: grasper instances (middle-left and middle-top) and hook (middle-bottom) are grouped distinctly.}
    \label{fig:qual_cluster4}
\end{figure}

We assess the quality of clusters produced by our Sinkhorn-based spatio-temporal clustering in the state encoder. Figures~\ref{fig:qual_cluster1}--\ref{fig:qual_cluster4} show SurgFUTR-TS clustering results for three randomly selected CholecT50~\citep{nwoye2021rendezvous} validation videos under verb-level state-change pretraining, where patches with identical colors represent the same cluster centroid.

Figure~\ref{fig:qual_cluster1} demonstrates that verb-level state-change pretraining produces emergent instrument tracking capabilities. Clusters automatically identify \textit{hook} instruments (gray, bottom-left) by tracking the entire instrument body and \textit{grasper} instruments (orange, top-middle) by focusing on the functional tip region. Without explicit instrument supervision, our clustering discovers instrument-tissue interaction zones that become implicitly encoded in the centroid-based state representations.

Figure~\ref{fig:qual_cluster2} reveals how clusters maintain coherent \textit{scissor} instrument tracking, with the green cluster consistently localizing the tip and shaft regions throughout the video sequence. This consistent spatial localization creates stable temporal anchors that allow SurgFUTR-TS to learn underlying surgical action patterns. 
The ActDyn module exploits these temporally consistent cluster assignments to model centroid transitions in the state space, thereby improving anticipation performance on critical downstream tasks: predicting cystic-structure related action triplets.


Figure~\ref{fig:qual_cluster3} demonstrates adaptive cluster behavior: the purple cluster tracks the \textit{grasper} shaft and tip through frame 6, then transitions to a yellow cluster for the new instrument segment. Simultaneously, the light pink cluster consistently follows the gallbladder-liver boundary across frames. These patterns reveal that our state encoder learns object-centric centroids that capture both instrument dynamics and anatomical context. This enables SurgFUTR-TS to effectively detect temporal changes and map them to the four state-change categories: continuity, discontinuity, onset, and background.

Figure~\ref{fig:qual_cluster4} demonstrates instance-level clustering in scenarios with multiple instruments of the same type. Despite both being graspers, the instrument at middle-left (violet) and middle-top (green) are assigned to distinct clusters, while the hook at middle-bottom (pink) forms a separate cluster. This instance-level differentiation, rather than purely semantic grouping, suggests that state-change pretraining under fine-grained verb-level supervision captures not only instrument identity but also contextual differences in spatial positioning and potential interactions with anatomical structures.

\subsection{Computational Complexity}

Tables~\ref{tab:pretraining_complexity} and~\ref{tab:downstream_complexity_rsd} summarize the computational footprint of our models at pretraining and downstream fine-tuning, respectively. We report trainable parameter count together with per-iteration training cost, and for downstream evaluation we additionally report per-sample inference latency. This distinction is important because SurgFUTR-TS introduces extra complexity during pretraining, whereas only the student model is retained at deployment time.


At the pretraining stage, SurgFUTR-Lite and SurgFUTR-S remain relatively compact, with 30.54M and 55.54M trainable parameters, and similar training cost (159.459 and 163.781 ms/iter). SurgFUTR-TS is more expensive because it uses a multi-stage pipeline with paired current/future clip modeling and, in the second stage, additional teacher-student distillation. As a result, Stage 1 matches SurgFUTR-S in trainable parameter count (55.54M) but increases training cost to 308.416 ms/iter due to the heavier two-clip state-change training setup, while Stage 2 reaches 62.68M parameters and 379.510 ms/iter. This confirms that the full teacher-student formulation adds non-trivial pretraining overhead compared with the lighter variants. At the same time, SurgFUTR-TS remains substantially smaller than AVT at pretraining time (62.68M vs.\ 387.57M parameters in the heaviest stage) and is comparable to or lighter than multi-stage recurrent anticipation baselines such as RULSTM and SF-RULSTM in parameter count, although its training cost is still higher than SurgFUTR-Lite and SurgFUTR-S.

\begin{table}[h]
    \centering
    \caption{Parameter count and training cost at the pretraining stage. Trainable parameters are reported in millions (M). Train cost is reported in milliseconds per iteration (ms/iter) under the same measurement setting for all methods. Stage 1 and Stage 2 denote multi-stage pretraining pipelines.}
    \label{tab:pretraining_complexity}
    \setlength{\tabcolsep}{6pt}
    \resizebox{\columnwidth}{!}{%
    \begin{tabular}{llccc}
    \toprule
        \textbf{Method} &
        \textbf{Backbone} &
        \textbf{Stage} &
        \textbf{Trainable Params (M)} &
        \textbf{Train Cost} \\
        \midrule

        \multicolumn{5}{l}{\textit{Our Method (State-Change Pretraining)}} \\
        SurgFUTR-Lite & ViT-S & -- & 30.54 & 159.459 \\
        SurgFUTR-S & ViT-S & -- & 55.54 & 163.781 \\
        SurgFUTR-TS & ViT-S & 1 & 55.54 & 308.416 \\
        SurgFUTR-TS & ViT-S & 2 & 62.68 & 379.510 \\

        \midrule
        \multicolumn{5}{l}{\textit{Task-Specific Baselines}} \\
        AVT & ViT-B & -- & 387.57 & 516.943 \\
        RULSTM & Inception-V3 & 1 & 23.89 & 326.028 \\
        RULSTM & Inception-V3 & 2 & 40.69 & 350.041 \\
        SF-RULSTM & Inception-V3 & 1 & 23.89 & 355.692 \\
        SF-RULSTM & Inception-V3 & 2 & 57.49 & 359.943 \\

        \bottomrule
    \end{tabular}
    }
\end{table}

\begin{table}[h]
    \centering
    \caption{Parameter count and computational cost for the downstream RSD task. Trainable parameters are reported in millions (M). Train cost is reported in milliseconds per iteration (ms/iter), and inference latency is reported in milliseconds per sample (ms/sample, batch size 1) under the same measurement setting for all methods.}
    \label{tab:downstream_complexity_rsd}
    \setlength{\tabcolsep}{5pt}
    \resizebox{\columnwidth}{!}{%
    \begin{tabular}{llccc}
    \toprule
        \textbf{Method} &
        \textbf{Backbone} &
        \textbf{Trainable Params (M)} &
        \textbf{Train Cost} &
        \textbf{Inf. Latency} \\
        \midrule

        \multicolumn{5}{l}{\textit{General SSL / Foundation Pretraining}} \\
        VideoMAE & ViT-S & 21.88 & 111.117 & 11.740 \\
        MoCoV2 & ResNet50 & 23.30 & 178.838 & 64.912 \\
        DINO & ResNet50 & 23.30 & 177.822 & 72.117 \\
        EndoViT & ViT-B & 85.80 & 505.968 & 57.572 \\
        SurgeNetXL & CaFormerS18 & 23.24 & 538.999 & 129.791 \\

        \midrule
        \multicolumn{5}{l}{\textit{Task-Specific Baselines}} \\
        AVT & ViT-B & 389.56 & 496.078 & 63.999 \\
        RULSTM & Inception-V3 & 40.68 & 358.943 & 143.016 \\
        SF-RULSTM & Inception-V3 & 57.48 & 363.259 & 158.183 \\
        TimeLSTM & ResNet152 & 58.16 & 487.620 & 169.503 \\
        TransLocal & ResNet101 & 42.50 & 341.243 & 124.239 \\
        RSDNet & ResNet152 & 70.74 & 491.330 & 171.420 \\
        BD-Net & DenseNet169 & 17.08 & 677.158 & 230.968 \\

        \midrule
        \multicolumn{5}{l}{\textit{Our Method (State-Change Pretraining)}} \\
        SurgFUTR-Lite & ViT-S & 30.54 & 118.447 & 13.438 \\
        SurgFUTR-S & ViT-S & 21.88 & 110.691 & 11.252 \\
        SurgFUTR-TS & ViT-S & 26.61 & 121.101 & 14.820 \\

        \bottomrule 
    \end{tabular}
    }
\end{table}

The downstream picture is more favorable, and this trend is consistent across the downstream tasks we consider. For brevity, Table~\ref{tab:downstream_complexity_rsd} reports the complexity analysis on the RSD task, since it includes the broadest set of comparable baselines. Once transferred downstream, SurgFUTR-TS requires 26.61M trainable parameters, which is markedly smaller than AVT (389.56M), EndoViT (85.80M), RSDNet (70.74M), and the recurrent anticipation baselines RULSTM (40.68M) and SF-RULSTM (57.48M). Its downstream training cost is 121.101 ms/iter, only slightly above SurgFUTR-Lite (118.447 ms/iter) and SurgFUTR-S (110.691 ms/iter), while remaining well below several task-specific baselines such as AVT, RULSTM, SF-RULSTM, TimeLSTM, RSDNet, and BD-Net. Inference latency follows the same trend: SurgFUTR-TS runs at 14.820 ms/sample, which is close to the lighter SurgFUTR variants (13.438 and 11.252 ms/sample), slightly above VideoMAE (11.740 ms/sample), and much faster than the ResNet, Inception, and DenseNet-based baselines.

Overall, these results show a clear accuracy-complexity trade-off. The proposed state-change framework incurs additional cost during pretraining, especially for the teacher-student variant, but this overhead does not translate into a large deployment penalty. At downstream inference time, SurgFUTR-TS remains lightweight relative to the strongest task-specific baselines, suggesting that the added pretraining complexity is primarily an offline cost rather than a runtime burden.


\section{Limitations}
While our state-change formulation shows promise for future prediction in the surgical domain, the approach involves a complex task and system design requiring substantial development effort. 
Extending context beyond the current $3$‑second clips could enrich understanding of surgical state changes, though achieving this is non‑trivial and would demand significant computational resources and system complexity to maintain and operate over the full history of frames.
In addition, some downstream improvements, particularly for remaining surgery duration prediction, are moderate in absolute terms even when they are consistent across baselines. This suggests that the value of state-change pretraining should be interpreted in the broader context of multi-task performance and task difficulty, rather than through any single downstream metric in isolation.

\section{Conclusion}
Surgical future prediction represents an important capability for decision support systems in the OR, yet has received limited attention compared to recognition tasks. While existing anticipation methods focus on utilizing coarse-grained surgical information such as phases and tools, fine-grained annotations offer untapped potential for developing robust future prediction capabilities.
In this work, we introduce a unified state-change modeling framework for surgical future prediction. Our approach is grounded in the hypothesis that models capable of understanding state transitions between adjacent video clips develop future-aware representations compared to direct feature prediction methods. We propose SurgFUTR, a novel teacher-student architecture featuring clustering-based state representation, a state graph for centroid interaction modeling, and our ActDyn module for future centroid transition prediction. The teacher processes current and future clips while the student learns to predict future states using only current observations, enabled by ActDyn's ability to model centroid transitions in the state space.
To comprehensively evaluate future prediction capabilities, we introduce SFPBench, a surgical future prediction benchmark encompassing three long-term forecasting and two short-term anticipation tasks across multiple surgical procedures. 
Extensive experiments demonstrate that our state-change pretraining provides generally favorable and often competitive transfer performance relative to strong baselines, including recent surgical foundation models and task-specific anticipation baselines, across the benchmark tasks. The detailed analyses further show that the benefits are not uniform across all tasks, horizons, and event categories, but are strongest in settings where temporally structured state-change information appears particularly relevant.
Notably, cross-procedure transfer experiments from cholecystectomy to gastric bypass procedures show encouraging generalization of our approach, with SurgFUTR variants achieving encouraging transfer performance relative to general vision and surgical foundation models, although the gains vary across centers and task definitions. We hope our contributions advance surgical AI by demonstrating the effectiveness of fine-grained state-change modeling for future prediction, establishing a comprehensive evaluation framework, and validating cross-procedure generalization capabilities.
We believe this work highlights the importance of temporal reasoning in surgical AI and offers a foundation for developing more capable surgical assistance systems that can adapt across diverse surgical procedures.

\section*{Acknowledgment }
This work has received funding from the European Union (ERC, CompSURG, 101088553). Views and opinions expressed are however those of the authors only and do not necessarily reflect those of the European Union or the European Research Council. Neither the European Union nor the granting authority can be held responsible for them. This work was also supported by French state funds managed by the ANR within the National AI Chair program under Grant ANR-20-CHIA-0029-01 (Chair AI4ORSafety) and within the Investments for the future program under Grant ANR-10-IAHU-02 (IHU Strasbourg). This work was also granted access to the servers/HPC resources managed by CAMMA, IHU Strasbourg, Unistra Mesocentre, and GENCI-IDRIS (Grant AD011013710R3, AD011013710R2, AD011013710R1).

\section*{Declaration}
During the preparation of this work the author used Cursor in order to perform minor grammar checks. After using this tool, the author reviewed and edited the content as needed and take full responsibility for the content of the published article.

\bibliographystyle{model2-names}\biboptions{authoryear}
\bibliography{main}


\end{document}